\documentclass[10pt,twocolumn,letterpaper]{article}

\usepackage[pagenumbers]{cvpr} %

\usepackage[dvipsnames]{xcolor}
\usepackage{microtype}
\usepackage{multirow}
\usepackage{tikz}
\usepackage{array, makecell}
\usetikzlibrary{tikzmark}
\makeatletter
\renewcommand\paragraph{\@startsection{paragraph}{4}{\z@}%
	{0.75ex \@plus.5ex \@minus.2ex}%
	{-1em}%
	{\normalfont\normalsize\bfseries\maybe@addperiod}}
\newcommand{\maybe@addperiod}[1]{#1\@addpunct{.}}
\makeatother

\newcommand{\OURS}{ViewDiff}

\newcommand{\norm}[1]{\left\lVert#1\right\rVert}

\definecolor{cvprblue}{rgb}{0.21,0.49,0.74}
\usepackage[pagebackref,breaklinks,colorlinks,citecolor=cvprblue]{hyperref}

\def\paperID{10257} %
\def\confName{CVPR}
\def\confYear{2024}

\title{\OURS: 3D-Consistent Image Generation with Text-to-Image Models}

\author{
Lukas H{\"o}llein$^{1,2}$ \quad
Alja\v{z} Bo\v{z}i\v{c}$^{2}$ \quad
Norman M{\"u}ller$^2$ \quad
David Novotny$^2$ \quad
Hung-Yu Tseng$^2$ \\
Christian Richardt$^2$ \quad
Michael Zollh{\"o}fer$^2$ \quad
Matthias Nie{\ss}ner$^1$ \\[0.5em]
$^1$Technical University of Munich \quad $^2$Meta \\
\url{https://lukashoel.github.io/ViewDiff/} \\[-0.8em]
}

\begin{document}
\begin{figure}

\twocolumn[{
\renewcommand\twocolumn[1][]{#1}
\maketitle

\centering
\setlength\tabcolsep{0pt}
\begin{tabular}{c@{}c}
\includegraphics[width=0.16976\textwidth]{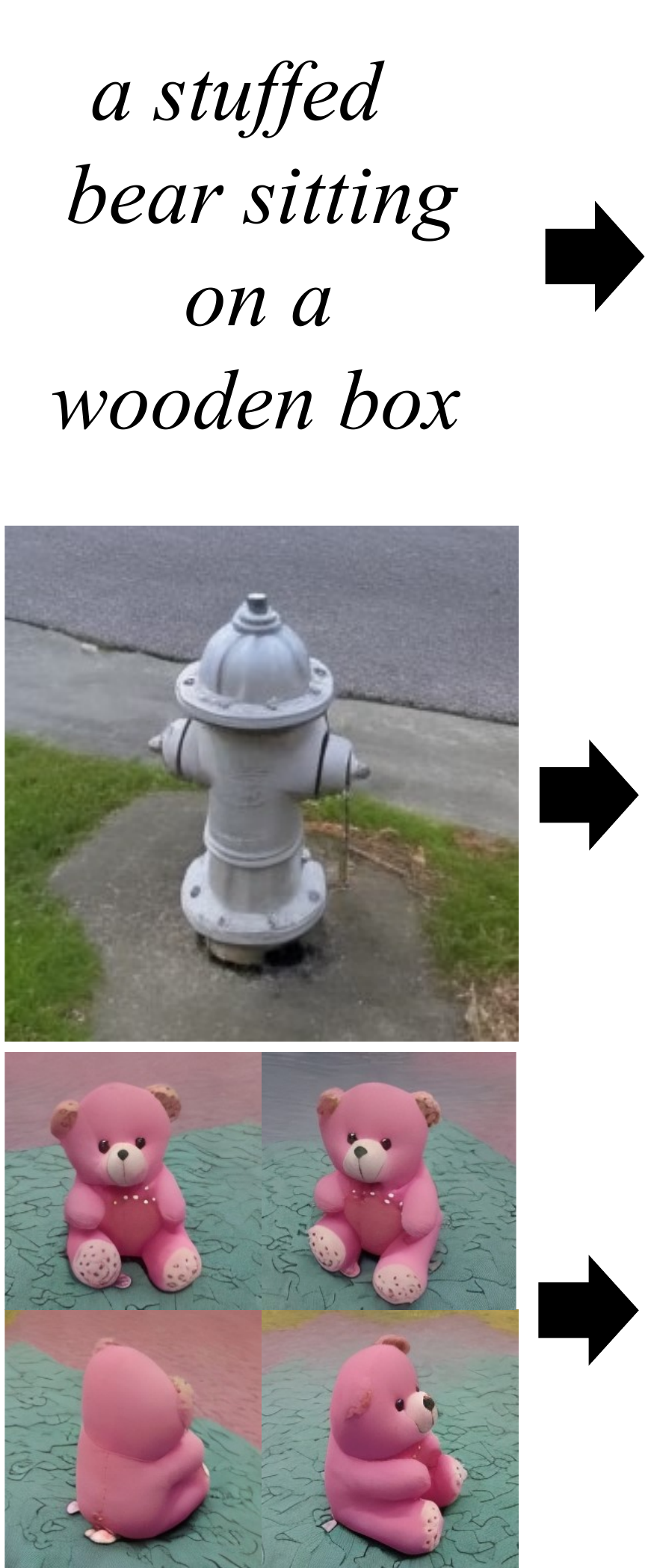} & 
\includegraphics[width=0.83023\textwidth]{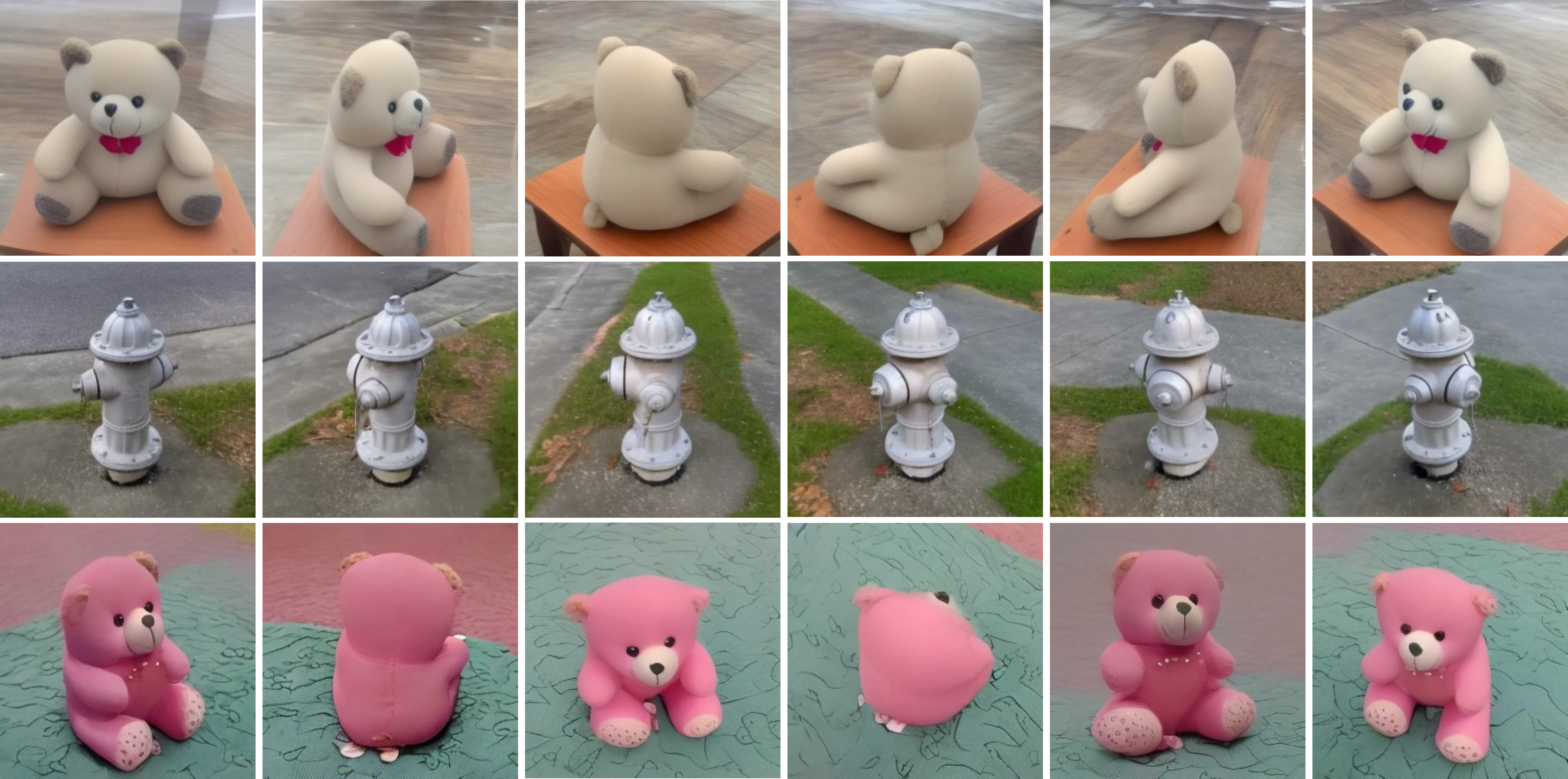} \\
Input & Multi-view generated images  \\
\end{tabular}
\vspace{-2mm}
\caption{
\textbf{Multi-view consistent image generation.}
Our method takes as input a text description, or any number of posed input images, and generates high-quality, multi-view consistent images of a real-world 3D object in authentic surroundings from any desired camera poses.
}
\vspace{2mm}
\label{fig:teaser}
}]
\end{figure}

\begin{abstract}
3D asset generation is getting massive amounts of attention, inspired by the recent success of text-guided 2D content creation.
Existing text-to-3D methods use pretrained text-to-image diffusion models in an optimization problem or fine-tune them on synthetic data, which often results in non-photorealistic 3D objects without backgrounds.
In this paper, we present a method that leverages pretrained text-to-image models as a prior, and learn to generate multi-view images in a single denoising process from real-world data.
Concretely, we propose to integrate 3D volume-rendering and cross-frame-attention layers into each block of the existing U-Net network of the text-to-image model.
Moreover, we design an autoregressive generation that renders more 3D-consistent images at any viewpoint.
We train our model on real-world datasets of objects and showcase its capabilities to generate instances with a variety of high-quality shapes and textures in authentic surroundings.
Compared to the existing methods, the results generated by our method are consistent, and have favorable visual quality ($-30\%$ FID, $-37\%$ KID).
\end{abstract}

\vspace{-4mm}
    
\section{Introduction}
\label{sec:intro}

In recent years, text-to-image (T2I) diffusion models~\cite{SaharCSLWDGAMLSHFN2022, RamesDNCC2022} %
have emerged as cutting-edge technologies, revolutionizing high-quality and imaginative 2D content creation guided by text descriptions.
These frameworks have found widespread applications, including extensions such as ControlNet~\cite{ZhangA2023} and DreamBooth~\cite{ruiz2023dreambooth}, showcasing their versatility and potential.
An intriguing direction in this domain is to use T2I models as powerful 2D priors for generating three-dimensional (3D) assets.
How can we effectively use these models to create photo-realistic and diverse 3D assets?

Existing methods like DreamFusion~\cite{PooleJBM2023}, Fantasia3D~\cite{chen2023fantasia}, and ProlificDreamer~\cite{WangLWBLSZ2023} have demonstrated exciting results by optimizing a 3D representation through score distillation sampling~\cite{PooleJBM2023} from pretrained T2I diffusion models.
The 3D assets generated by these methods exhibit compelling diversity.
However, their visual quality is not consistently as high as that of the images generated by T2I models.
A key step to obtaining 3D assets is the ability to generate consistent multi-view images of the desired objects and their surroundings.
These images can then be fitted to 3D representations like NeRF~\cite{mildenhall2021nerf} or NeuS~\cite{wang2021neus}.
HoloDiffusion~\cite{karnewar2023holodiffusion} and ViewsetDiffusion~\cite{szymanowicz23viewset_diffusion} train a diffusion model \emph{from scratch} using multi-view images and output 3D-consistent images.
GeNVS~\cite{ChanNCBPLAMKW2023} and DFM~\cite{tewari2023forwarddiffusion} additionally produce object surroundings, thereby increasing the realism of the generation.
These methods ensure (photo)-realistic results by training on real-world 3D datasets~\cite{ReizeSHSLN2021, zhou2018stereo}.
However, these datasets are orders of magnitude smaller than the 2D dataset %
used to train T2I diffusion models.
As a result, these approaches produce realistic but less-diverse 3D assets.
Alternatively, recent works like Zero-1-to-3~\cite{Liu_2023_ICCV} and One-2-3-45~\cite{liu2023one} leverage a pretrained T2I model and fine-tune it for 3D consistency.
These methods successfully preserve the diversity of generated results by training on a large synthetic 3D dataset~\cite{deitke2023objaverse}.
Nonetheless, the produced objects can be less photo-realistic and are without surroundings.

\begin{figure*}[t]
	\centering
	\includegraphics[width=\linewidth]{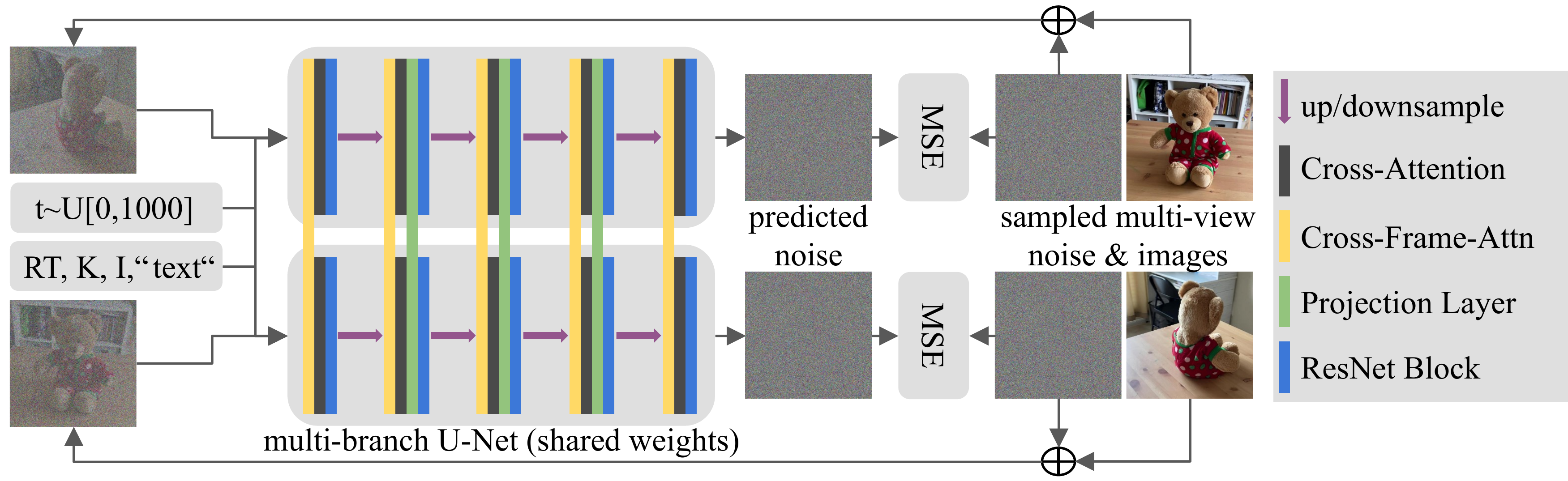}
	\caption{\textbf{Method Overview.}
		We augment the U-Net architecture of pretrained text-to-image models with new layers in every U-Net block.
		These layers facilitate communication between multi-view images in a batch, resulting in a denoising process that jointly produces 3D-consistent images.
		First, we replace self-attention with cross-frame-attention (yellow) which compares the spatial features of all views.
		We condition all attention layers on pose ($RT$), intrinsics ($K$), and intensity ($I$) of each image.
		Second, we add a projection layer (green) into the inner blocks of the U-Net.
		It creates a 3D representation from multi-view features and renders them into 3D-consistent features.
		We fine-tune the U-Net using the diffusion denoising objective~(\cref{eq:ddpm-eps-loss}) at timestep $t$, supervised from captioned multi-view images.
	}
	\label{fig:method}
\end{figure*}

In this paper, we propose a method that leverages the 2D priors of the pretrained T2I diffusion models to produce \emph{photo-realistic} and \emph{3D-consistent} 3D asset renderings.
As shown in the first two rows of \cref{fig:teaser}, the input is a text description or an image of an object, along with the camera poses of the desired rendered images.
The proposed approach produces multiple images of the same object in a single forward pass.
Moreover, we design an autoregressive generation scheme that allows to render more images at \emph{any} novel viewpoint (\cref{fig:teaser}, third row).
Concretely, we introduce projection and cross-frame-attention layers, that are strategically placed into the existing U-Net architecture, to encode explicit 3D knowledge about the generated object (see \cref{fig:method}).
By doing so, our approach paves the way to fine-tune T2I models on real-world 3D datasets, such as CO3D \cite{ReizeSHSLN2021}, while benefiting from the large 2D prior encoded in the pretrained weights.
Our generated images are consistent, diverse, and realistic renderings of objects.

\noindent To summarize, our contributions are:
\begin{itemize}
    \item a method that utilizes the pretrained 2D prior of text-to-image models and turns them into 3D-consistent image generators. We train our approach on real-world multi-view datasets, allowing us to produce realistic and high-quality images of objects and their surroundings (\cref{subsec:theory}).
    \item a novel U-Net architecture that combines commonly used 2D layers with 3D-aware layers. Our projection and cross-frame-attention layers encode explicit 3D knowledge into each block of the U-Net architecture (\cref{subsec:aug-layers}).
    \item an autoregressive generation scheme that renders images of a 3D object from \emph{any} desired viewpoint \emph{directly} with our model in a 3D-consistent way (\cref{subsec:autoreg-gen}).
\end{itemize}

\section{Related Work}
\label{sec:related-work}

\paragraph{Text-To-2D}
Denoising diffusion probabilistic models (DDPM) \cite{ho2020denoising} model a data distribution by learning to invert a Gaussian noising process with a deep network.
Recently, DDPMs were shown to be superior to generative adversarial networks \cite{dhariwal2021diffusion}, becoming the state-of-the-art framework for image generation.
Soon after, large text-conditioned models trained on billion-scale data were proposed in Imagen \cite{SaharCSLWDGAMLSHFN2022} or Dall-E 2 \cite{RamesDNCC2022}.
While \cite{dhariwal2021diffusion} achieved conditional generation via guidance with a classifier, \cite{ho2022classifier} proposed classifier-free guidance.
ControlNet \cite{ZhangA2023} proposed a way to tune the diffusion outputs by conditioning on various modalities, such as image segmentation or normal maps.
Similar to ControlNet, our method builds on the strong 2D prior of a pretrained text-to-image (T2I) model.
We further demonstrate how to adjust this prior to generate 3D-consistent images of objects.

\paragraph{Text-To-3D}
2D DDPMs were applied to the generation of 3D shapes~\cite{wang2023score,zhu2023hifa,tang2023makeit3d,xiang2023ivid,seo2023let,tsalicoglou2024textmesh,qian2023magic123} or scenes~\cite{Tang2023mvdiffusion,Hollein_2023_ICCV,instructnerf2023} from text descriptions.
DreamFusion \cite{PooleJBM2023} proposed score distillation sampling (SDS) which optimizes a 3D shape whose renders match the belief of the DDPM.
Improved sample quality was achieved by a second-stage mesh optimization \cite{lin2023magic3d, chen2023fantasia}, and smoother SDS convergence \cite{WangLWBLSZ2023, shi2023MVDream}.
Several methods use 3D data to train a novel-view synthesis model whose multi-view samples can be later converted to 3D,
e.g. conditioning
a 2D DDPM on an image and a relative camera motion to generate novel views \cite{watson2023novel,Liu_2023_ICCV}.
However, due to no explicit modelling of geometry, the outputs are view-inconsistent. %
Consistency can be improved with epipolar attention \cite{tseng2023consistent, zhou2023sparsefusion}, or optimizing a 3D shape from multi-view proposals \cite{liu2023one}.
Our work fine-tunes a 2D T2I model to generate renders of a 3D object; however, we propose explicit 3D unprojection and rendering operators to improve view-consistency.
Concurrently, SyncDreamer~\cite{liu2023syncdreamer} also add 3D layers in their 2D DDPM.
We differ by training on real data with backgrounds and by showing that autoregressive generation is sufficient to generate consistent images, making the second 3D reconstruction stage expendable. %

\paragraph{Diffusion on 3D Representations}
Several works model the distribution of 3D representations.
While DiffRF \cite{MuelleSPBKN2023} leverages ground-truth 3D shapes, HoloDiffusion \cite{karnewar2023holodiffusion} is supervised only with 2D images.
HoloFusion \cite{karnewar2023holofusion} extends
this work
with a 2D diffusion render post-processor.
Images can also be denoised by rendering a reconstructing 3D shape  \cite{szymanowicz23viewset_diffusion,anciukevivcius2023renderdiffusion}.
Unfortunately, the limited scale of existing 3D datasets prevents these 3D diffusion models from extrapolating beyond the training distribution.
Instead, we exploit a large 2D pretrained DDPM and add 3D components that are tuned on smaller-scale multi-view data.
This leads to improved multi-view consistency while maintaining the expressivity brought by pretraining on billion-scale image data.
\looseness-1

\section{Method}
\label{sec:method}

We propose a method that produces 3D-consistent images from a given text or posed image input (see \cref{fig:teaser} top/mid).
Concretely, given desired output poses, we jointly generate all images corresponding to the condition.
We leverage pretrained text-to-image (T2I) models~\cite{SaharCSLWDGAMLSHFN2022, RamesDNCC2022} %
and fine-tune them on multi-view data~\cite{ReizeSHSLN2021}.
We propose to augment the existing U-Net architecture by adding new layers into each block (see \cref{fig:method}).
At test time, we can condition our method on multiple images (see \cref{fig:teaser} bottom), which allows us to autoregressively render the same object from \emph{any} viewpoint \emph{directly} with the diffusion model (see \cref{subsec:autoreg-gen}).

\subsection{3D-Consistent Diffusion}
\label{subsec:theory}

Diffusion models \cite{sohl2015deep, ho2020denoising} are a class of generative models that learn the probability distribution $p_{\theta}(x_0) {=} \int p_{\theta}(x_{0{:}T})dx_{1{:}T}$ over data $x_0 {\sim} q(x_0)$ and latent variables $x_{1{:}T} {=} x_1, \ldots, x_T$.
Our method is based on pretrained text-to-image models, which are diffusion models $p_{\theta}(x_0 \mid c)$ with an additional text condition $c$.
For clarity, we drop the condition $c$ for the remainder of this section.

To produce multiple images $x_0^{0{:}N}$ at once, which are 3D-consistent with each other, we seek to model their joint probability distribution $p_{\theta}(x_0^{0{:}N}) {=} \int p_{\theta}(x_{0{:}T}^{0{:}N})dx_{1{:}T}^{0{:}N}$.
Similarly to concurrent work by \citet{liu2023syncdreamer}, we generate one set of images $p_{\theta}(x_0^{0{:}N})$ by adapting the \emph{reverse process} of DDPMs \cite{ho2020denoising} as a Markov chain over all images jointly:
\begin{align}
\label{eq:ddpm-reverse-process}
p_{\theta}(x_{0{:}T}^{0{:}N}) := p(x_T^{0{:}N}) \prod_{t=1}^T \prod_{n=0}^N p_{\theta}(x_{t-1}^n \mid x_t^{0{:}N}) \text{,}
\end{align}
where we start the generation from Gaussian noise sampled separately per image $p(x_T^n) = \mathcal{N}(x_T^n;\mathbf{0}, \mathbf{I})$, $\forall n \in [0, N]$.
We gradually denoise samples $p_{\theta}(x_{t-1}^n \mid x_t^{0{:}N}) = \mathcal{N}(x_{t-1}; \mu_{\theta}^n(x_t^{0{:}N}, t), \sigma_t^2\mathbf{I})$ by predicting the per-image mean $\mu_{\theta}^n(x_t^{0{:}N}, t)$ through a neural network $\mu_{\theta}$ that is shared between all images.
Importantly, at each step, the model uses the previous states $x_t^{0{:}N}$ of all images, i.e., there is communication between images during the model prediction.
We refer to \cref{subsec:aug-layers} for details on how this is implemented.
To train $\mu_{\theta}$, we define the \emph{forward process} as a Markov chain:
\begin{align}
\label{eq:ddpm-forward-process}
q(x_{1{:}T}^{0{:}N} \mid x_0^{0{:}N}) = \prod_{t=1}^T \prod_{n=0}^N q(x_t^n \mid x_{t-1}^n) \text{,}
\end{align}
where $q(x_t^n \mid x_{t-1}^n) = \mathcal{N}(x_t^n; \sqrt{1-\beta_t}x_{t-1}^n, \beta_t\mathbf{I})$ and $\beta_1, \ldots, \beta_T$ define a constant variance schedule, i.e., we apply separate noise per image to produce training samples.

We follow \citet{ho2020denoising} by learning a \emph{noise predictor} $\epsilon_{\theta}$ instead of $\mu_{\theta}$.
This allows to train $\epsilon_{\theta}$ with an $L2$ loss:
\begin{align}
\label{eq:ddpm-eps-loss}
\mathbb{E}_{x_0^{0{:}N}\!\!, \epsilon^{0{:}N} \sim \mathcal{N}(\mathbf{0}, \mathbf{I}), n}\left[\norm{ \epsilon^n - \epsilon_{\theta}^n(x_t^{0{:}N}, t) }^2\right] \text{.}
\end{align}

\subsection{Augmentation of the U-Net architecture}
\label{subsec:aug-layers}

To model a 3D-consistent denoising process over all images, we predict per-image noise $\epsilon_{\theta}^n(x_t^{0{:}N}, t)$ through a neural network $\epsilon_{\theta}$.
This neural network is initialized from the pretrained weights of existing text-to-image models, and is usually defined as a U-Net architecture \cite{SaharCSLWDGAMLSHFN2022, RamesDNCC2022}. %
We seek to leverage the previous states $x_t^{0{:}N}$ of all images to arrive at a 3D-consistent denoising step.
To this end, we propose to add two layers into the U-Net architecture, namely a cross-frame-attention layer and a projection layer.
We note that the predicted per-image noise needs to be image specific, since all images are generated starting from separate Gaussian noise.
It is therefore important to keep around 2D layers that act separately on each image, which we achieve by finetuning the existing ResNet \cite{HeZRS2016} and ViT \cite{DosovBKWZUDMHGUH2021} blocks.
We summarize our architecture in \cref{fig:method}.
In the following, we discuss our two proposed layers in more detail.

\paragraph{Cross-Frame Attention}

Inspired by video diffusion \cite{wu2022tune, yang2023rerender}, we add cross-frame-attention layers into the U-Net architecture.
Concretely, we modify the existing self-attention layers to calculate $\emph{CFAttn}(Q, K, V){=}\emph{softmax}\big(\frac{QK^T}{\sqrt{d}}\big)V$ with
\begin{align}
\label{eq:cfa}
Q = W^Qh_i \text{,~~}
K = W^K[h_j]_{j \neq i} \text{,~~}
V = W^V[h_j]_{j \neq i} \text{,}
\end{align}
where $W^Q, W^K, W^V$ are the pretrained weights for feature projection, and $h_i{\in} \mathbb{R}^{C{\times}H{\times}W}$ is the input spatial feature of each image $i{\in}[1,N]$.
Intuitively, this matches features across all frames, which allows generating the same global style.

Additionally, we add a conditioning vector to all cross-frame and cross-attention layers to inform the network about the viewpoint of each image.
First, we add pose information by encoding each image's camera matrix $p \in \mathbb{R}^{4 \times 4}$ into an embedding $z_1 \in \mathbb{R}^4$ similar to Zero-1-to-3 \cite{Liu_2023_ICCV}.
Additionally, we concatenate the focal length and principal point of each camera into an embedding $z_2 \in \mathbb{R}^4$.
Finally, we provide an intensity encoding $z_3 \in \mathbb{R}^2$, which stores the mean and variance of the image RGB values.
At training time, we set $z_3$ to the true values of each input image, and at test time, we set $z_3{=}[0.5,0]$ for all images.
This helps to reduce the view-dependent lighting differences contained in the dataset (e.g., due to different camera exposure).
We construct the conditioning vector as $z{=}[z_1,z_2,z_3]$, and add it through a LoRA-linear-layer \cite{hu2021lora} $W'^Q$ to the feature projection matrix $Q$.
Concretely, we compute the projected features as:
\begin{align}
\label{eq:cfa-lora}
Q = W^Qh_i + s \cdot W'^Q[h_i; z] \text{,}
\end{align}
where we set $s{=}1$.
Similarly, we add the condition via $W'^K$ to $K$, and $W'^V$ to $V$.

\paragraph{Projection Layer}

\begin{figure}
\centering
\includegraphics[width=\linewidth]{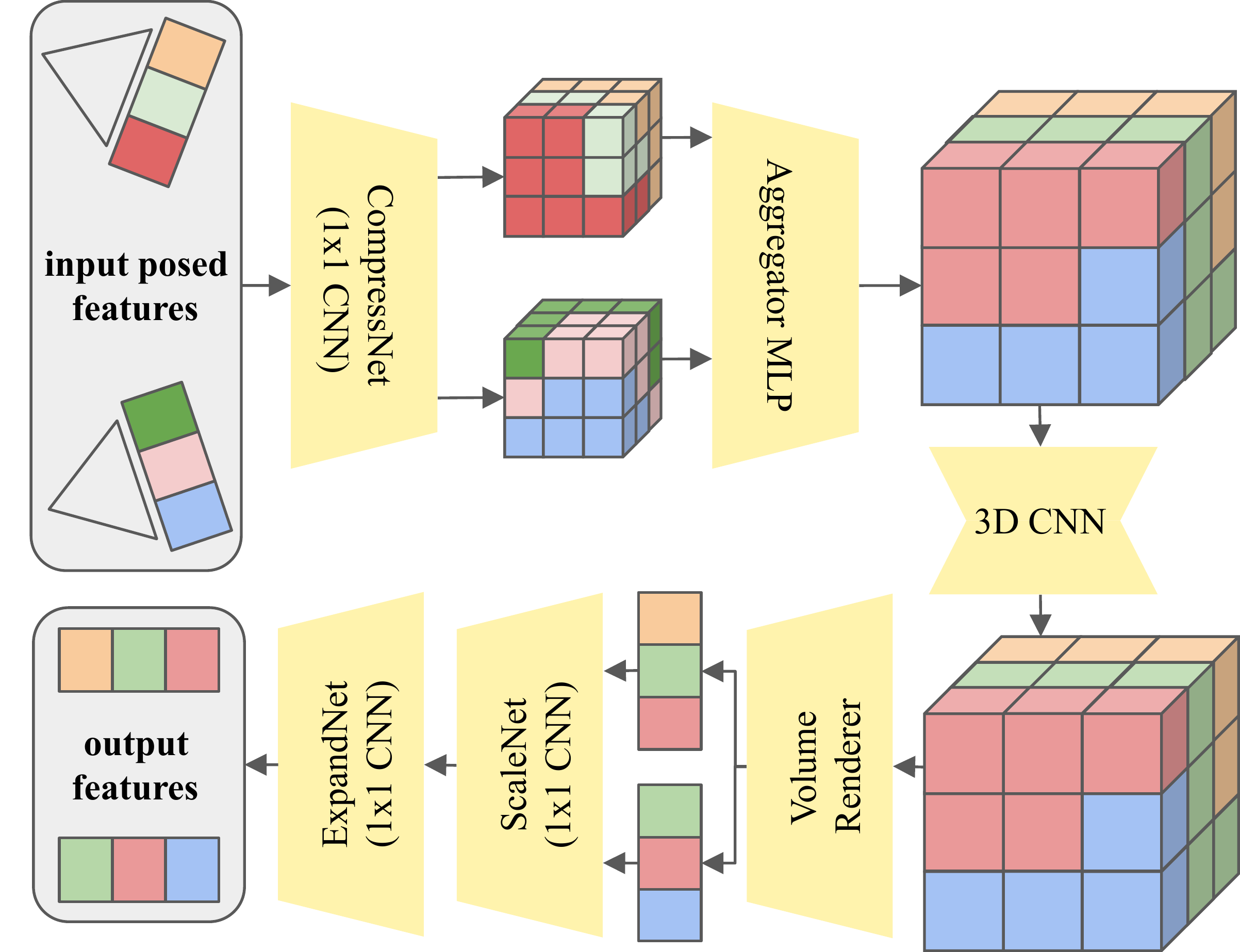}
\caption{\textbf{Architecture of the projection layer.}
We produce 3D-consistent output features from posed input features.
First, we unproject the compressed image features into 3D and aggregate them into a joint voxel grid with an MLP.
Then we refine the voxel grid with a 3D CNN.
A volume renderer similar to NeRF \cite{mildenhall2021nerf} renders 3D-consistent features from the grid.
Finally, we apply a learned scale function and expand the feature dimension.
}
\label{fig:proj-layer}
\end{figure}

Cross-frame attention layers are helpful to produce globally 3D-consistent images.
However, the objects do not precisely follow the specified poses, which leads to view-inconsistencies (see \cref{fig:uncond-360} and \cref{tab:ablation}).
To this end, we add a projection layer into the U-Net architecture (\cref{fig:proj-layer}).
The idea of this layer is to create 3D-consistent features that are then further processed by the next U-Net layers (e.g. ResNet blocks).
By repeating this layer across all stages of the U-Net, we ensure that the per-image features are in a 3D-consistent space.
We do not add the projection layer to the first and last U-Net blocks, as we saw no benefit from them at these locations.
We reason that the network processes image-specific information at those stages and thus does not need a 3D-consistent feature space.

Inspired by multi-view stereo literature \cite{izacard2022few, SunXCZB2021, BozicPTDN2021}, we create a 3D feature voxel grid from all input spatial features $h_\text{in}^{0:N} {\in} \mathbb{R}^{C{\times}H{\times}W}$ by projecting each voxel into each image plane.
First, we compress $h_\text{in}^{0:N}$ with a $1{\times}1$ convolution to a reduced feature dimension $C'{=}16$.
We then take the bilinearly interpolated feature at the image plane location and place it into the voxel.
This way, we create a separate voxel grid per view, and merge them into a single grid through an aggregator MLP.
Inspired by IBRNet \cite{WangWGSZBMSF2021}, the MLP predicts per-view weights followed by a weighted feature average.
We then run a small 3D CNN on the voxel grid to refine the 3D feature space.
Afterwards, we render the voxel grid into output features $h_\text{out}^{0:N} {\in} \mathbb{R}^{C'{\times}H{\times}W}$ with NeRF-like volumetric rendering \cite{mildenhall2021nerf} using \cite{cao2024lightplane}.
We dedicate half of the voxel grid to foreground and half to background and apply the background model from MERF~\cite{reiser2023merf} during ray-marching.

We found it is necessary to add a scale function after the volume rendering output.
The volume renderer typically uses a \emph{sigmoid} activation function as the final layer during ray-marching \cite{mildenhall2021nerf}.
However, the input features are defined in an arbitrary floating-point range.
To convert $h_\text{out}^{0:N}$ back into the same range, we non-linearly scale the features with 1$\times$1 convolutions and \emph{ReLU} activations.
Finally, we expand $h_\text{out}^{0:N}$ to the input feature dimension $C$.
We refer to the supplemental material for details about each component's architecture.

\subsection{Autoregressive Generation}
\label{subsec:autoreg-gen}

Our method takes as input multiple samples $x_t^{0{:}N}$ at once and denoises them 3D-consistently.
During training, we set $N{=}5$, but can increase it at inference time up to memory constraints, e.g., $N{=}30$.
However, we want to render an object from \emph{any} possible viewpoint directly with our network.
To this end, we propose an autoregressive image generation scheme, i.e., we condition the generation of next viewpoints on previously generated images.
We provide the timesteps $t^{0{:}N}$ of each image as input to the U-Net.
By varying $t^{0{:}N}$, we can achieve different types of conditioning.

\paragraph{Unconditional Generation}
All samples are initialized to Gaussian noise and are denoised jointly.
The timesteps $t^{0{:}N}$ are kept identical for all samples throughout the \emph{reverse process}.
We provide different cameras per image and a single text prompt.
The generated images are 3D-consistent, showing the object from the desired viewpoints (\cref{fig:uncond-generation,fig:uncond-360}).

\paragraph{Image-Conditional Generation}
We divide the total number of samples $N{=}n_\text{c}{+}n_\text{g}$ into a conditional part $n_\text{c}$ and generative part $n_\text{g}$.
The first $n_\text{c}$ samples correspond to images and cameras that are provided as input.
The other $n_\text{g}$ samples should generate novel views that are similar to the conditioning images.
We start the generation from Gaussian noise for the $n_\text{g}$ samples and provide the un-noised images for the other samples.
Similarly, we set $t^{0{:}n_\text{c}}{=}0$ for all denoising steps, while gradually decreasing $t^{n_\text{g}:N}$.

When $n_\text{c}{=}1$, our method performs single-image reconstruction (\cref{fig:cond-generation}).
Setting $n_\text{c}{>}1$ allows to autoregressively generate novel views from previous images (\cref{fig:teaser} bottom).
In practice, we first generate one batch of images unconditionally and then condition the next batches on a subset of previous images.
This allows us to render smooth trajectories around 3D objects (see the supplemental material).

\subsection{Implementation Details}
\label{subsec:impl-details}

\paragraph{Dataset}
We train our method on the large-scale CO3Dv2 \cite{ReizeSHSLN2021} dataset, which consists of posed multi-view images of real-world objects.
Concretely, we choose the categories \texttt{Teddybear}, \texttt{Hydrant}, \texttt{Apple}, and \texttt{Donut}.
Per category, we train on 500–1000 objects with each 200 images at resolution 256$\times$256.
We generate text captions with the BLIP-2 model \cite{li2023blip} and sample one of 5 proposals per object.

\paragraph{Training}
We base our model on a pretrained latent-diffusion %
text-to-image model.
We only fine-tune the U-Net and keep the VAE encoder and decoder frozen.
In each iteration, we select $N{=}5$ images and their poses.
We sample one denoising timestep $t {\sim} [0, 1000]$, add noise to the images according to \cref{eq:ddpm-forward-process}, and compute the loss according to \cref{eq:ddpm-eps-loss}.
In the projection layers, we skip the last image when building the voxel grid, which enforces to learn a 3D representation that can be rendered from novel views.
We train our method by varying between unconditional and image-conditional generation (\cref{subsec:autoreg-gen}).
Concretely, with probabilities $p_1{=}0.25$ and $p_2{=}0.25$ we provide the first and/or second image as input and set the respective timestep to zero.
Similar to \citet{ruiz2023dreambooth}, we create a \emph{prior dataset} with the pretrained text-to-image model and use it during training to maintain the 2D prior (see supplemental material for details).

We fine-tune the model on 2$\times$ A100 GPUs for 60K iterations (7 days) with a total batch size of 64.
We set the learning rate for the volume renderer to 0.005 and for all other layers to
$5{\times}10^{-5}$,
and use the AdamW optimizer \cite{SaharCSLWDGAMLSHFN2022}.
During inference, we can increase $N$ and generate up to 30 images/batch on an RTX 3090 GPU.
We use the UniPC~\cite{zhao2023unipc} sampler with 10 denoising steps, which takes 15 seconds.

\section{Results}
\label{sec:results}

\paragraph{Baselines}
We compare against recent state-of-the-art works for 3D generative modeling.
Our goal is to create multi-view consistent images from real-world, realistic objects with authentic surroundings.
Therefore, we consider methods that are trained on real-world datasets and select HoloFusion (HF) \cite{karnewar2023holofusion}, ViewsetDiffusion (VD) \cite{szymanowicz23viewset_diffusion}, and DFM \cite{tewari2023forwarddiffusion}.
We show results on two tasks: unconditional generation (\cref{subsec:res-uncond}) and single-image reconstruction (\cref{subsec:res-cond}).

\paragraph{Metrics}
We report FID~\cite{heusel2017gans} and KID~\cite{binkowski2018demystifying} as common metrics for 2D/3D generation and measure the multi-view consistency of generated images with peak signal-to-noise ratio (PSNR), structural similarity index (SSIM), and LPIPS~\cite{zhang2018unreasonable}.
To ensure comparability, we evaluate all metrics on images without backgrounds, as not every baseline models them.

\subsection{Unconditional Generation}
\label{subsec:res-uncond}

\begin{figure*}
\centering
\setlength\tabcolsep{1pt}
\def\arraystretch{0.7}%
\begin{tabular}{c c@{}c@{}c  c@{}c@{}c  c@{}c@{}c}
 &
\multicolumn{3}{c}{HoloFusion~(HF)~\cite{karnewar2023holofusion}} &
\multicolumn{3}{c}{ViewsetDiffusion~(VD)~\cite{szymanowicz23viewset_diffusion}} &
\multicolumn{3}{c}{Ours} \\ 
\multirow{3}{*}{\rotatebox{90}{Teddybear}} &
\includegraphics[width=0.11\textwidth]{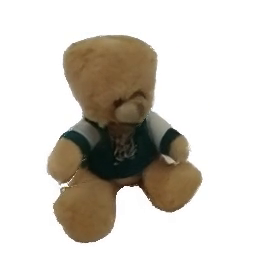} &
\includegraphics[width=0.11\textwidth]{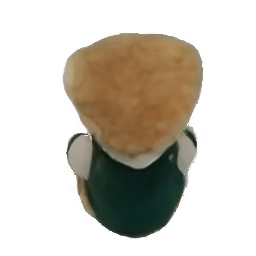} &
\includegraphics[width=0.11\textwidth]{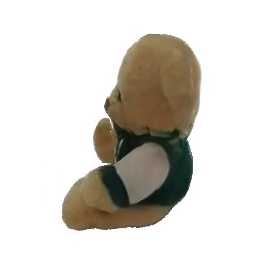} &
\includegraphics[width=0.11\textwidth]{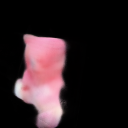} &
\includegraphics[width=0.11\textwidth]{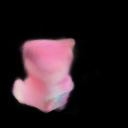} &
\includegraphics[width=0.11\textwidth]{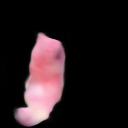} &
\includegraphics[width=0.11\textwidth]{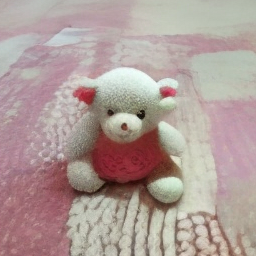} &
\includegraphics[width=0.11\textwidth]{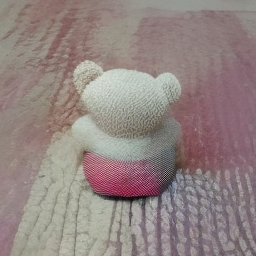} &
\includegraphics[width=0.11\textwidth]{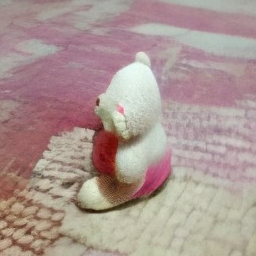} \\ 
&
\includegraphics[width=0.11\textwidth]{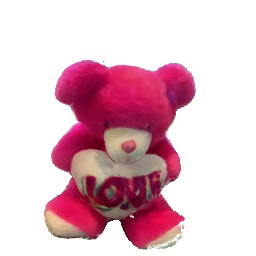} &
\includegraphics[width=0.11\textwidth]{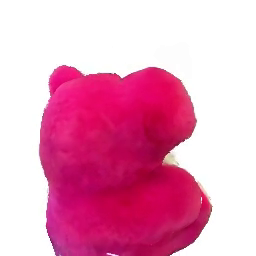} &
\includegraphics[width=0.11\textwidth]{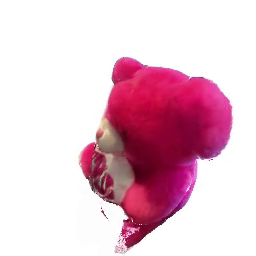} &
\includegraphics[width=0.11\textwidth]{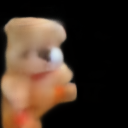} &
\includegraphics[width=0.11\textwidth]{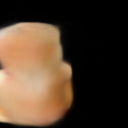} &
\includegraphics[width=0.11\textwidth]{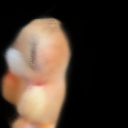} &
\includegraphics[width=0.11\textwidth]{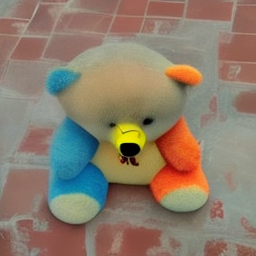} &
\includegraphics[width=0.11\textwidth]{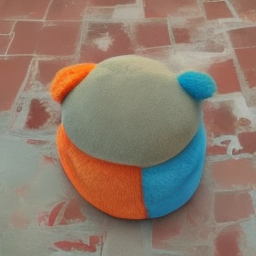} &
\includegraphics[width=0.11\textwidth]{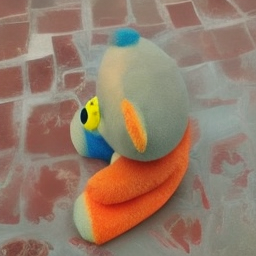} \\ 
\hline \\[-1.5ex]
\multirow{3}{*}{\rotatebox{90}{Hydrant}} &
\includegraphics[width=0.11\textwidth]{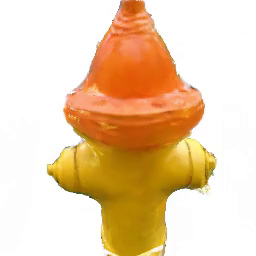} &
\includegraphics[width=0.11\textwidth]{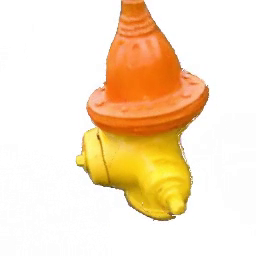} &
\includegraphics[width=0.11\textwidth]{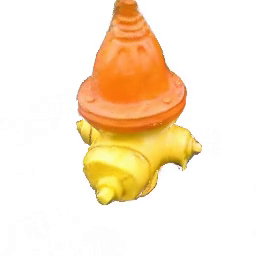} &
\includegraphics[width=0.11\textwidth]{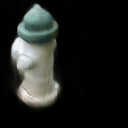} &
\includegraphics[width=0.11\textwidth]{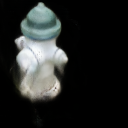} &
\includegraphics[width=0.11\textwidth]{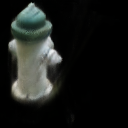} &
\includegraphics[width=0.11\textwidth]{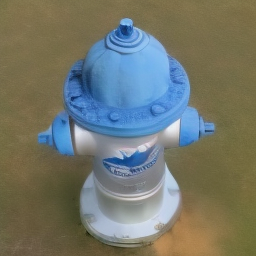} &
\includegraphics[width=0.11\textwidth]{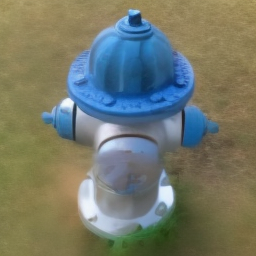} &
\includegraphics[width=0.11\textwidth]{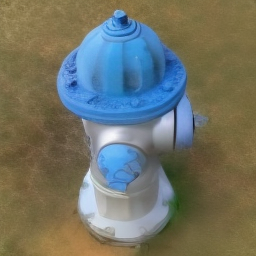} \\
 &
\includegraphics[width=0.11\textwidth]{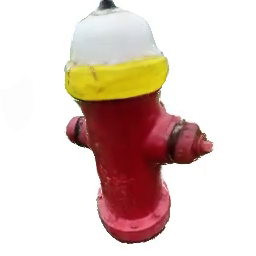} &
\includegraphics[width=0.11\textwidth]{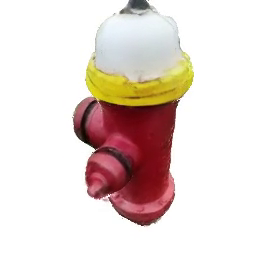} &
\includegraphics[width=0.11\textwidth]{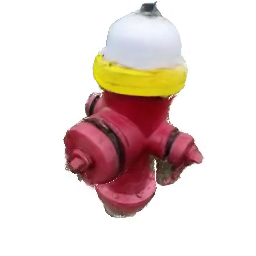} &
\includegraphics[width=0.11\textwidth]{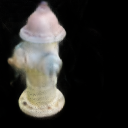} &
\includegraphics[width=0.11\textwidth]{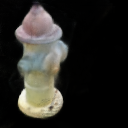} &
\includegraphics[width=0.11\textwidth]{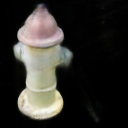} &
\includegraphics[width=0.11\textwidth]{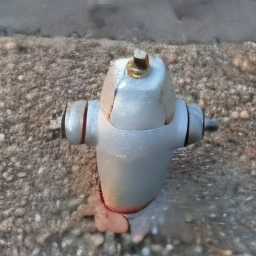} &
\includegraphics[width=0.11\textwidth]{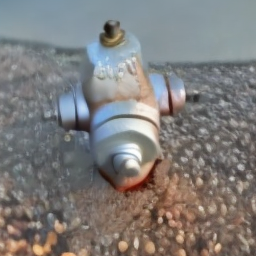} &
\includegraphics[width=0.11\textwidth]{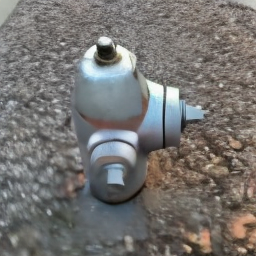} \\
\hline \\[-1.5ex]
\multirow{3}{*}{\rotatebox{90}{Apple}} &
\includegraphics[width=0.11\textwidth]{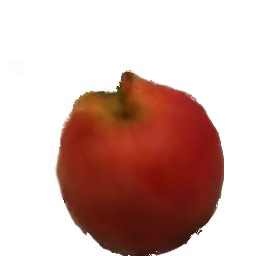} &
\includegraphics[width=0.11\textwidth]{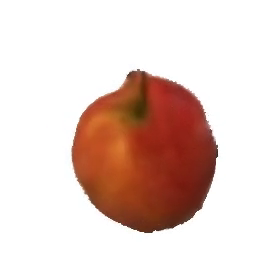} &
\includegraphics[width=0.11\textwidth]{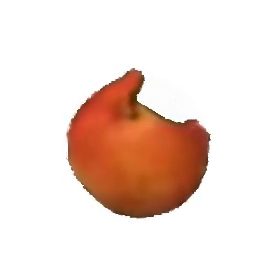} &
\includegraphics[width=0.11\textwidth]{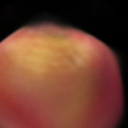} &
\includegraphics[width=0.11\textwidth]{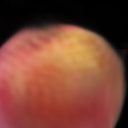} &
\includegraphics[width=0.11\textwidth]{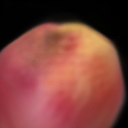} &
\includegraphics[width=0.11\textwidth]{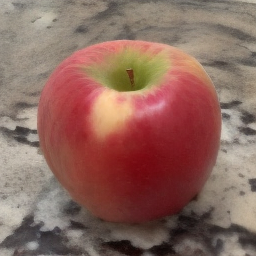} &
\includegraphics[width=0.11\textwidth]{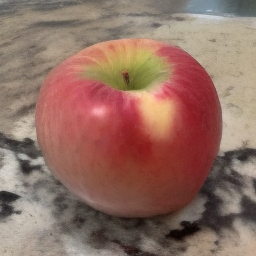} &
\includegraphics[width=0.11\textwidth]{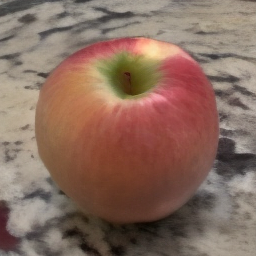} \\
&
\includegraphics[width=0.11\textwidth]{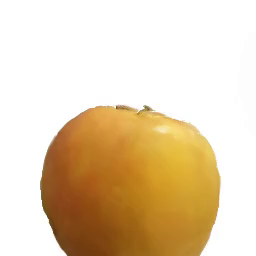} &
\includegraphics[width=0.11\textwidth]{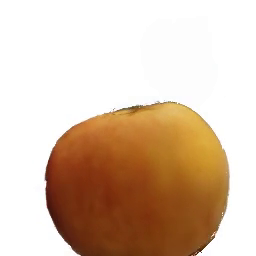} &
\includegraphics[width=0.11\textwidth]{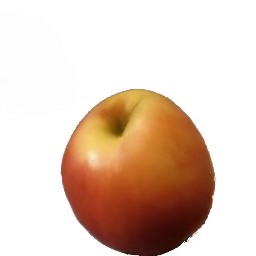} &
\includegraphics[width=0.11\textwidth]{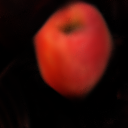} &
\includegraphics[width=0.11\textwidth]{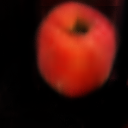} &
\includegraphics[width=0.11\textwidth]{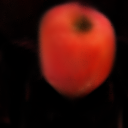} &
\includegraphics[width=0.11\textwidth]{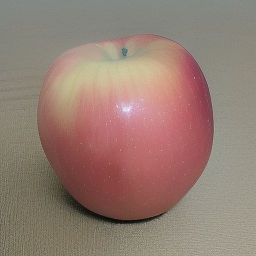} &
\includegraphics[width=0.11\textwidth]{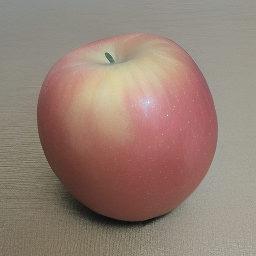} &
\includegraphics[width=0.11\textwidth]{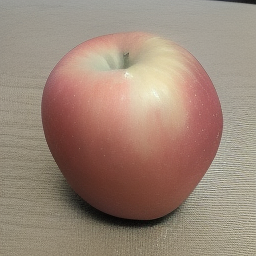} \\
\hline \\[-1.5ex]
\multirow{3}{*}{\rotatebox{90}{Donut}} &
\includegraphics[width=0.11\textwidth]{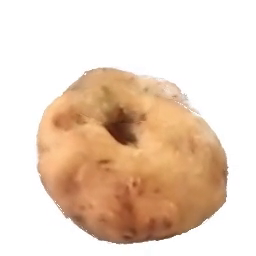} &
\includegraphics[width=0.11\textwidth]{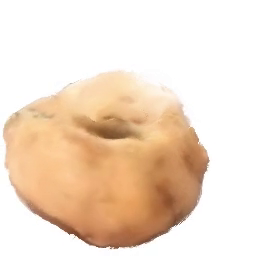} &
\includegraphics[width=0.11\textwidth]{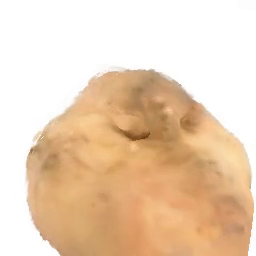} &
\includegraphics[width=0.11\textwidth]{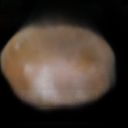} &
\includegraphics[width=0.11\textwidth]{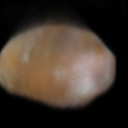} &
\includegraphics[width=0.11\textwidth]{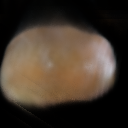} &
\includegraphics[width=0.11\textwidth]{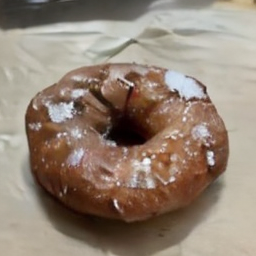} &
\includegraphics[width=0.11\textwidth]{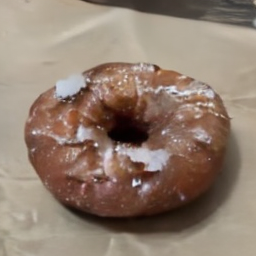} &
\includegraphics[width=0.11\textwidth]{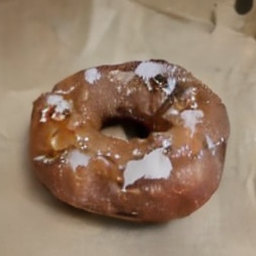} \\
&
\includegraphics[width=0.11\textwidth]{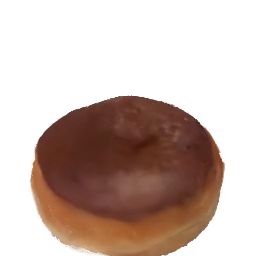} &
\includegraphics[width=0.11\textwidth]{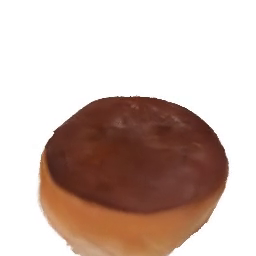} &
\includegraphics[width=0.11\textwidth]{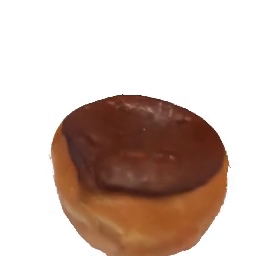} &
\includegraphics[width=0.11\textwidth]{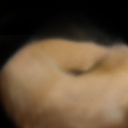} &
\includegraphics[width=0.11\textwidth]{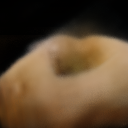} &
\includegraphics[width=0.11\textwidth]{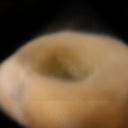} &
\includegraphics[width=0.11\textwidth]{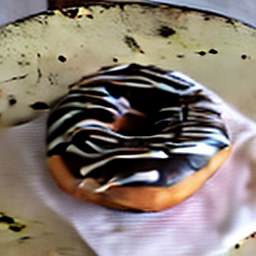} &
\includegraphics[width=0.11\textwidth]{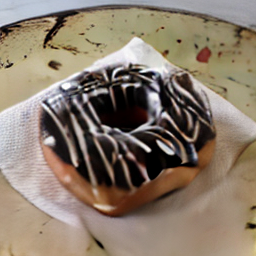} &
\includegraphics[width=0.11\textwidth]{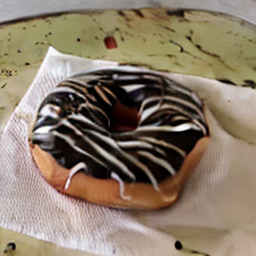} \\

\end{tabular}
\caption{
\textbf{Unconditional image generation of our method and baselines.}
We show renderings from different viewpoints for multiple objects and categories.
Our method produces consistent objects and backgrounds.
Our textures are sharper in comparison to baselines.
Please see the supplemental material for more examples and animations.
}
\label{fig:uncond-generation}
\end{figure*}

\begin{table}
  \caption{
	\textbf{Quantitative comparison of unconditional image generation.}
	We report average FID~\cite{heusel2017gans} and KID~\cite{binkowski2018demystifying} per category and improve by a significant margin.
	This signals that our images are more similar to the distribution of real images in the dataset.
	We mask away the background for our method and the real images to ensure comparability of numbers with the baselines.
}
  \centering
  \setlength\tabcolsep{3pt}
  \begin{tabular}{l rr rr rr}
    \toprule
        \multirow{2}{*}{Category} & \multicolumn{2}{c}{HF~\cite{karnewar2023holofusion}} & \multicolumn{2}{c}{VD~\cite{szymanowicz23viewset_diffusion}} & \multicolumn{2}{c}{Ours} \\
        \cmidrule(l{2pt}r{2pt}){2-3} \cmidrule(l{2pt}r{2pt}){4-5} \cmidrule(l{2pt}r{2pt}){6-7}
    & FID$\downarrow$ & KID$\downarrow$ & FID$\downarrow$ & KID$\downarrow$ & FID$\downarrow$ & KID$\downarrow$ \\
    \midrule
    Teddybear & 81.93 & 0.072 & 201.71 & 0.169 & \textbf{49.39} & \textbf{0.036} \\
    Hydrant   & 61.19 & 0.042 & 138.45 & 0.118 & \textbf{46.45} & \textbf{0.033} \\
    Donut     & 105.97 & 0.091 & 199.14 & 0.136 & \textbf{68.86} & \textbf{0.054} \\
    Apple     & 62.19 & 0.056 & 183.67 & 0.149 & \textbf{56.85} & \textbf{0.043} \\
    \bottomrule
  \end{tabular}
  \label{tab:fid}
\end{table}

\begin{figure*}
\centering
\setlength\tabcolsep{0pt}
\renewcommand\cellset{\renewcommand\arraystretch{0}%
\setlength\extrarowheight{0pt}}
\begin{tabular}{c ccccccccc}
 &
0\textdegree \tikzmark{a}&
 &
 &
 &
 &
 &
 &
 &
\tikzmark{b} 340\textdegree \\ 

\rotatebox[origin=c]{90}{\scalebox{1.0}{HF~\cite{karnewar2023holofusion}}} &
\raisebox{-0.5\height}{\includegraphics[width=0.11\textwidth]{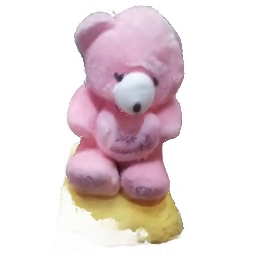}} &
\raisebox{-0.5\height}{\includegraphics[width=0.11\textwidth]{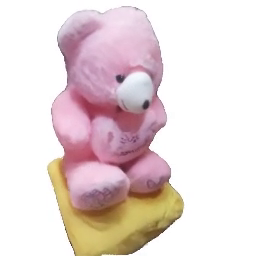}} &
\raisebox{-0.5\height}{\includegraphics[width=0.11\textwidth]{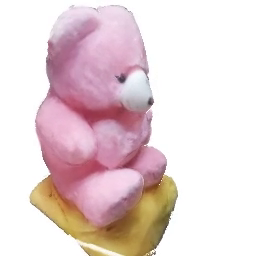}} &
\raisebox{-0.5\height}{\includegraphics[width=0.11\textwidth]{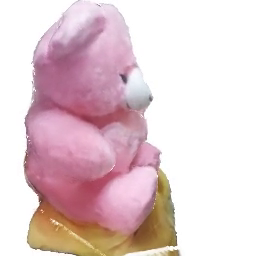}} &
\raisebox{-0.5\height}{\includegraphics[width=0.11\textwidth]{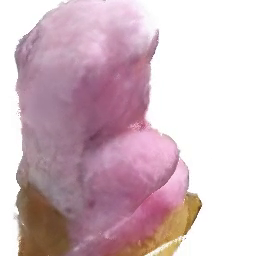}} &
\raisebox{-0.5\height}{\includegraphics[width=0.11\textwidth]{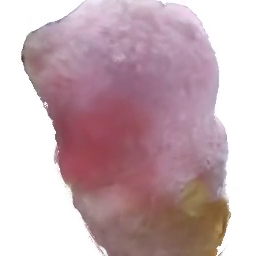}} &
\raisebox{-0.5\height}{\includegraphics[width=0.11\textwidth]{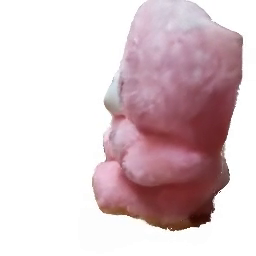}} &
\raisebox{-0.5\height}{\includegraphics[width=0.11\textwidth]{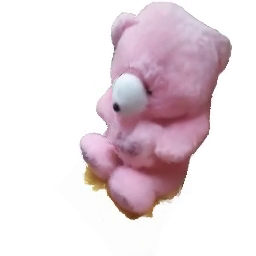}} &
\raisebox{-0.5\height}{\includegraphics[width=0.11\textwidth]{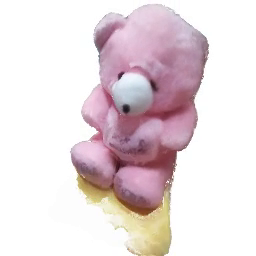}} \\

\rotatebox[origin=c]{90}{\scalebox{1.0}{VD~\cite{szymanowicz23viewset_diffusion}}} &
\raisebox{-0.5\height}{\includegraphics[width=0.11\textwidth]{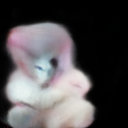}} &
\raisebox{-0.5\height}{\includegraphics[width=0.11\textwidth]{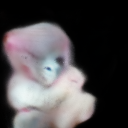}} &
\raisebox{-0.5\height}{\includegraphics[width=0.11\textwidth]{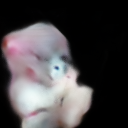}} &
\raisebox{-0.5\height}{\includegraphics[width=0.11\textwidth]{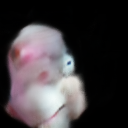}} &
\raisebox{-0.5\height}{\includegraphics[width=0.11\textwidth]{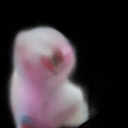}} &
\raisebox{-0.5\height}{\includegraphics[width=0.11\textwidth]{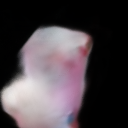}} &
\raisebox{-0.5\height}{\includegraphics[width=0.11\textwidth]{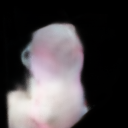}} &
\raisebox{-0.5\height}{\includegraphics[width=0.11\textwidth]{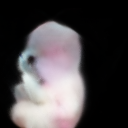}} &
\raisebox{-0.5\height}{\includegraphics[width=0.11\textwidth]{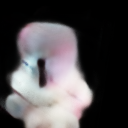}} \\

\rotatebox[origin=c]{90}{\scalebox{1.0}{\makecell{Ours \\ (no proj)}}} &
\raisebox{-0.5\height}{\includegraphics[width=0.11\textwidth]{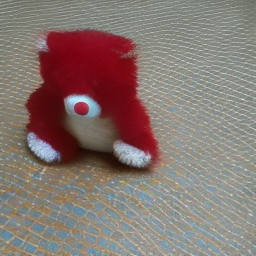}} &
\raisebox{-0.5\height}{\includegraphics[width=0.11\textwidth]{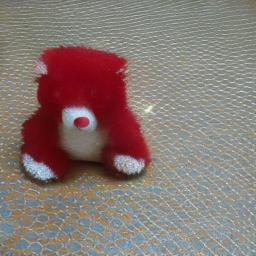}} &
\raisebox{-0.5\height}{\includegraphics[width=0.11\textwidth]{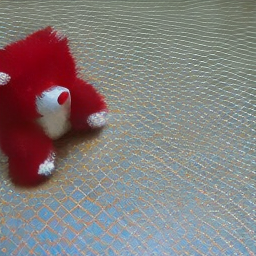}} &
\raisebox{-0.5\height}{\includegraphics[width=0.11\textwidth]{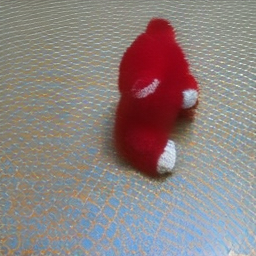}} &
\raisebox{-0.5\height}{\includegraphics[width=0.11\textwidth]{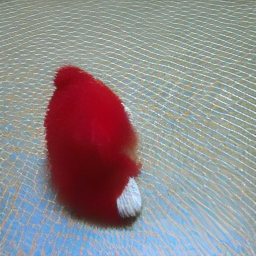}} &
\raisebox{-0.5\height}{\includegraphics[width=0.11\textwidth]{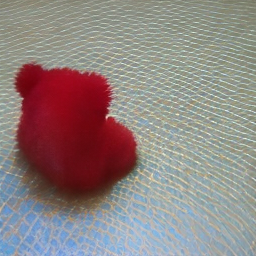}} &
\raisebox{-0.5\height}{\includegraphics[width=0.11\textwidth]{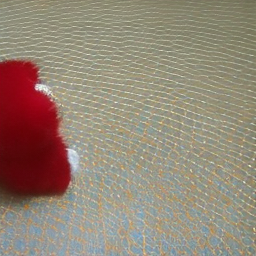}} &
\raisebox{-0.5\height}{\includegraphics[width=0.11\textwidth]{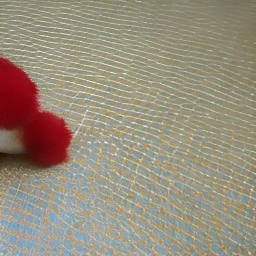}} &
\raisebox{-0.5\height}{\includegraphics[width=0.11\textwidth]{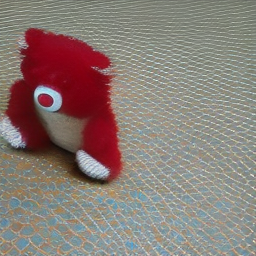}} \\

\rotatebox[origin=c]{90}{\scalebox{1.0}{\makecell{Ours \\ (no cfa)}}} &
\raisebox{-0.5\height}{\includegraphics[width=0.11\textwidth]{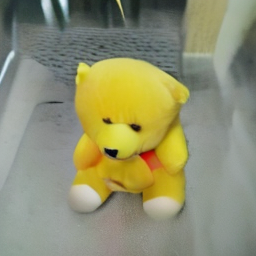}} &
\raisebox{-0.5\height}{\includegraphics[width=0.11\textwidth]{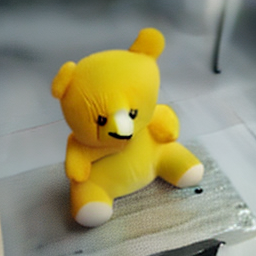}} &
\raisebox{-0.5\height}{\includegraphics[width=0.11\textwidth]{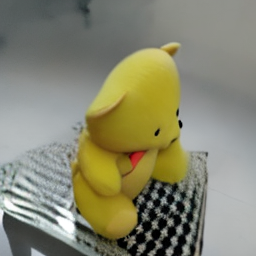}} &
\raisebox{-0.5\height}{\includegraphics[width=0.11\textwidth]{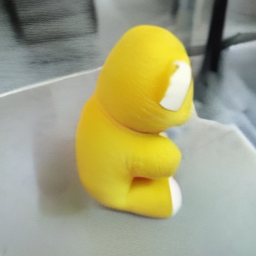}} &
\raisebox{-0.5\height}{\includegraphics[width=0.11\textwidth]{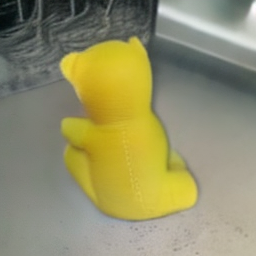}} &
\raisebox{-0.5\height}{\includegraphics[width=0.11\textwidth]{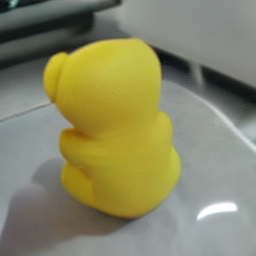}} &
\raisebox{-0.5\height}{\includegraphics[width=0.11\textwidth]{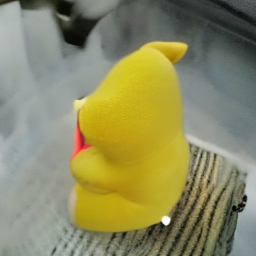}} &
\raisebox{-0.5\height}{\includegraphics[width=0.11\textwidth]{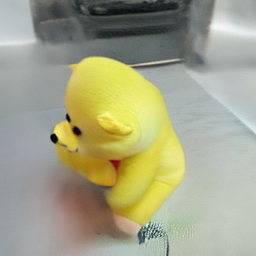}} &
\raisebox{-0.5\height}{\includegraphics[width=0.11\textwidth]{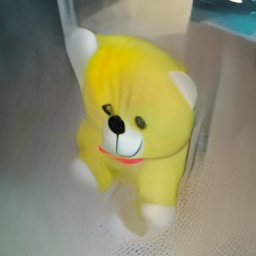}} \\

\rotatebox[origin=c]{90}{\scalebox{1.0}{Ours}} &
\raisebox{-0.5\height}{\includegraphics[width=0.11\textwidth]{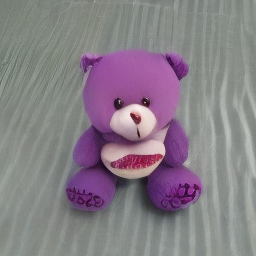}} &
\raisebox{-0.5\height}{\includegraphics[width=0.11\textwidth]{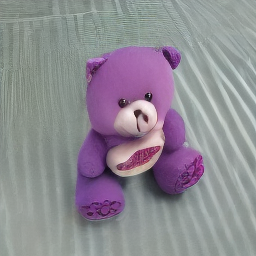}} &
\raisebox{-0.5\height}{\includegraphics[width=0.11\textwidth]{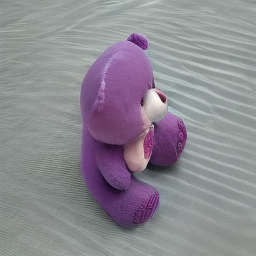}} &
\raisebox{-0.5\height}{\includegraphics[width=0.11\textwidth]{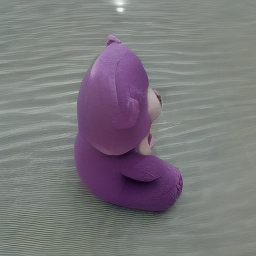}} &
\raisebox{-0.5\height}{\includegraphics[width=0.11\textwidth]{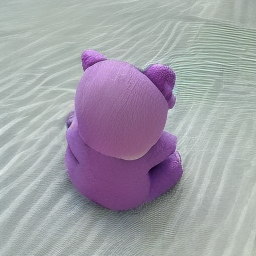}} &
\raisebox{-0.5\height}{\includegraphics[width=0.11\textwidth]{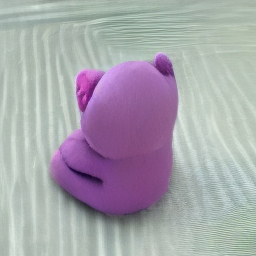}} &
\raisebox{-0.5\height}{\includegraphics[width=0.11\textwidth]{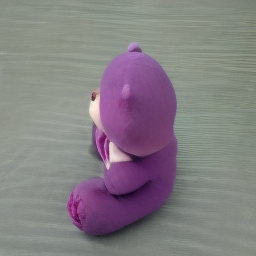}} &
\raisebox{-0.5\height}{\includegraphics[width=0.11\textwidth]{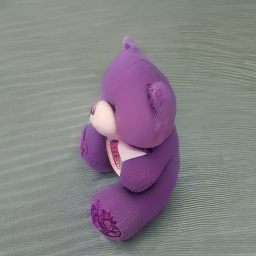}} &
\raisebox{-0.5\height}{\includegraphics[width=0.11\textwidth]{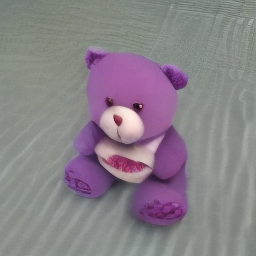}} \\

\end{tabular}
\begin{tikzpicture}[overlay,remember picture]
\draw[->,black,thick] ([yshift=0.12cm]pic cs:a) -- ([yshift=0.12cm]pic cs:b);
\end{tikzpicture}
\caption{
\textbf{Multi-view consistency of unconditional image generation.}
HoloFusion (HF) \cite{karnewar2023holofusion} has view-dependent floating artifacts (the base in first row).
ViewsetDiffusion (VD) \cite{szymanowicz23viewset_diffusion} has blurrier renderings (second row).
Without the projection layer, our method has no precise control over viewpoints (third row).
Without cross-frame-attention, our method suffers from identity changes of the object (fourth row).
Our full method produces detailed images that are 3D-consistent (fifth row).
}
\label{fig:uncond-360}
\end{figure*}

Our method can be used to generate 3D-consistent views of an object from any pose with only text as input by using our autoregressive generation (\cref{subsec:autoreg-gen}).
Concretely, we sample an (unobserved) image caption from the test set for the first batch and generate $N{=}10$ images with a guidance scale~\cite{ho2022classifier} of $\lambda_\text{cfg}{=}7.5$.
We then set $\lambda_\text{cfg}{=}{0}$ for subsequent batches, and create a total of 100 images per object.

We evaluate against HoloFusion~(HF)~\cite{karnewar2023holofusion} and ViewsetDiffusion~(VD)~\cite{szymanowicz23viewset_diffusion}.
We report quantitative results in \cref{tab:fid} and qualitative results in \cref{fig:uncond-generation,fig:uncond-360}.
HF~\cite{karnewar2023holofusion} creates diverse images that sometimes show view-dependent floating artifacts (see \cref{fig:uncond-360}).
VD~\cite{szymanowicz23viewset_diffusion} creates consistent but blurry images.
In contrast, our method produces images with backgrounds and higher-resolution object details.
Please see the suppl. material for more examples and animated results.

\subsection{Single-Image Reconstruction}
\label{subsec:res-cond}

\begin{figure}
\centering
\setlength\tabcolsep{0pt}
\def\arraystretch{0.7}%
\begin{tabular}{ccccc}
Input &
VD~\cite{szymanowicz23viewset_diffusion} &
DFM~\cite{tewari2023forwarddiffusion} &
Ours &
Real Image \\
\includegraphics[width=0.1\textwidth]{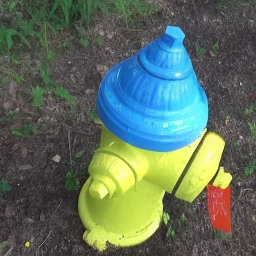} &
\includegraphics[width=0.1\textwidth]{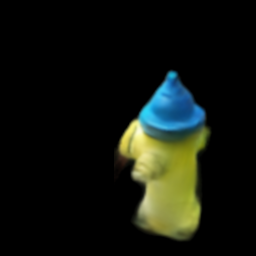} &
\includegraphics[width=0.1\textwidth]{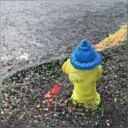} &
\includegraphics[width=0.1\textwidth]{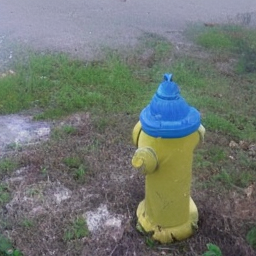} & 
\includegraphics[width=0.1\textwidth]{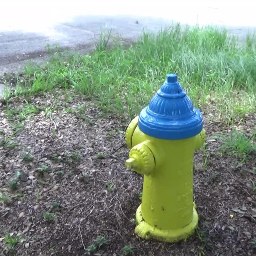} \\
\includegraphics[width=0.1\textwidth]{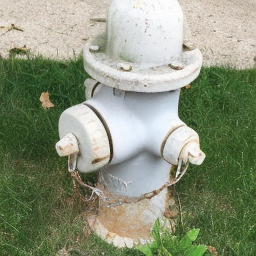} &
\includegraphics[width=0.1\textwidth]{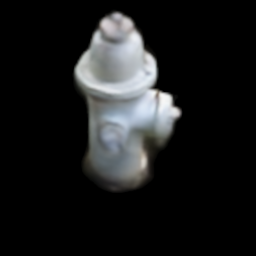} &
\includegraphics[width=0.1\textwidth]{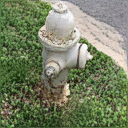} &
\includegraphics[width=0.1\textwidth]{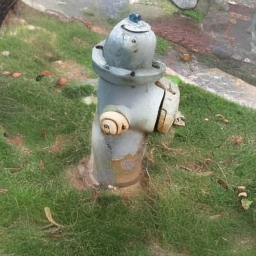} & 
\includegraphics[width=0.1\textwidth]{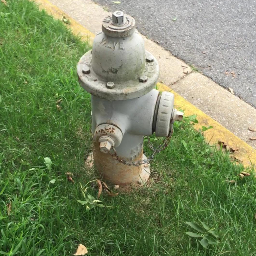} \\

\end{tabular}
\caption{
\textbf{Single-image reconstruction of our method and baselines.}
Given one image/pose as input, our method produces plausible novel views that are consistent with the real shape and texture.
We can also produce detailed backgrounds that match the input.
}
\label{fig:cond-generation}
\end{figure}

\begin{table*}
	\caption{
		\textbf{Quantitative comparison of single-image reconstruction.}
		Given a single image as input, we measure the quality of novel views through average PSNR, SSIM, and LPIPS~\cite{zhang2018unreasonable} per category.
		We mask away the generated backgrounds to ensure comparability across all methods.
		We improve over VD~\cite{szymanowicz23viewset_diffusion} while being on-par with DFM~\cite{tewari2023forwarddiffusion}. 
	}
  \centering
  \setlength\tabcolsep{3pt}
  \begin{tabular}{l ccc ccc ccc ccc}
    \toprule
        \multirow{2}{*}{Method} & \multicolumn{3}{c}{Teddybear} & \multicolumn{3}{c}{Hydrant} & \multicolumn{3}{c}{Donut} & \multicolumn{3}{c}{Apple} \\
                        \cmidrule(l{2pt}r{2pt}){2-4} \cmidrule(l{2pt}r{2pt}){5-7} \cmidrule(l{2pt}r{2pt}){8-10} \cmidrule(l{2pt}r{2pt}){11-13}
    & PSNR$\uparrow$ & SSIM$\uparrow$ & LPIPS$\downarrow$ & PSNR$\uparrow$ & SSIM$\uparrow$ & LPIPS$\downarrow$ & PSNR$\uparrow$ & SSIM$\uparrow$ & LPIPS$\downarrow$ & PSNR$\uparrow$ & SSIM$\uparrow$ & LPIPS$\downarrow$ \\
    \midrule
    VD~\cite{szymanowicz23viewset_diffusion} & 19.68 & 0.70 & 0.30 & 22.36 & 0.80 & 0.19 & 18.27 & 0.68 & 0.14 & 19.54 & 0.64 & 0.31 \\
    DFM~\cite{tewari2023forwarddiffusion} & 21.81 & 0.82 & 0.16 & \textbf{22.67} & 0.83 & 0.12 & \textbf{23.91} & \textbf{0.86} & \textbf{0.10} & 25.79 & \textbf{0.91} & \textbf{0.07} \\
    Ours & \textbf{21.98} & \textbf{0.84} & \textbf{0.13} & 22.49 & \textbf{0.85} & \textbf{0.11} & 21.50 & 0.85 & 0.18 & \textbf{25.94} & \textbf{0.91} & 0.11 \\
    \bottomrule
  \end{tabular}
  \label{tab:cond-generation}
\end{table*}

Our method can be conditioned on multiple images in order to render any novel view in an autoregressive fashion (\cref{subsec:autoreg-gen}).
To measure the 3D-consistency of our generated images, we compare single-image reconstruction against ViewsetDiffusion~(VD)~\cite{szymanowicz23viewset_diffusion} and DFM~\cite{tewari2023forwarddiffusion}.
Concretely, we sample one image from the dataset and generate 20 images at novel views also sampled from the dataset.
We follow \citet{szymanowicz23viewset_diffusion} and report the per-view maximum PSNR/SSIM and average LPIPS across multiple objects and viewpoints for all methods.
We report quantitative results in \cref{tab:cond-generation} and show qualitative results in \cref{fig:cond-generation}.
VD~\cite{szymanowicz23viewset_diffusion} creates plausible results without backgrounds.
DFM~\cite{tewari2023forwarddiffusion} creates consistent results with backgrounds at a lower image resolution (128$\times$128).
Our method produces higher resolution images with similar reconstruction results and backgrounds as DFM~\cite{tewari2023forwarddiffusion}.
Please see the supplemental material for more examples and animated results.

\subsection{Ablations}
\label{subsec:ablation}

\begin{figure}
\centering
\setlength\tabcolsep{0.5pt}
\begin{tabular}{c c@{}c c@{}c}

\rotatebox[origin=c]{90}{\scalebox{1.0}{sample 1}} &
\raisebox{-0.5\height}{\includegraphics[width=0.12\textwidth]{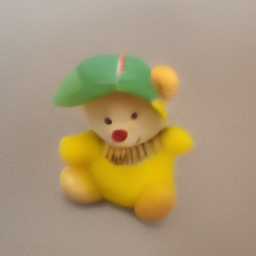}} & 
\raisebox{-0.5\height}{\includegraphics[width=0.12\textwidth]{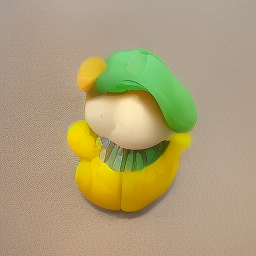}} & 

\raisebox{-0.5\height}{\includegraphics[width=0.12\textwidth]{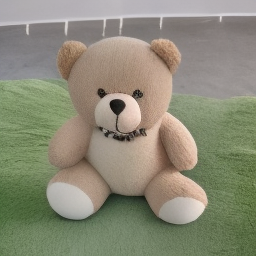}} & 
\raisebox{-0.5\height}{\includegraphics[width=0.12\textwidth]{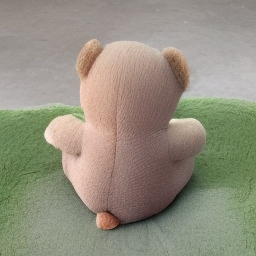}} \\

\rotatebox[origin=c]{90}{\scalebox{1.0}{sample 2}} &
\raisebox{-0.5\height}{\includegraphics[width=0.12\textwidth]{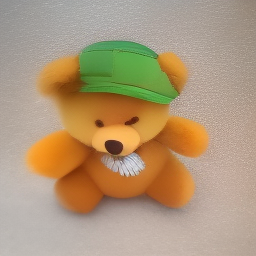}} & 
\raisebox{-0.5\height}{\includegraphics[width=0.12\textwidth]{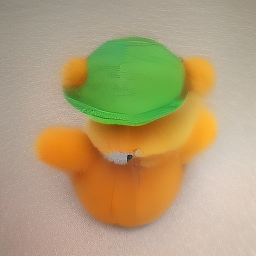}} & 

\raisebox{-0.5\height}{\includegraphics[width=0.12\textwidth]{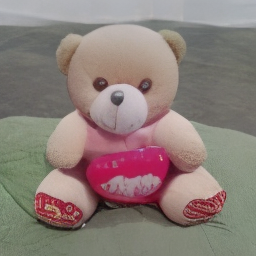}} & 
\raisebox{-0.5\height}{\includegraphics[width=0.12\textwidth]{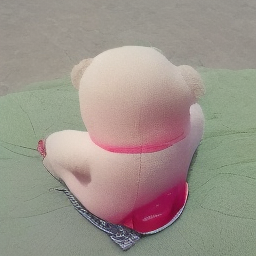}} \\

& \multicolumn{2}{c}{\textit{\texttt{[C]} wearing a green hat}} & \multicolumn{2}{c}{\textit{\texttt{[C]} sits on a green blanket}} \\

\rotatebox[origin=c]{90}{\scalebox{1.0}{sample 1}} &
\raisebox{-0.5\height}{\includegraphics[width=0.12\textwidth]{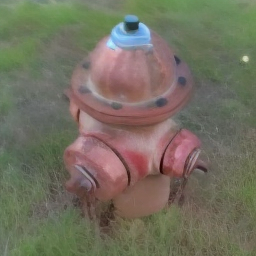}} & 
\raisebox{-0.5\height}{\includegraphics[width=0.12\textwidth]{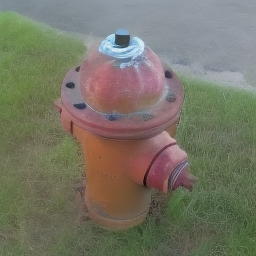}} &

\raisebox{-0.5\height}{\includegraphics[width=0.12\textwidth]{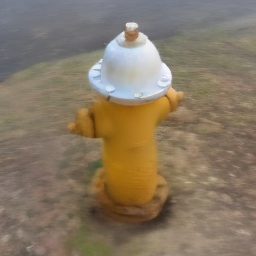}} & 
\raisebox{-0.5\height}{\includegraphics[width=0.12\textwidth]{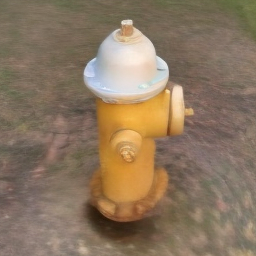}} \\

\rotatebox[origin=c]{90}{\scalebox{1.0}{sample 2}} &
\raisebox{-0.5\height}{\includegraphics[width=0.12\textwidth]{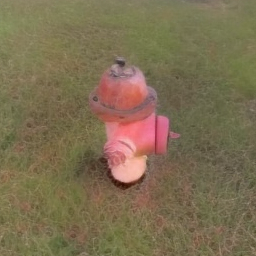}} & 
\raisebox{-0.5\height}{\includegraphics[width=0.12\textwidth]{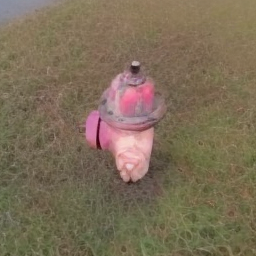}} &

\raisebox{-0.5\height}{\includegraphics[width=0.12\textwidth]{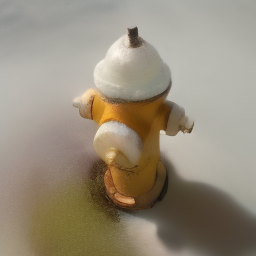}} & 
\raisebox{-0.5\height}{\includegraphics[width=0.12\textwidth]{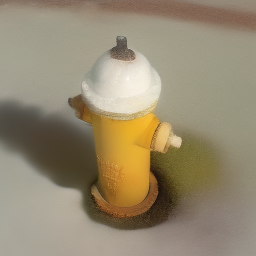}} \\

& \multicolumn{2}{c}{\textit{rusty \texttt{[C]} in the grass}} & \multicolumn{2}{c}{\textit{yellow and white \texttt{[C]}}} \\

\end{tabular}
\caption{
\textbf{Diversity of generated results.}
We condition our method on text input, which allows to create objects in a desired style.
We show samples for hand-crafted text descriptions that combine attributes (e.g., color, shape, background) in a novel way.
Each row shows a different generation proposal from our method and we denote the object category (\texttt{Teddybear}, \texttt{Hydrant}) as \texttt{[C]}.
This showcases the diversity of generated results, i.e., multiple different objects are generated for the same description.
}
\label{fig:diversity}
\end{figure}

\begin{table}
	\caption{
		\textbf{Quantitative comparison of our method and ablations.}
		We report average PSNR, SSIM, LPIPS~\cite{zhang2018unreasonable}, FID~\cite{heusel2017gans}, and KID \cite{binkowski2018demystifying} over the \texttt{Teddybear} and \texttt{Hydrant} categories.
		We compare against dropping the projection layer (``Ours no proj'') and the cross-frame-attention layer (``Ours no cfa'') from the U-Net (see \cref{subsec:aug-layers}).
		While still producing high-quality images with similar FID/KID scores, this demonstrates that our proposed layers are necessary to obtain 3D-consistent images.
	}
  \centering
  \setlength\tabcolsep{3pt}
  \begin{tabular}{l ccccc}
    \toprule
        Method & PSNR$\uparrow$ & SSIM$\uparrow$ & LPIPS$\downarrow$ & FID$\downarrow$ & KID$\downarrow$ \\
    \midrule
    Ours (no proj) & 16.55 & 0.71 & 0.29 & 47.95 & \textbf{0.034} \\
    Ours (no cfa) & 18.15 & 0.76 & 0.25 & 47.93 & \textbf{0.034} \\
    Ours & \textbf{22.24} & \textbf{0.84} & \textbf{0.11} & \textbf{47.92} & \textbf{0.034} \\
    \bottomrule
  \end{tabular}
  \label{tab:ablation}
\end{table}

The key ingredients of our method are the cross-frame-attention and projection layers that we add to the U-Net (\cref{subsec:aug-layers}).
We highlight their importance in \cref{tab:ablation} and \cref{fig:uncond-360}.

\paragraph{How important are the projection layers?}
They are necessary to allow precise control over the image viewpoints (e.g., \cref{fig:uncond-360} row 3 does not follow the specified rotation).
Our goal is to generate a consistent set of images from \emph{any} viewpoint \emph{directly} with our model (\cref{subsec:autoreg-gen}).
Being able to control the pose of the object is therefore an essential part of our contribution.
The projection layers build up a 3D representation of the object that is explicitly rendered into 3D-consistent features through volume rendering.
This allows us to achieve viewpoint consistency, as also demonstrated through single-image reconstruction (\cref{tab:ablation}).

\paragraph{How important are cross-frame-attention layers?}
They are necessary to create images of the same object.
Without them, the teddybear in \cref{fig:uncond-360} (row 4) has the same general color scheme and follows the specified poses.
However, differences in shape and texture lead to an inconsistent set of images.
We reason that the cross-frame-attention layers are essential for defining a consistent \emph{object identity}.

\paragraph{Does the 2D prior help?}
We utilize a 2D prior in form of the pretrained text-to-image model that we fine-tune in a 3D-consistent fashion (\cref{subsec:theory}).
This enables our method to produce sharp and detailed images of objects from different viewpoints.
Also, we train our method on captioned images to retain the controllable generation through text descriptions (\cref{subsec:impl-details}).
We show the diversity and controllability of our generations in \cref{fig:diversity} with hand-crafted text prompts.
This highlights that, after finetuning, our model is still faithful to text input, and can combine attributes in a novel way, i.e., our model learns to extrapolate from the training set.

\subsection{Limitations}
Our method generates 3D-consistent, high-quality images of diverse objects according to text descriptions or input images.
Nevertheless, there are several limitations.
First, our method sometimes produces images with slight inconsistency, as shown in the supplement.
Since the model is fine-tuned on a real-world dataset consisting of view-dependent effects (e.g., exposure changes), our framework learns to generate such variations across different viewpoints.
A potential solution is to add lighting condition through a ControlNet \cite{ZhangA2023}.
Second, our work focuses on objects, but similarly scene-scale generation on large datasets~\cite{yeshwanthliu2023scannetpp,dai2017scannet} can be explored.%

\section{Conclusion}
\label{sec:conclusion}

We presented \OURS, a method that, given text or image input, generates 3D-consistent images of real-world objects that are placed in authentic surroundings.
Our method leverages the expressivity of large 2D text-to-image models and fine-tunes this 2D prior on real-world 3D datasets to produce diverse multi-view images in a joint denoising process.
The core insight of our work are two novel layers, namely cross-frame-attention and the projection layer (\cref{subsec:aug-layers}).
Our autoregressive generation scheme (\cref{subsec:autoreg-gen}) allows to directly render high-quality and realistic novel views of a generated 3D object.

\section{Acknowledgements}
This work was done during Lukas' internship at Meta Reality Labs Zurich as well as at TU Munich,
funded by a Meta sponsored research agreement.
Matthias Nie{\ss}ner was also supported by the ERC Starting Grant Scan2CAD (804724).
We also thank Angela Dai for the video voice-over.

{
    \small
    \bibliographystyle{ieeenat_fullname}
    \bibliography{main,3D-CIGTIM-CR}
}

\clearpage
\maketitlesupplementary
\appendix

\section{Supplemental Video}
Please watch our attached video~\footnote[1]{\url{https://youtu.be/SdjoCqHzMMk}} for a comprehensive evaluation of the proposed method.
We include rendered videos of multiple generated objects from novel trajectories at different camera elevations (showcasing unconditional generation as in \cref{fig:uncond-generation}).
We also show animated results for single-image reconstruction (\cref{fig:cond-generation}) and sample diversity (\cref{fig:diversity}).

\section{Training Details}

\subsection{Data Preprocessing}
\label{sec:data-preproc}

We train our method on the large-scale CO3Dv2 \cite{ReizeSHSLN2021} dataset, which consists of posed multi-view images of real-world objects.
Concretely, we choose the categories \texttt{Teddybear}, \texttt{Hydrant}, \texttt{Apple}, and \texttt{Donut}.
Per category, we train on 500–1000 objects, with 200 images at resolution $256{\times}256$ for each object.
We generate text captions with the BLIP-2 model \cite{li2023blip} and sample one of 5 proposals per object during each training iteration.
With probability $p_1{=}0.5$ we select the training images randomly per object and with probability $p_2{=}0.5$ we select consecutive images from the captured trajectory that goes around the object.
We randomly crop the images to a resolution of 256$\times$256 and normalize the camera poses such that the captured object lies in an axis-aligned unit cube.
Specifically, we follow \citet{szymanowicz23viewset_diffusion} and calculate a rotation transformation such that all cameras align on an axis-aligned plane.
Then, we translate and scale the camera positions, such that their bounding box is contained in the unit cube.

\subsection{Prior Preservation Loss}
\label{sec:prior-preservation}
Inspired by \citet{ruiz2023dreambooth}, we create a \textit{prior preservation} dataset of 300 images and random poses per category with the pretrained text-to-image model.
We use it during training to maintain the image generation prior.
This has been shown to be successful when fine-tuning a large 2D diffusion model on smaller-scale data~\cite{ruiz2023dreambooth}.
For each of the 300 images we randomly sample a text description from the training set of CO3Dv2 \cite{ReizeSHSLN2021}.
We then generate an image with the pretrained text-to-image model given that text description as input.
During each training iteration we first calculate the diffusion objective (\cref{eq:ddpm-eps-loss}) on the $N {=} 5$ multi-view images sampled from the dataset and obtain $L_d$.
Then, we sample one image of the \textit{prior preservation} dataset and apply noise to it (\cref{eq:ddpm-forward-process}).
Additionally, we sample a camera (pose and intrinsics) that lies within the distribution of cameras for each object category.
We then similarly calculate the loss (\cref{eq:ddpm-eps-loss}) on the prediction of our model and obtain $L_p$.
Since we only sample a single image instead of multiple, this does not train the diffusion model on 3D-consistency.
Instead, it trains the model to maintain its image generation prior.
Concretely, the cross-frame-attention layers are treated again as self-attention layers and the projection layers perform unprojection and rendering normally, but only from a single image as input.
In practice, we scale the prior preservation loss with factor $0.1$ and add it to the dataset loss to obtain the final loss: $L {=} L_d + 0.1 L_p$.

\section{Evaluation Details}
\label{sec:eval-details}

\subsection{Autoregressive Generation}
We showcase unconditional generation of our method in \cref{subsec:res-uncond}.
To obtain these results, we employ our autoregressive generation scheme (\cref{subsec:autoreg-gen}).
Concretely, we sample an (unobserved) image caption from the test set for the first batch and generate $N{=}10$ images with a guidance scale~\cite{ho2022classifier} of $\lambda_\text{cfg}{=}7.5$.
Then we set $\lambda_\text{cfg}{=}{0}$ for subsequent batches and create a total of 100 images per object.
We found that the results are most consistent, if the first batch generates $N$ images in a 360° rotation around the object.
This way, we globally define the object shape and texture in a single denoising forward pass.
All subsequent batches are conditioned on all $N$ images of the first batch.
To render a smooth trajectory, we sample the camera poses in other batches in a sequence.
That is, the next $N$ images are close to each other with only a small rotation between them.
We visualize this principle in our supplemental video.

\subsection{Metric Computation}

We give additional details on how we computed the metrics as shown in \cref{tab:fid,tab:cond-generation,tab:ablation}.
To ensure comparability, we evaluate all metrics on images without backgrounds as not every baseline models them.

\paragraph{FID/KID}
We report FID~\cite{heusel2017gans} and KID~\cite{binkowski2018demystifying} as common metrics for 2D/3D generation.
We calculate these metrics to compare \textit{unconditional} image generation against HoloFusion~\cite{karnewar2023holofusion} and ViewsetDiffusion~\cite{szymanowicz23viewset_diffusion}.
This quantifies the similarity of the generated images to the dataset and thereby provides insight about their quality (e.g., texture details and sharpness) and diversity (e.g., different shapes and colors).
Following the baselines~\cite{karnewar2023holodiffusion,karnewar2023holofusion}, we sample 20,000 images from the CO3Dv2 \cite{ReizeSHSLN2021} dataset for each object category.
We remove the background from each object by using the foreground mask probabilities contained in the dataset.
Similarly, we generate 20,000 images with each method and remove the background from our generated images with CarveKit \cite{carvekit}.
For our method, we set the text prompt to an empty string during the generation to facilitate complete unconditional generation.

\paragraph{PSNR/SSIM/LPIPS}
We measure the multi-view consistency of generated images with peak signal-to-noise ratio (PSNR), structural similarity index (SSIM), and LPIPS~\cite{zhang2018unreasonable}.
We calculate these metrics to compare single-image reconstruction against ViewsetDiffusion~\cite{szymanowicz23viewset_diffusion} and DFM~\cite{tewari2023forwarddiffusion}.
We resize all images to the resolution 256$\times$256 to obtain comparable numbers.
First, we obtain all objects that were not used during training for every method (hereby obtaining a unified test set across all methods).
Then, we randomly sample 20 posed image pairs from each object.
We use the first image/pose as input and predict the novel view at the second pose.
We then calculate the metrics as the similarity of the prediction to the ground-truth second image.
We remove the background from the prediction and ground-truth images by obtaining the foreground mask with CarveKit~\cite{carvekit} from the \textit{prediction} image.
We use the same mask to remove background from both images.
This way, we calculate the metrics only on similarly masked images.
If the method puts the predicted object at a wrong position, we would thus quantify this as a penalty by comparing the segmented object to the background of the ground-truth image at that location.

\section{Limitations}
\label{sec:limitations}

\begin{figure}
\centering
\setlength\tabcolsep{0.5pt}
\begin{tabular}{c@{}c c@{}c}

\includegraphics[width=0.124\textwidth]{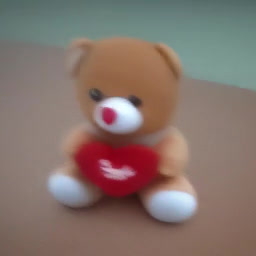} & 
\includegraphics[width=0.124\textwidth]{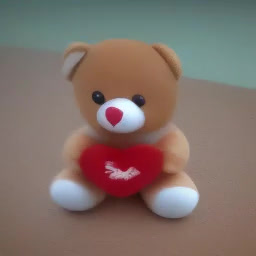} & 

\includegraphics[width=0.124\textwidth]{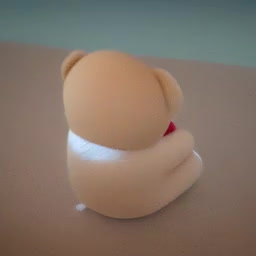} & 
\includegraphics[width=0.124\textwidth]{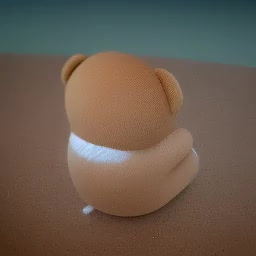} \\

\includegraphics[width=0.124\textwidth]{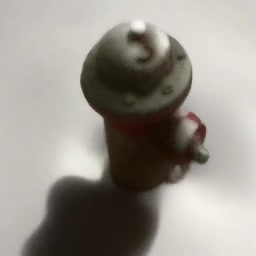} & 
\includegraphics[width=0.124\textwidth]{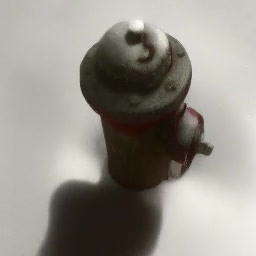} & 

\includegraphics[width=0.124\textwidth]{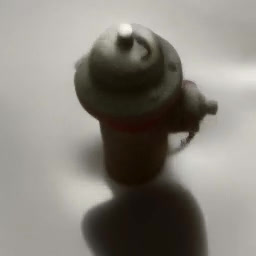} & 
\includegraphics[width=0.124\textwidth]{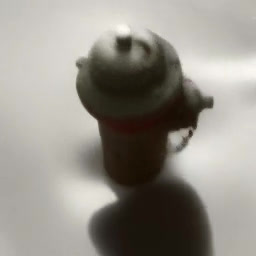} \\

\multicolumn{2}{c}{different sharpness} & \multicolumn{2}{c}{different lighting}

\end{tabular}
\caption{
\textbf{Limitations.}
Our method generates consistent images at different camera poses.
However, there can be slight inconsistencies like different sharpness and lighting between images.
Since our model is fine-tuned on a real-world dataset consisting of view-dependent effects (e.g., exposure changes), our framework learns to generate such variations across different viewpoints.
}
\label{fig:limitations}
\end{figure}

Our method generates 3D-consistent, high-quality images of diverse objects according to text descriptions or input images.
Nevertheless, there are several limitations.
Our method sometimes produces images with slight inconsistency, as shown in \cref{fig:limitations}.
Since the model is fine-tuned on a real-world dataset consisting of view-dependent effects (e.g., exposure changes), our framework learns to generate such variations across different viewpoints.
This can lead to flickering artifacts when rendering a smooth video from a generated set of images (e.g., see the supplemental video).
A potential solution is to (i) filter blurry frames from the dataset, and (ii) add lighting-condition through a ControlNet~\cite{ZhangA2023}.

\section{Projection Layer Architecture}
\label{sec:model-architecture}

\begin{figure}
\centering
\setlength\tabcolsep{0.5pt}
\begin{tabular}{cc}

\includegraphics[width=0.24\textwidth]{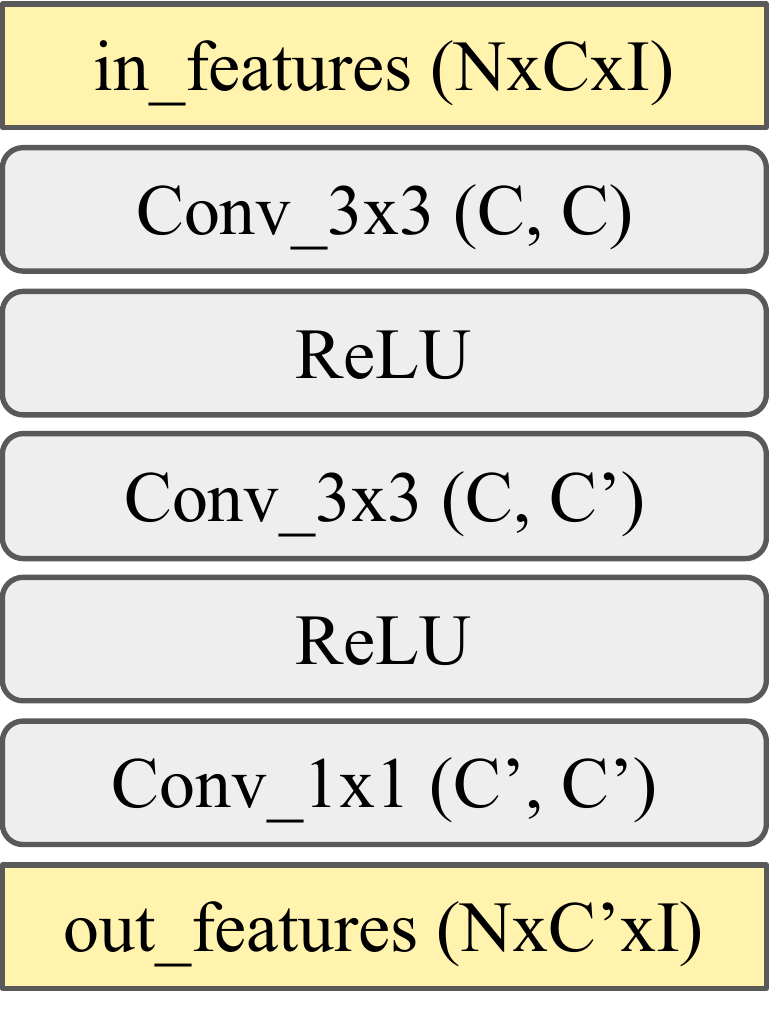} & 
\includegraphics[width=0.24\textwidth]{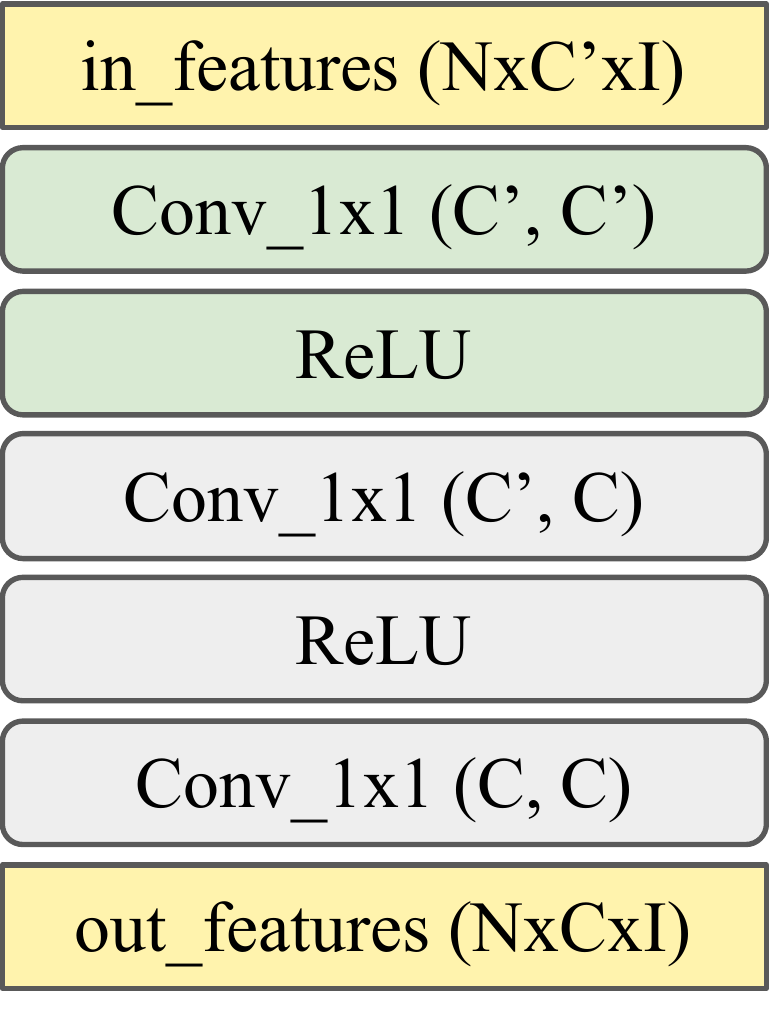} \\

CompressNet & \makecell[c]{ScaleNet \& \\ ExpandNet}

\end{tabular}
\caption{
\textbf{Architecture of projection layer components.}
The projection layer contains the components \textit{CompressNet}, \textit{ScaleNet} (green), and \textit{ExpandNet}.
We implement these networks as small CNN networks.
}
\label{fig:arch-in-out}
\end{figure}

\begin{figure}
\centering
\setlength\tabcolsep{0.5pt}
\begin{tabular}{c}

\includegraphics[width=0.5\textwidth]{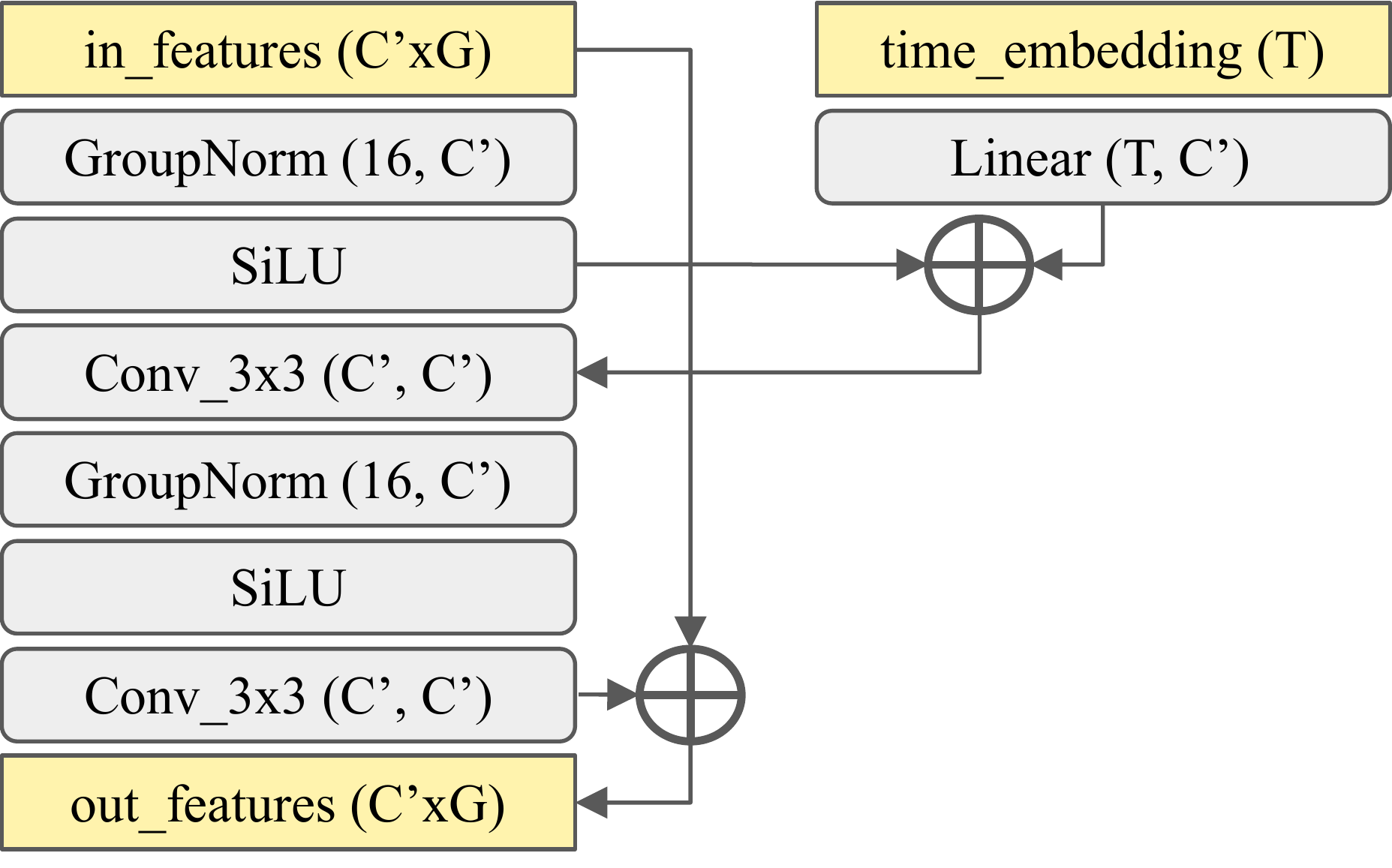} \\

ResNet block of the 3D CNN

\end{tabular}
\caption{
\textbf{Architecture of projection layer components.}
The projection layer contains the component \textit{3D CNN}.
We implement this networks as a series of 5 3D ResNet \cite{HeZRS2016} blocks with timestep embeddings.
}
\label{fig:arch-3d-net}
\end{figure}

\begin{figure*}
\centering
\setlength\tabcolsep{0.5pt}
\begin{tabular}{c}

\includegraphics[width=\textwidth]{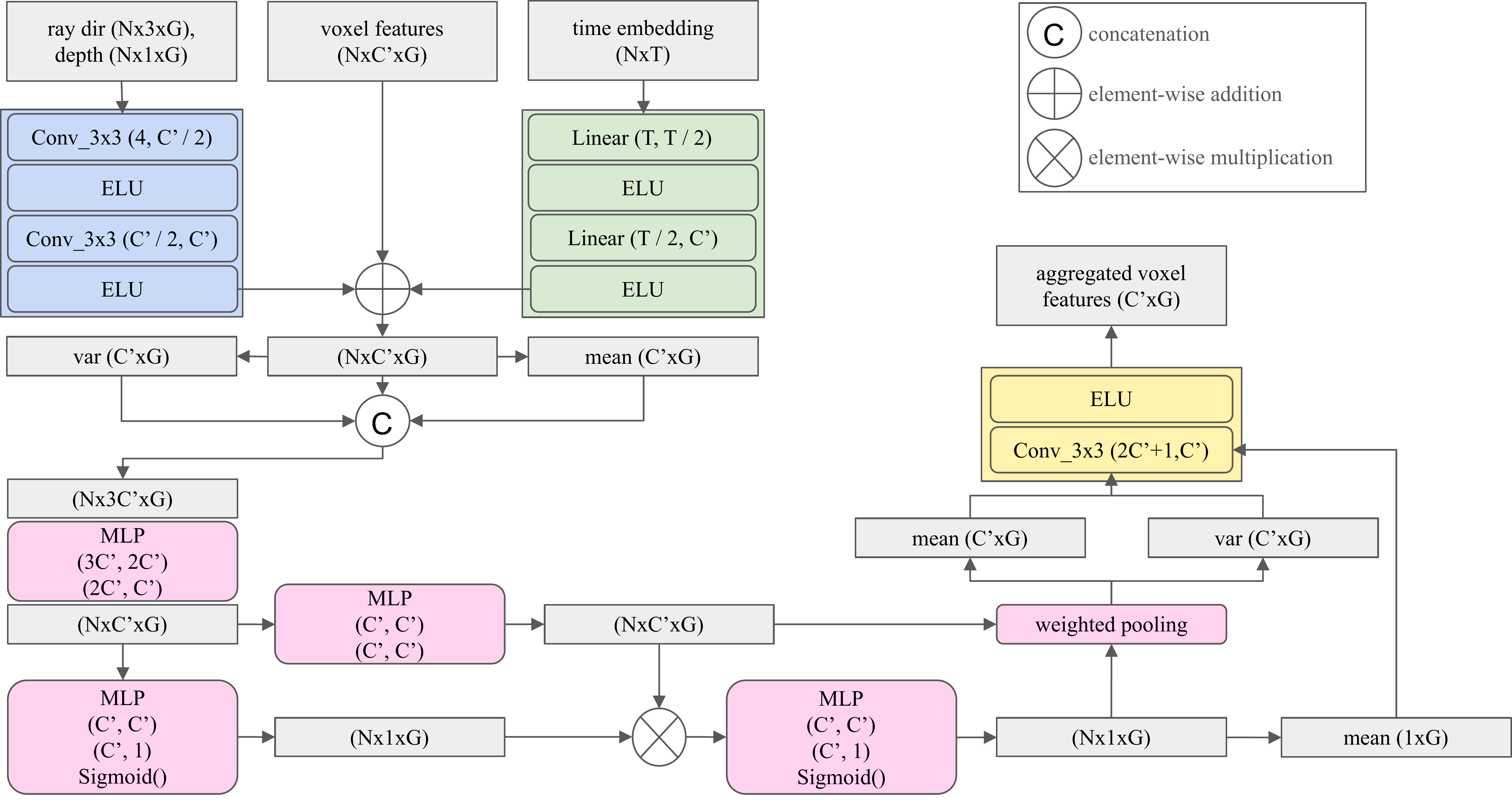} \\

Aggregator MLP

\end{tabular}
\caption{
\textbf{Architecture of projection layer components.}
The projection layer contains the component \textit{Aggregator MLP}.
First, we combine per-view voxel grids with their ray-direction/depth encodings (blue) and the temporal embedding (green).
Inspired by IBRNet \cite{WangWGSZBMSF2021}, the MLPs (pink) then predict per-view weights followed by a weighted feature average.
Finally, we combine the per-voxel weights with the mean and variance grids (yellow) to obtain the aggregated feature grid.
}
\label{fig:arch-agg-net}
\end{figure*}

We add a projection layer into the U-Net architecture of pretrained text-to-image models (see \cref{fig:proj-layer} and \cref{subsec:aug-layers}).
The idea of this layer is to create 3D-consistent features that are then further processed by the next U-Net layers (e.g. ResNet blocks).
Concretely, we create a 3D representation from all input features in form of a voxel grid, that is defined inside of the axis-aligned unit cube.
We set the 3D feature dimension as $C'=16$ and define the base resolution of the voxel grid as 128$\times$128$\times$128.
Throughout the U-Net, we apply the same up/downsampling as for the 2D features, i.e., the resolution decreases to 8$\times$8$\times$8 in the bottleneck layer.
The projection layer consists of multiple network components.
We show detailed network architectures of these components in \cref{fig:arch-in-out,fig:arch-3d-net,fig:arch-agg-net}.

\subsection{CompressNet and ExpandNet}
We apply the 3D layers on features that are defined in a unified dimensionality of $C' {=} 16$.
Since our 3D layers act on dense voxel grids this helps to lower the memory requirements.
To convert to/from this compressed feature space, we employ small CNNs, as depicted in \cref{fig:arch-in-out}.
In these schematics, we define $N$ as the number of images in a batch, $C$ as the uncompressed feature dimension and $I$ as the spatial dimension of the features.

\subsection{Aggregator MLP}
After creating per-view voxel grids via raymarching (see \cref{subsec:aug-layers}), we combine $N$ voxel grids into one voxel grid that represents the features for all viewpoints.
To this end, we employ a series of networks, as depicted in  \cref{fig:arch-agg-net}.
In these schematics, we define $N$ as the number of images in a batch, $C'$ as the compressed feature dimension, $T$ as the dimension of the timestep embedding, $G$ as the 3D voxel grid resolution, and $I$ as the spatial dimension of the features.
The MLPs are defined as a sequence of linear layers of specified input and output dimensionality with \textit{ELU} activations in between.

First, we concatenate the voxel features with an encoding of the ray-direction and depth that was used to project the image features into each voxel.
We also concatenate the timestep embedding to each voxel.
This allows to combine per-view voxel grids of different timesteps (e.g., as proposed in image conditional generation in \cref{subsec:autoreg-gen}).
It is also useful to inform the subsequent networks about the denoising timestep, which allows to perform the aggregation differently throughout the denoising process.
Inspired by IBRNet \cite{WangWGSZBMSF2021}, a set of MLPs then predict per-view weights followed by a weighted feature average.
We perform this averaging operation elementwise: since all voxel grids are defined in the same unit cube, we can combine the same voxel across all views.
Finally, we combine the per-voxel weights with the mean and variance grids to obtain the final aggregated feature grid.

\subsection{3D CNN}
After aggregating the per-view voxel grids into a joint grid, we further refine that grid.
The goal of this network is to add additional details to the feature representation such as the global orientation of the shape.
To achieve this, we employ a series of 5 3D ResNet \cite{HeZRS2016} blocks with timestep embeddings, as depicted in \cref{fig:arch-3d-net}.
In these schematics, we define $C'$ as the compressed feature dimension, $T$ as the dimension of the timestep embedding, and $G$ as the 3D voxel grid resolution.

\subsection{Volume Renderer and ScaleNet}
After we obtain a refined 3D feature representation in form of the voxel grid, we render that grid back into per-view image features (see \cref{fig:proj-layer}).
Concretely, we employ a volume renderer similar to NeRF~\cite{mildenhall2021nerf} and implement it as a grid-based renderer similar to DVGO~\cite{SunSC22}.
This allows to render features in an efficient way that is not a bottleneck for the forward pass of the network.
In contrast to NeRF, we render down \textit{features} instead of \textit{rgb} colors.
Concretely, we sample 128 points along a ray and for each point we trilinearly interpolate the voxel grid features to obtain a feature vector $f \in \mathbb{R}^{C'}$.
Then, we employ a small 3-layer MLP that transforms $f$ into the density $d \in \mathbb{R}$ and a sampled feature $s \in \mathbb{R}^{C'}$.
Using alpha-compositing, we accumulate all pairs $(d_0, s_0), ..., (d_{127}, s_{127})$ along a ray into a final rendered feature $r \in \mathbb{R}^{C'}$.
We dedicate half of the voxel grid to foreground and half to background and apply the background model from MERF~\cite{reiser2023merf} during ray-marching.

We found it is necessary to add a scale function after the volume rendering output.
The volume renderer typically uses a \emph{sigmoid} activation function as the final layer during ray-marching \cite{mildenhall2021nerf}.
However, the input features are defined in an arbitrary floating-point range.
To convert $r$ back into the same range, we non-linearly scale the features with 1$\times$1 convolutions and \emph{ReLU} activations.
We depict the architecture of this \textit{ScaleNet} as the green layers in \cref{fig:arch-in-out}.

\section{Additional Results}

\subsection{Comparison To Additional Baselines}
\label{sec:add-results-rebuttal}

\begin{figure}[ht]
\setlength\tabcolsep{1pt}
\def\arraystretch{0.7}%
\resizebox{\linewidth}{!}{%
\begin{tabular}{c@{}c|c@{}c}
\includegraphics[width=0.12\textwidth]{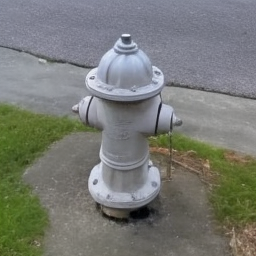} &
\multicolumn{1}{c}{\includegraphics[width=0.12\textwidth]{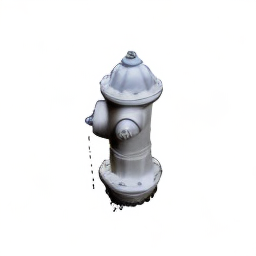}} &
\multicolumn{1}{c}{\includegraphics[width=0.12\textwidth]{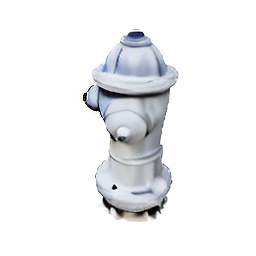}} &
\includegraphics[width=0.12\textwidth]{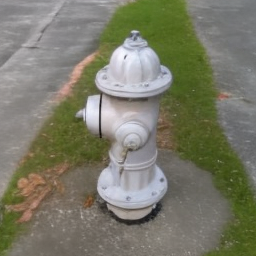} \\
{\small Input Image} & \multicolumn{1}{c}{{\small Zero123-XL~\cite{Liu_2023_ICCV}}} & \multicolumn{1}{c}{{\small SyncDreamer~\cite{liu2023syncdreamer}}} & {\small Ours} \\
\hline & \\[-1.5ex]
\multicolumn{2}{c}{\small \textit{teddy sitting on a wooden box}} & \multicolumn{2}{c}{\small \textit{donut on top of a white plate}} \\
\includegraphics[width=0.12\textwidth]{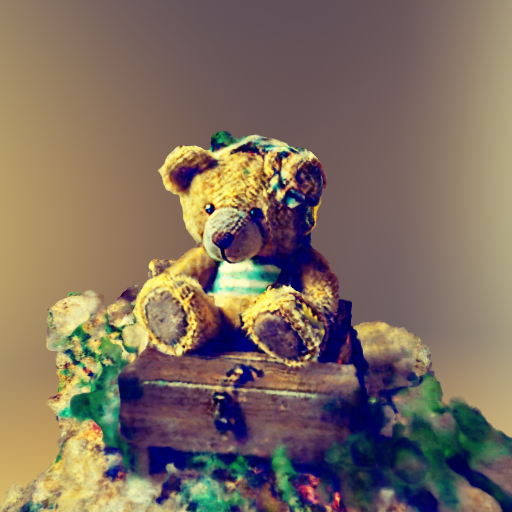} &
\includegraphics[width=0.12\textwidth]{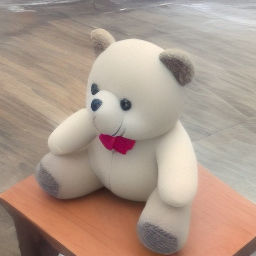} &
\includegraphics[width=0.12\textwidth]{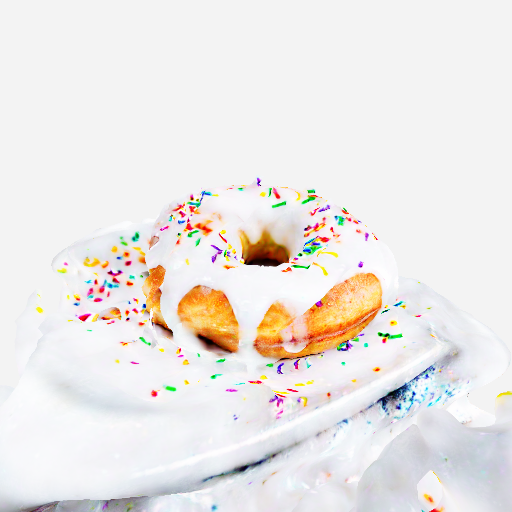} &
\includegraphics[width=0.12\textwidth]{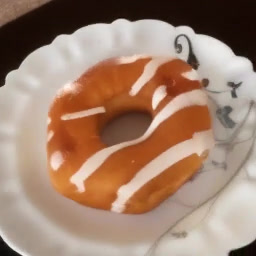} \\
{\small ProlificDreamer~\cite{WangLWBLSZ2023}} & {\small Ours} & {\small ProlificDreamer~\cite{WangLWBLSZ2023}} & {\small Ours}
\end{tabular}}
\caption{
\textbf{Comparison to other text-to-3D baselines from image- (top) and text-input (bottom).}
Our method produces images with higher photorealism and authentic surroundings.
}
\label{fig:rebuttal-qual-comp}
\end{figure}

\begin{table}
  \caption{
    \textbf{Comparison of consistency (mid) and photorealism (FID).}
    Our method shows similar 3D-consistency as baselines, while producing more photorealistic images.
  }
  \label{tab:rebuttal-quant-comp}
  \centering
  \setlength\tabcolsep{3pt}
  \begin{tabular}{l|ccc|c}
    \toprule
        Method & $E_\text{warp}\downarrow$ & \#Points$\uparrow$ & PSNR$\uparrow$ & FID$\downarrow$ \\
    \midrule
    DFM~\cite{tewari2023forwarddiffusion} & 0.0034 & 17,470 & 32.32 & --- \\
    VD~\cite{szymanowicz23viewset_diffusion} & 0.0021 & --- & --- & --- \\
    HF~\cite{karnewar2023holofusion} & 0.0031 & --- & --- & --- \\
    SyncDreamer~\cite{liu2023syncdreamer} & 0.0042 & \phantom{0}4,126 & 33.81 & 135.78 \\
    Zero123-XL (SDS)~\cite{Liu_2023_ICCV} & 0.0039 & --- & --- & 126.83 \\
    \bf Ours & 0.0036 & 18,358 & 33.65 & \phantom{0}85.08 \\
    \bottomrule
  \end{tabular}
\end{table}

We compare against additional text-to-3D baselines that also utilize a pretrained text-to-image model in \cref{fig:rebuttal-qual-comp}.
We choose ProlificDreamer~\cite{WangLWBLSZ2023} as representative of score distillation~\cite{PooleJBM2023} methods.
Rendered images are less photorealistic since the optimization may create noisy surroundings and over-saturated textures.
Similar to us, Zero123-XL~\cite{Liu_2023_ICCV} and SyncDreamer~\cite{liu2023syncdreamer} circumvent this problem by generating 3D-consistent images directly.
However, they finetune on a large synthetic dataset~\cite{deitke2023objaverse} instead of real-world images.
As a result, their images have synthetic textures and lighting effects and no backgrounds.
We quantify this in~\cref{tab:rebuttal-quant-comp} with the FID between sets of generated images (conditioned on an input view), and real images of the same object (without backgrounds).
Our method has better scores since the generated images are more photorealistic.

We calculate temporal stability ($E_\text{warp}$) of video renderings with optical flow warping following~\cite{Lai-ECCV-2018}.
Also, we measure the consistency of generated images for methods that do not directly produce a 3D representation.
Concretely, we report the number of point correspondences following~\cite{liu2023syncdreamer} and the PSNR between NeRF~\cite{mildenhall2021nerf} re-renderings and input images.
\Cref{tab:rebuttal-quant-comp} shows that our method is on-par with baselines in terms of 3D consistency, while generating higher quality images.

\subsection{Unconditional Generation}
\label{sec:add-results-uncond}

\begin{figure*}
\centering
\setlength\tabcolsep{0pt}
\renewcommand\cellset{\renewcommand\arraystretch{0}%
\setlength\extrarowheight{0pt}}
\begin{tabular}{cccccccc}

\includegraphics[width=0.1217\textwidth]{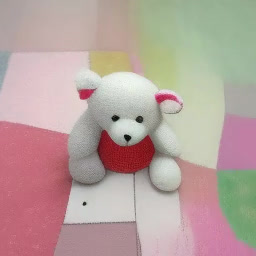} &
\includegraphics[width=0.1217\textwidth]{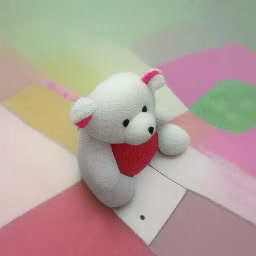} &
\includegraphics[width=0.1217\textwidth]{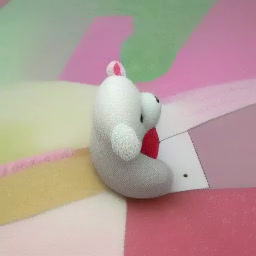} &
\includegraphics[width=0.1217\textwidth]{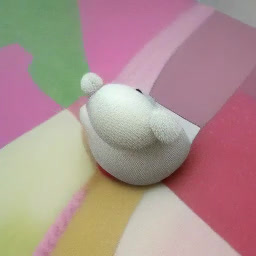} &
\includegraphics[width=0.1217\textwidth]{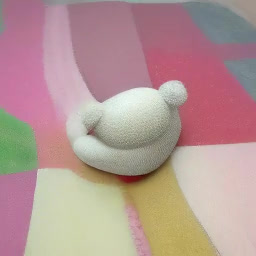} &
\includegraphics[width=0.1217\textwidth]{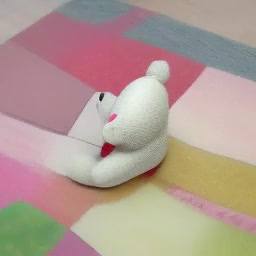} &
\includegraphics[width=0.1217\textwidth]{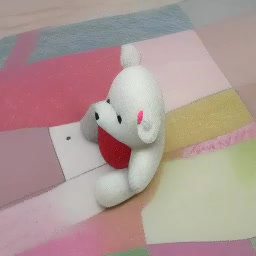} &
\includegraphics[width=0.1217\textwidth]{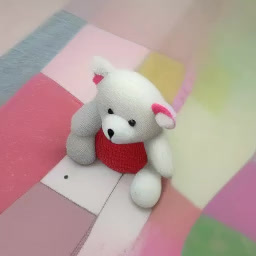} \\
\multicolumn{8}{c}{\textit{a teddy bear sitting on a colorful rug}} \\

\includegraphics[width=0.1217\textwidth]{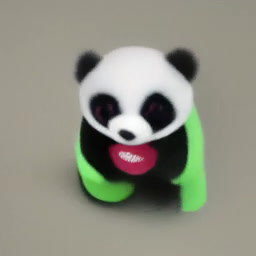} &
\includegraphics[width=0.1217\textwidth]{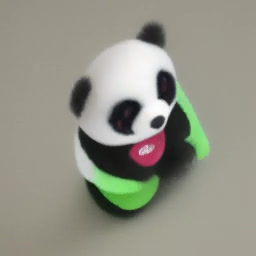} &
\includegraphics[width=0.1217\textwidth]{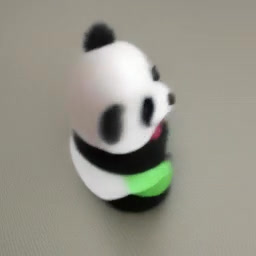} &
\includegraphics[width=0.1217\textwidth]{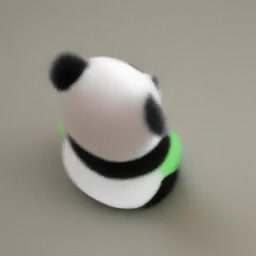} &
\includegraphics[width=0.1217\textwidth]{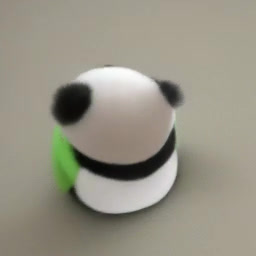} &
\includegraphics[width=0.1217\textwidth]{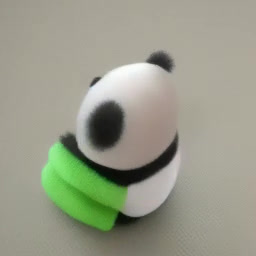} &
\includegraphics[width=0.1217\textwidth]{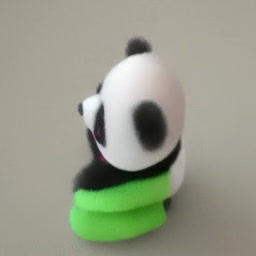} &
\includegraphics[width=0.1217\textwidth]{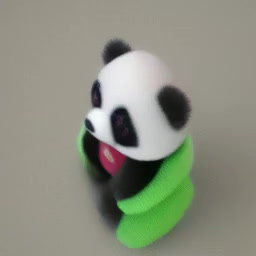} \\
\multicolumn{8}{c}{\textit{a stuffed panda bear with a heart on its chest}} \\

\includegraphics[width=0.1217\textwidth]{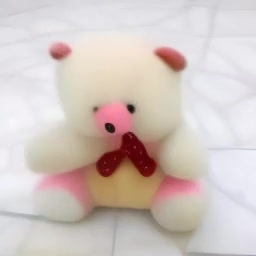} &
\includegraphics[width=0.1217\textwidth]{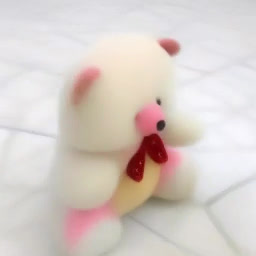} &
\includegraphics[width=0.1217\textwidth]{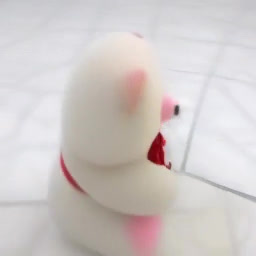} &
\includegraphics[width=0.1217\textwidth]{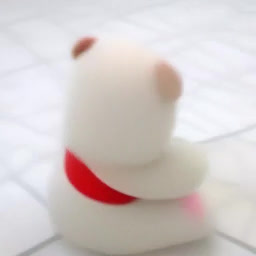} &
\includegraphics[width=0.1217\textwidth]{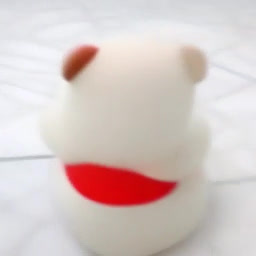} &
\includegraphics[width=0.1217\textwidth]{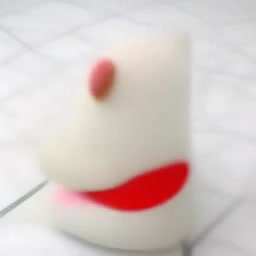} &
\includegraphics[width=0.1217\textwidth]{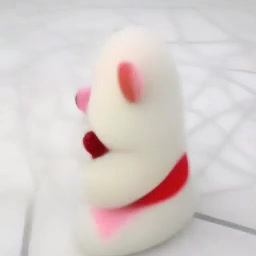} &
\includegraphics[width=0.1217\textwidth]{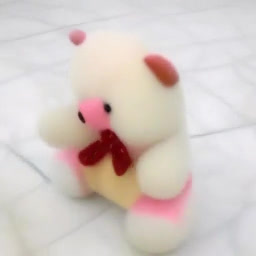} \\
\multicolumn{8}{c}{\textit{a stuffed animal sitting on a tile floor}} \\

\includegraphics[width=0.1217\textwidth]{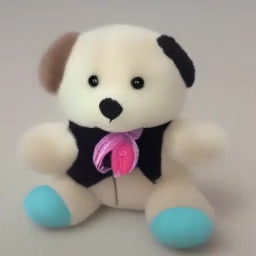} &
\includegraphics[width=0.1217\textwidth]{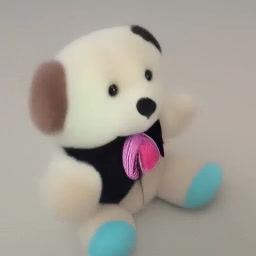} &
\includegraphics[width=0.1217\textwidth]{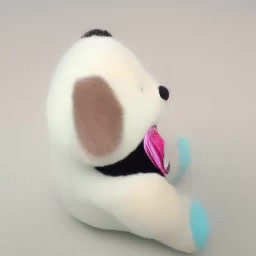} &
\includegraphics[width=0.1217\textwidth]{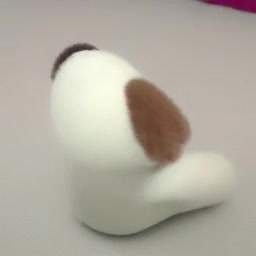} &
\includegraphics[width=0.1217\textwidth]{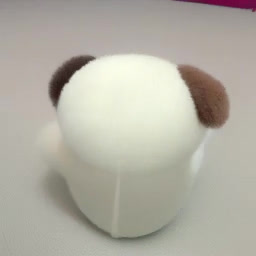} &
\includegraphics[width=0.1217\textwidth]{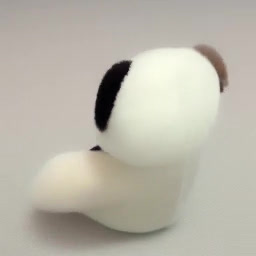} &
\includegraphics[width=0.1217\textwidth]{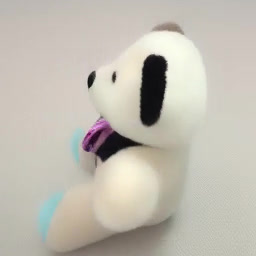} &
\includegraphics[width=0.1217\textwidth]{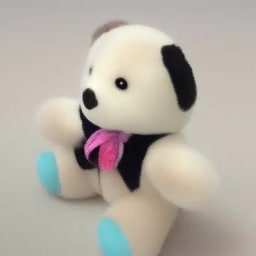} \\
\multicolumn{8}{c}{\textit{a black and white teddybear with blue feet}} \\

\includegraphics[width=0.1217\textwidth]{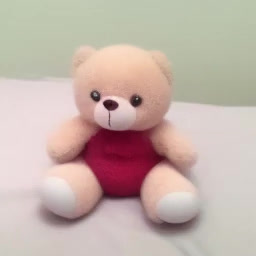} &
\includegraphics[width=0.1217\textwidth]{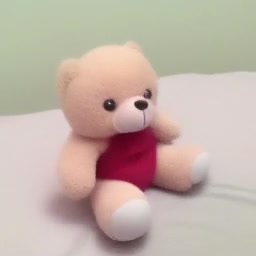} &
\includegraphics[width=0.1217\textwidth]{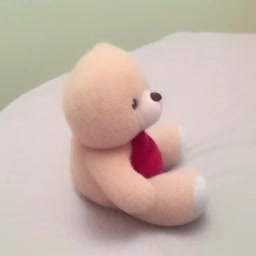} &
\includegraphics[width=0.1217\textwidth]{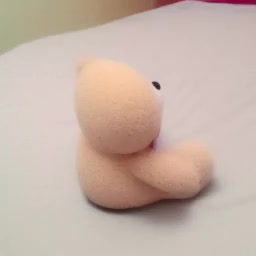} &
\includegraphics[width=0.1217\textwidth]{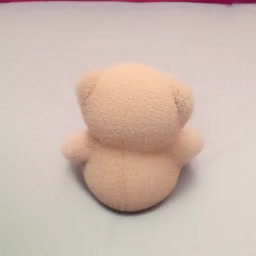} &
\includegraphics[width=0.1217\textwidth]{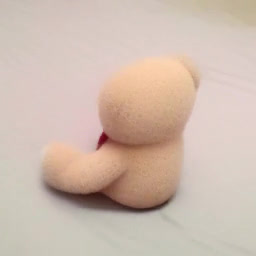} &
\includegraphics[width=0.1217\textwidth]{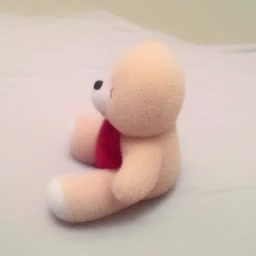} &
\includegraphics[width=0.1217\textwidth]{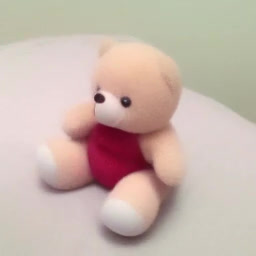} \\
\multicolumn{8}{c}{\textit{a teddy bear laying on a bed}} \\

\includegraphics[width=0.1217\textwidth]{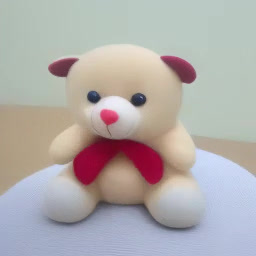} &
\includegraphics[width=0.1217\textwidth]{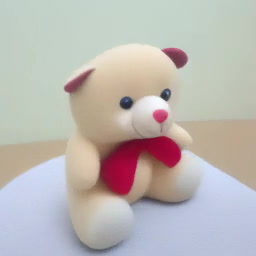} &
\includegraphics[width=0.1217\textwidth]{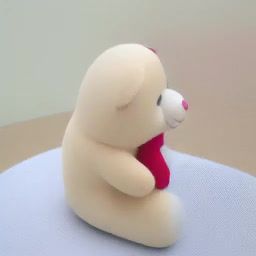} &
\includegraphics[width=0.1217\textwidth]{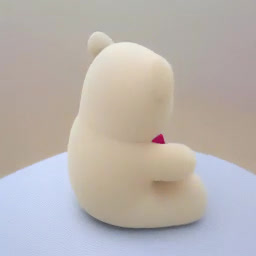} &
\includegraphics[width=0.1217\textwidth]{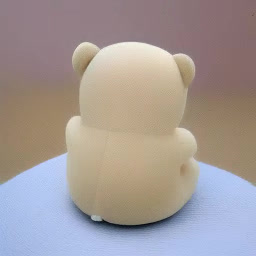} &
\includegraphics[width=0.1217\textwidth]{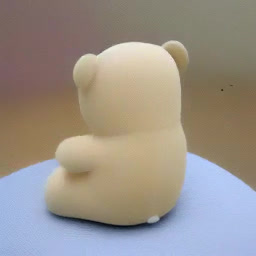} &
\includegraphics[width=0.1217\textwidth]{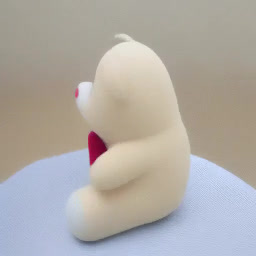} &
\includegraphics[width=0.1217\textwidth]{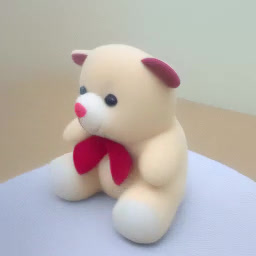} \\
\multicolumn{8}{c}{\textit{a stuffed animal sitting on a chair}} \\

\includegraphics[width=0.1217\textwidth]{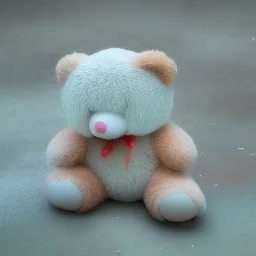} &
\includegraphics[width=0.1217\textwidth]{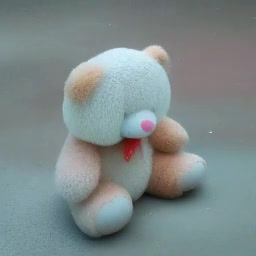} &
\includegraphics[width=0.1217\textwidth]{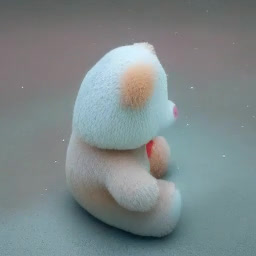} &
\includegraphics[width=0.1217\textwidth]{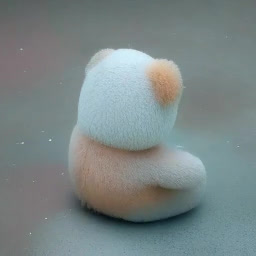} &
\includegraphics[width=0.1217\textwidth]{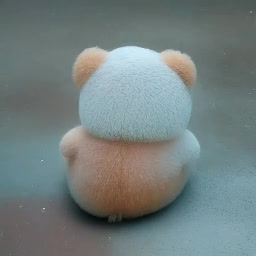} &
\includegraphics[width=0.1217\textwidth]{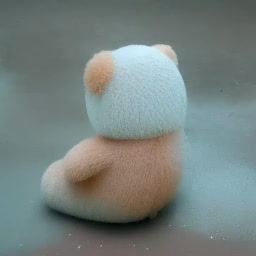} &
\includegraphics[width=0.1217\textwidth]{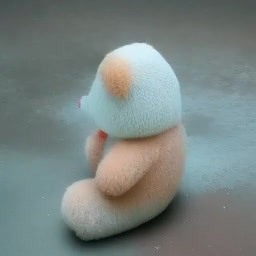} &
\includegraphics[width=0.1217\textwidth]{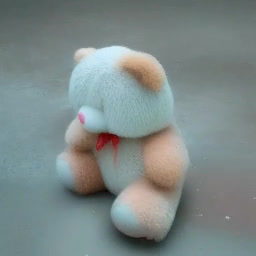} \\
\multicolumn{8}{c}{\textit{a teddy bear sitting on the ground in the dark}} \\

\includegraphics[width=0.1217\textwidth]{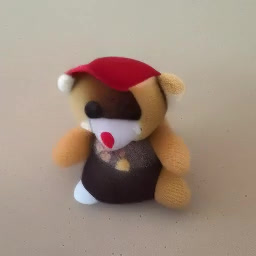} &
\includegraphics[width=0.1217\textwidth]{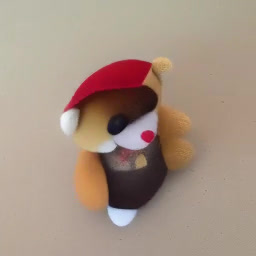} &
\includegraphics[width=0.1217\textwidth]{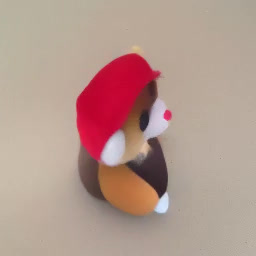} &
\includegraphics[width=0.1217\textwidth]{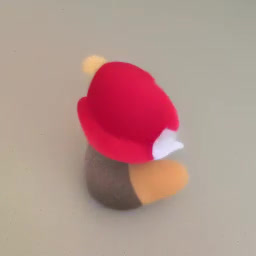} &
\includegraphics[width=0.1217\textwidth]{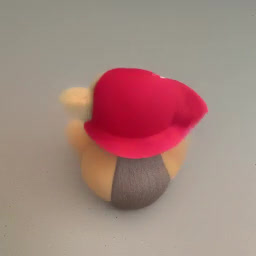} &
\includegraphics[width=0.1217\textwidth]{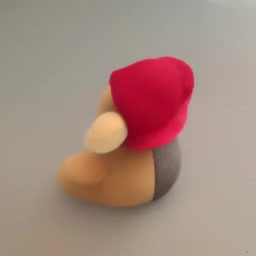} &
\includegraphics[width=0.1217\textwidth]{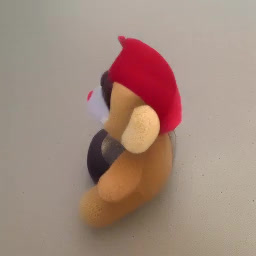} &
\includegraphics[width=0.1217\textwidth]{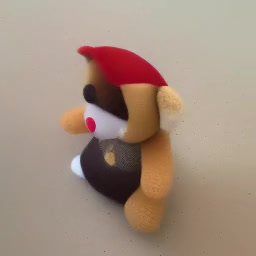} \\
\multicolumn{8}{c}{\textit{a stuffed bear wearing a red hat and a cloak}} \\

\end{tabular}
\caption{
\textbf{Additional examples of our method.}
Given a text prompt as input, we generate a smooth trajectory around an object with our autoregressive generation scheme (\cref{subsec:autoreg-gen}).
Please see the supplemental video for animations of the generated samples.
}
\label{fig:suppl-uncond-1}
\end{figure*}

\begin{figure*}
\centering
\setlength\tabcolsep{0pt}
\renewcommand\cellset{\renewcommand\arraystretch{0}%
\setlength\extrarowheight{0pt}}
\begin{tabular}{cccccccc}

\includegraphics[width=0.1217\textwidth]{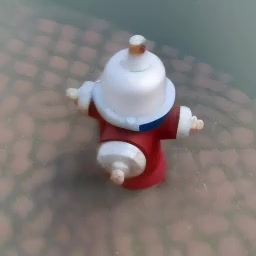} &
\includegraphics[width=0.1217\textwidth]{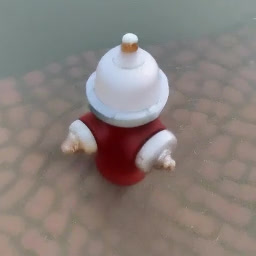} &
\includegraphics[width=0.1217\textwidth]{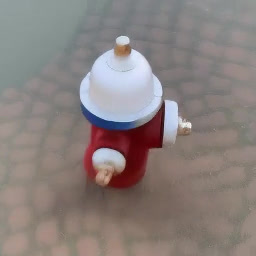} &
\includegraphics[width=0.1217\textwidth]{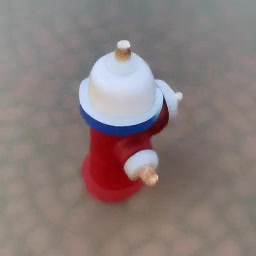} &
\includegraphics[width=0.1217\textwidth]{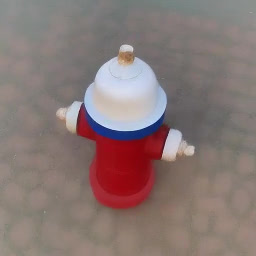} &
\includegraphics[width=0.1217\textwidth]{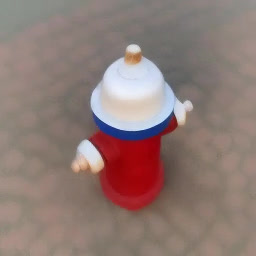} &
\includegraphics[width=0.1217\textwidth]{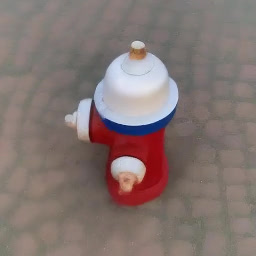} &
\includegraphics[width=0.1217\textwidth]{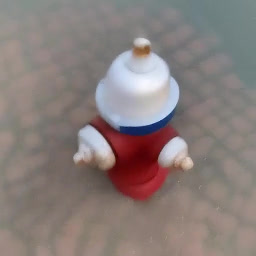} \\
\multicolumn{8}{c}{\textit{a red and white fire hydrant on a brick floor}} \\

\includegraphics[width=0.1217\textwidth]{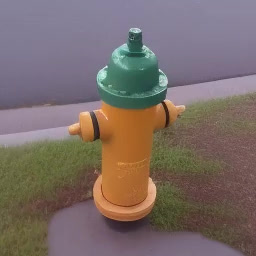} &
\includegraphics[width=0.1217\textwidth]{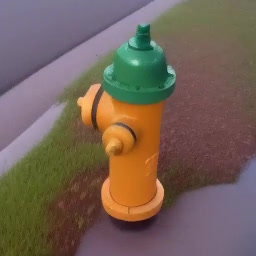} &
\includegraphics[width=0.1217\textwidth]{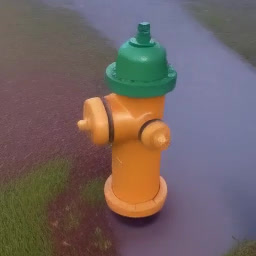} &
\includegraphics[width=0.1217\textwidth]{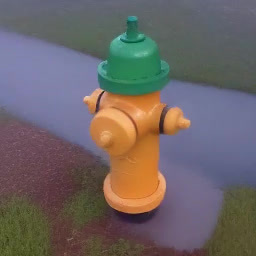} &
\includegraphics[width=0.1217\textwidth]{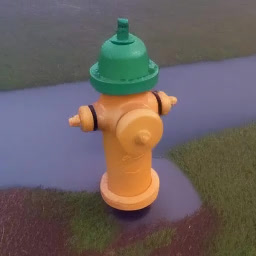} &
\includegraphics[width=0.1217\textwidth]{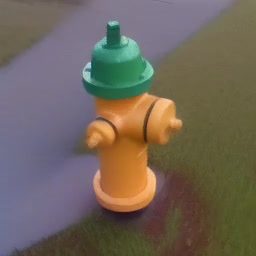} &
\includegraphics[width=0.1217\textwidth]{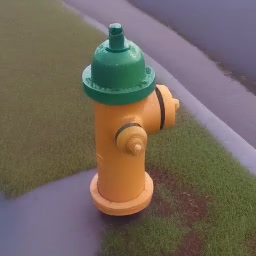} &
\includegraphics[width=0.1217\textwidth]{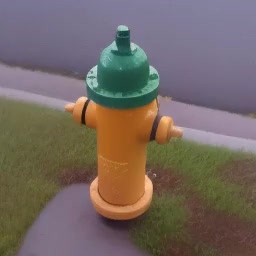} \\
\multicolumn{8}{c}{\textit{a yellow and green fire hydrant sitting on the ground}} \\

\includegraphics[width=0.1217\textwidth]{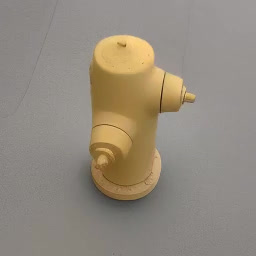} &
\includegraphics[width=0.1217\textwidth]{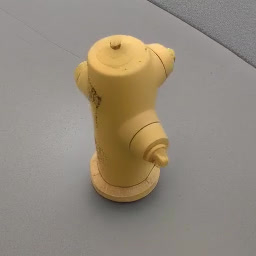} &
\includegraphics[width=0.1217\textwidth]{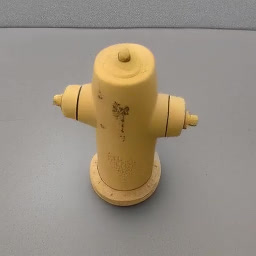} &
\includegraphics[width=0.1217\textwidth]{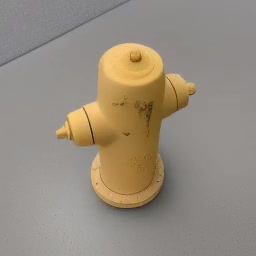} &
\includegraphics[width=0.1217\textwidth]{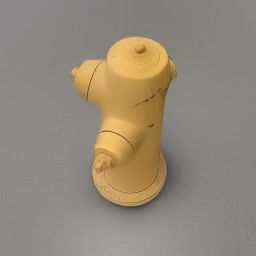} &
\includegraphics[width=0.1217\textwidth]{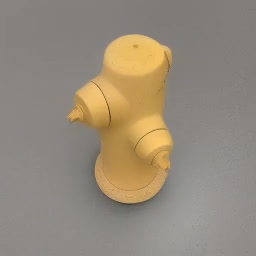} &
\includegraphics[width=0.1217\textwidth]{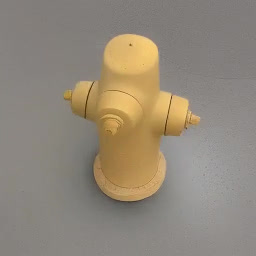} &
\includegraphics[width=0.1217\textwidth]{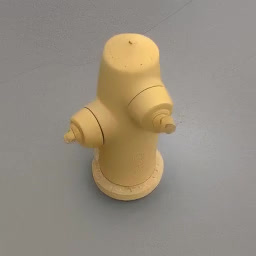} \\
\multicolumn{8}{c}{\textit{a yellow fire hydrant sitting on the sidewalk}} \\

\includegraphics[width=0.1217\textwidth]{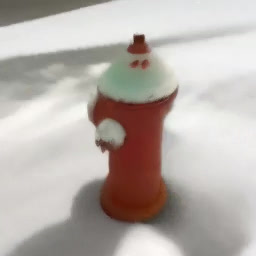} &
\includegraphics[width=0.1217\textwidth]{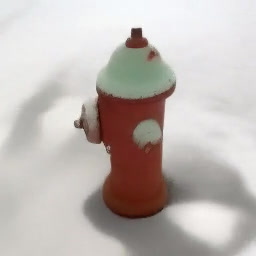} &
\includegraphics[width=0.1217\textwidth]{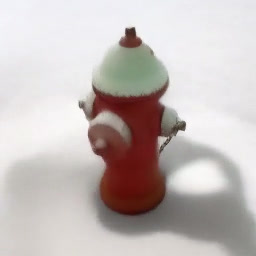} &
\includegraphics[width=0.1217\textwidth]{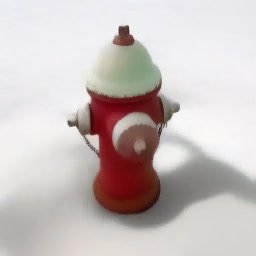} &
\includegraphics[width=0.1217\textwidth]{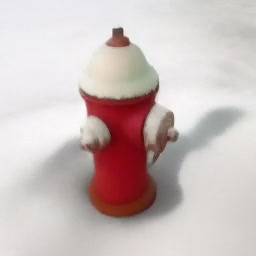} &
\includegraphics[width=0.1217\textwidth]{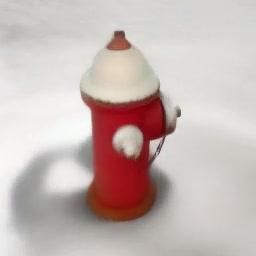} &
\includegraphics[width=0.1217\textwidth]{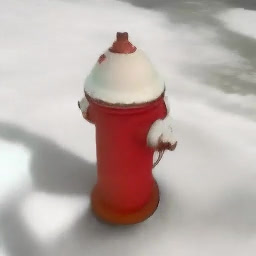} &
\includegraphics[width=0.1217\textwidth]{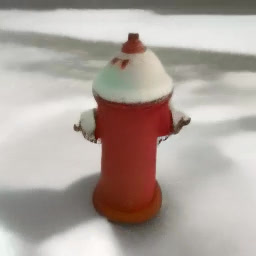} \\
\multicolumn{8}{c}{\textit{a red fire hydrant in the snow}} \\

\includegraphics[width=0.1217\textwidth]{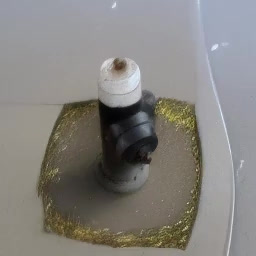} &
\includegraphics[width=0.1217\textwidth]{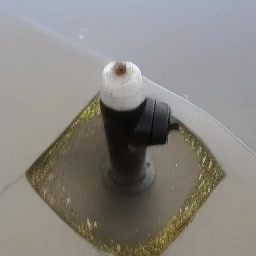} &
\includegraphics[width=0.1217\textwidth]{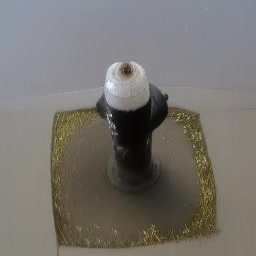} &
\includegraphics[width=0.1217\textwidth]{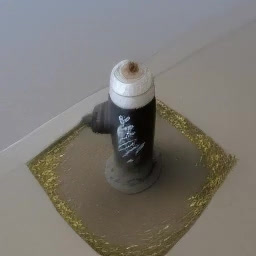} &
\includegraphics[width=0.1217\textwidth]{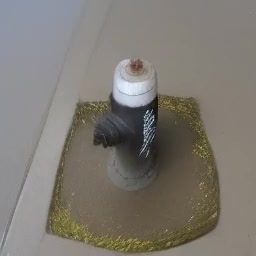} &
\includegraphics[width=0.1217\textwidth]{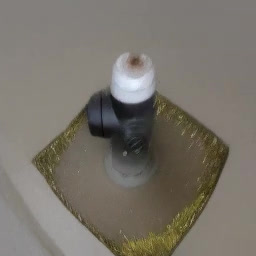} &
\includegraphics[width=0.1217\textwidth]{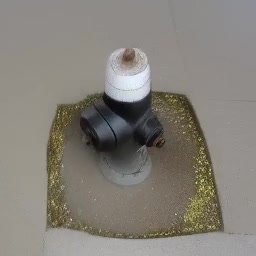} &
\includegraphics[width=0.1217\textwidth]{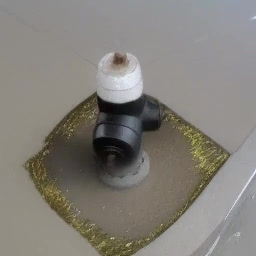} \\
\multicolumn{8}{c}{\textit{a fire hydrant on the sidewalk}} \\

\includegraphics[width=0.1217\textwidth]{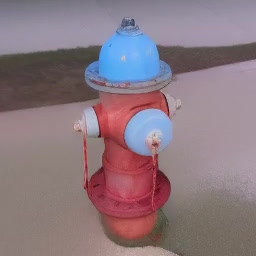} &
\includegraphics[width=0.1217\textwidth]{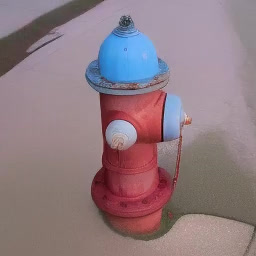} &
\includegraphics[width=0.1217\textwidth]{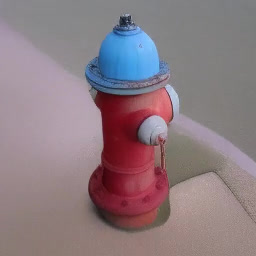} &
\includegraphics[width=0.1217\textwidth]{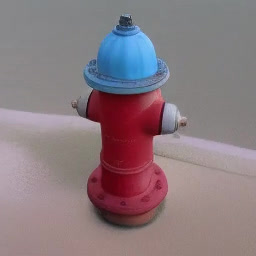} &
\includegraphics[width=0.1217\textwidth]{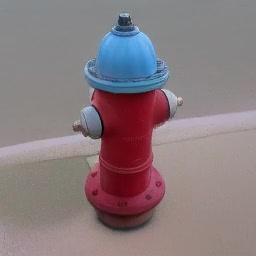} &
\includegraphics[width=0.1217\textwidth]{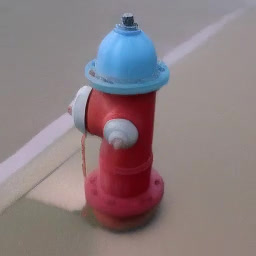} &
\includegraphics[width=0.1217\textwidth]{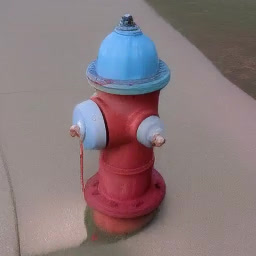} &
\includegraphics[width=0.1217\textwidth]{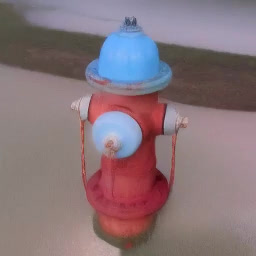} \\
\multicolumn{8}{c}{\textit{a red and blue fire hydrant}} \\

\includegraphics[width=0.1217\textwidth]{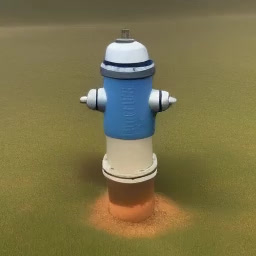} &
\includegraphics[width=0.1217\textwidth]{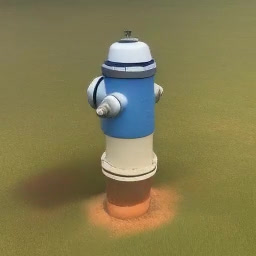} &
\includegraphics[width=0.1217\textwidth]{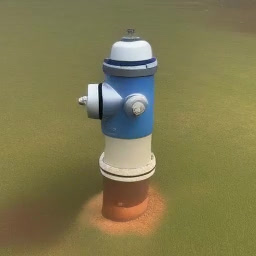} &
\includegraphics[width=0.1217\textwidth]{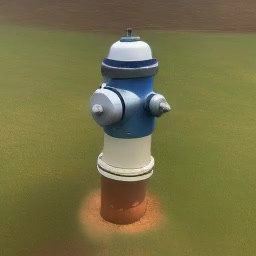} &
\includegraphics[width=0.1217\textwidth]{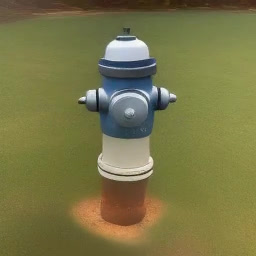} &
\includegraphics[width=0.1217\textwidth]{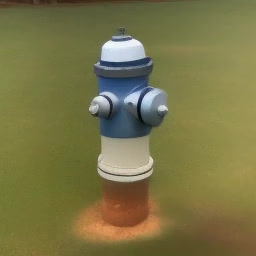} &
\includegraphics[width=0.1217\textwidth]{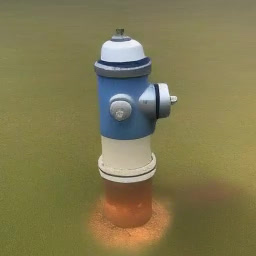} &
\includegraphics[width=0.1217\textwidth]{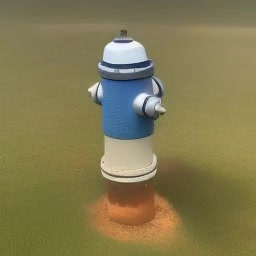} \\
\multicolumn{8}{c}{\textit{a blue and white fire hydrant sitting in the grass}} \\

\includegraphics[width=0.1217\textwidth]{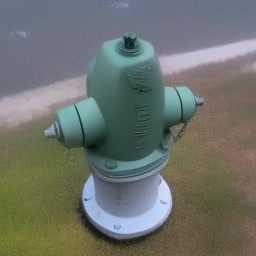} &
\includegraphics[width=0.1217\textwidth]{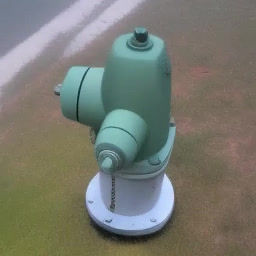} &
\includegraphics[width=0.1217\textwidth]{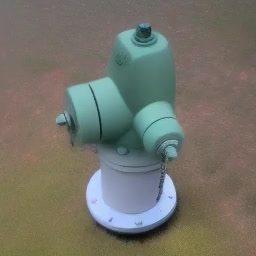} &
\includegraphics[width=0.1217\textwidth]{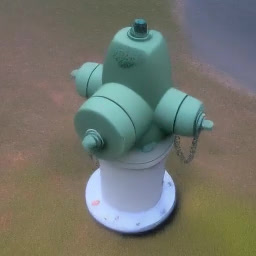} &
\includegraphics[width=0.1217\textwidth]{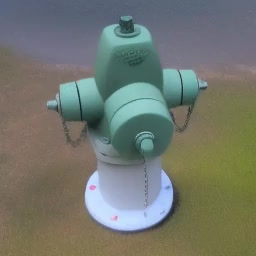} &
\includegraphics[width=0.1217\textwidth]{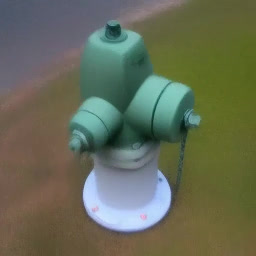} &
\includegraphics[width=0.1217\textwidth]{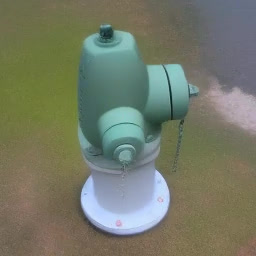} &
\includegraphics[width=0.1217\textwidth]{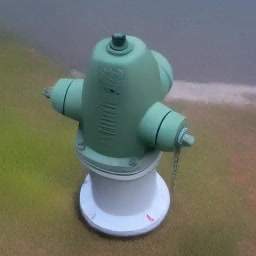} \\
\multicolumn{8}{c}{\textit{a green fire hydrant with a tag on it}} \\

\end{tabular}
\caption{
\textbf{Additional examples of our method.}
Given a text prompt as input, we generate a smooth trajectory around an object with our autoregressive generation scheme (\cref{subsec:autoreg-gen}).
Please see the supplemental video for animations of the generated samples.
}
\label{fig:suppl-uncond-2}
\end{figure*}

\begin{figure*}
\centering
\setlength\tabcolsep{0pt}
\renewcommand\cellset{\renewcommand\arraystretch{0}%
\setlength\extrarowheight{0pt}}
\begin{tabular}{cccccccc}

\includegraphics[width=0.1217\textwidth]{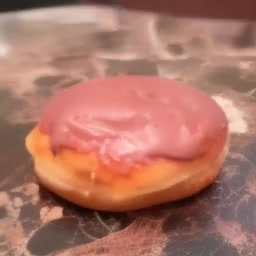} &
\includegraphics[width=0.1217\textwidth]{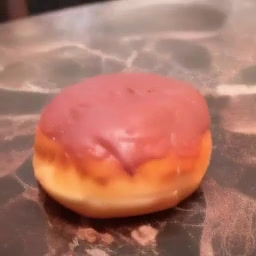} &
\includegraphics[width=0.1217\textwidth]{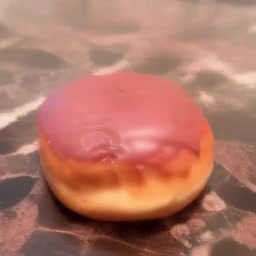} &
\includegraphics[width=0.1217\textwidth]{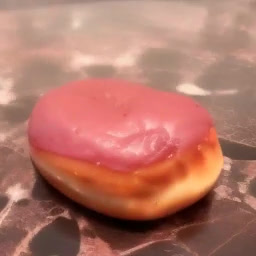} &
\includegraphics[width=0.1217\textwidth]{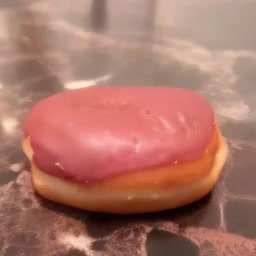} &
\includegraphics[width=0.1217\textwidth]{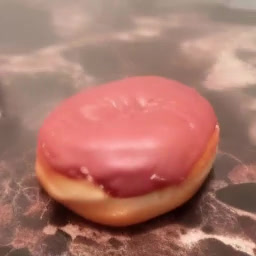} &
\includegraphics[width=0.1217\textwidth]{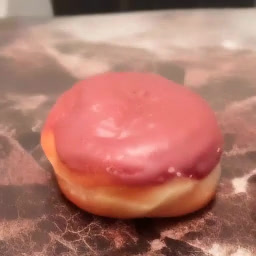} &
\includegraphics[width=0.1217\textwidth]{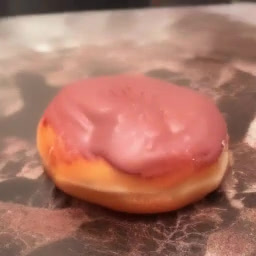} \\
\multicolumn{8}{c}{\textit{a glazed donut sitting on a marble counter}} \\

\includegraphics[width=0.1217\textwidth]{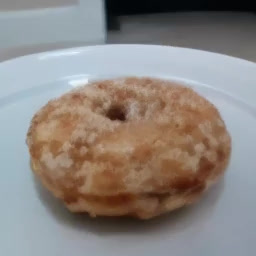} &
\includegraphics[width=0.1217\textwidth]{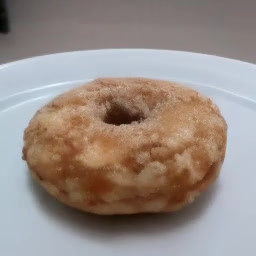} &
\includegraphics[width=0.1217\textwidth]{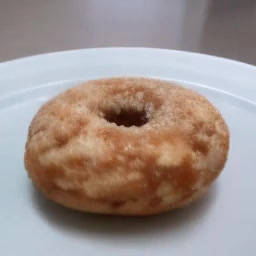} &
\includegraphics[width=0.1217\textwidth]{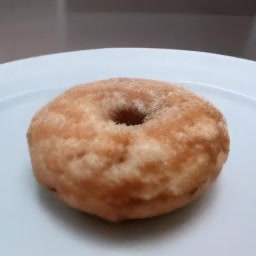} &
\includegraphics[width=0.1217\textwidth]{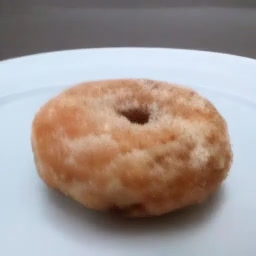} &
\includegraphics[width=0.1217\textwidth]{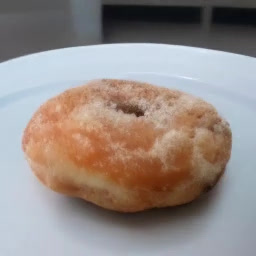} &
\includegraphics[width=0.1217\textwidth]{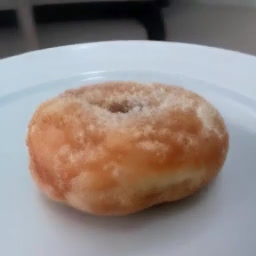} &
\includegraphics[width=0.1217\textwidth]{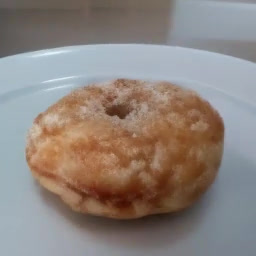} \\
\multicolumn{8}{c}{\textit{a donut on a clear plate}} \\

\includegraphics[width=0.1217\textwidth]{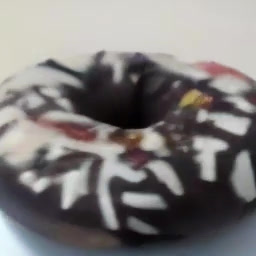} &
\includegraphics[width=0.1217\textwidth]{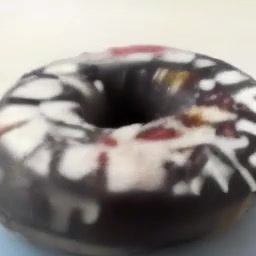} &
\includegraphics[width=0.1217\textwidth]{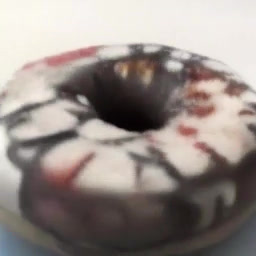} &
\includegraphics[width=0.1217\textwidth]{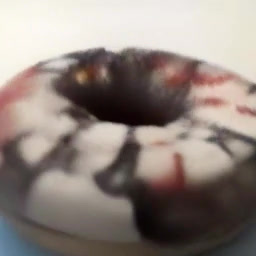} &
\includegraphics[width=0.1217\textwidth]{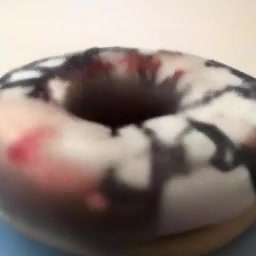} &
\includegraphics[width=0.1217\textwidth]{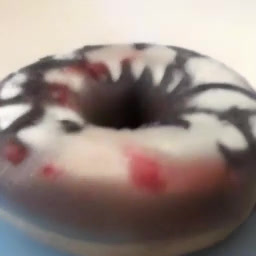} &
\includegraphics[width=0.1217\textwidth]{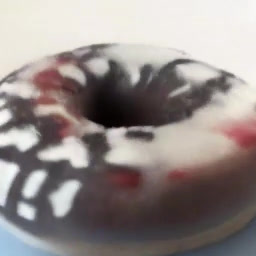} &
\includegraphics[width=0.1217\textwidth]{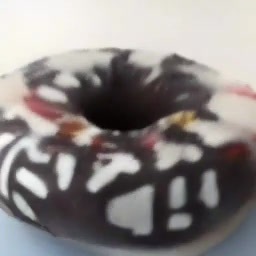} \\
\multicolumn{8}{c}{\textit{a chocolate donut with sprinkles on it}} \\

\includegraphics[width=0.1217\textwidth]{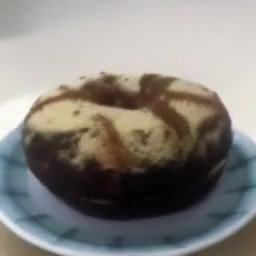} &
\includegraphics[width=0.1217\textwidth]{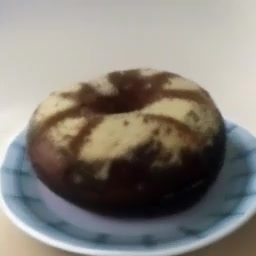} &
\includegraphics[width=0.1217\textwidth]{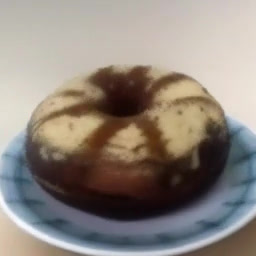} &
\includegraphics[width=0.1217\textwidth]{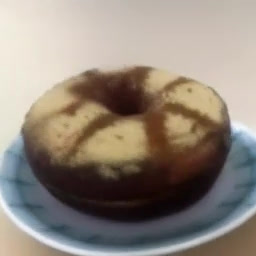} &
\includegraphics[width=0.1217\textwidth]{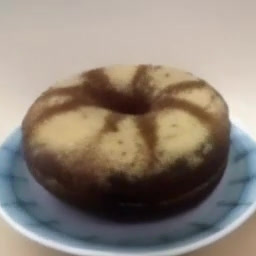} &
\includegraphics[width=0.1217\textwidth]{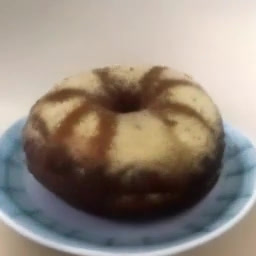} &
\includegraphics[width=0.1217\textwidth]{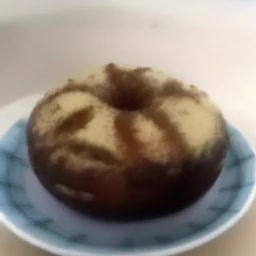} &
\includegraphics[width=0.1217\textwidth]{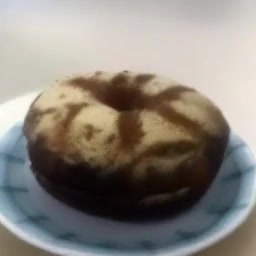} \\
\multicolumn{8}{c}{\textit{a donut on a plate with a hole in it}} \\

\includegraphics[width=0.1217\textwidth]{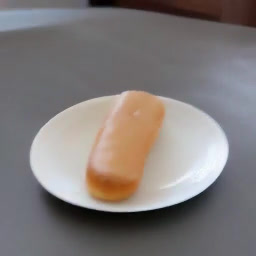} &
\includegraphics[width=0.1217\textwidth]{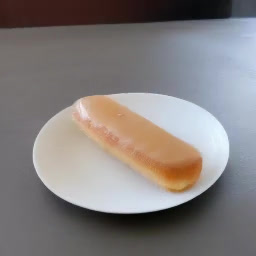} &
\includegraphics[width=0.1217\textwidth]{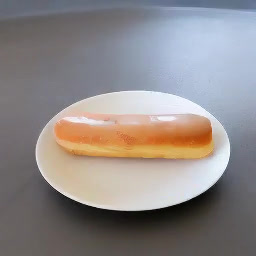} &
\includegraphics[width=0.1217\textwidth]{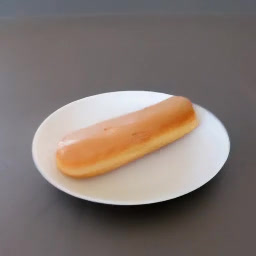} &
\includegraphics[width=0.1217\textwidth]{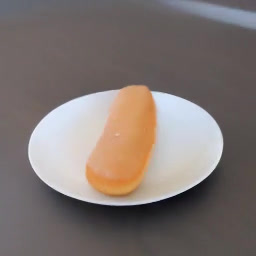} &
\includegraphics[width=0.1217\textwidth]{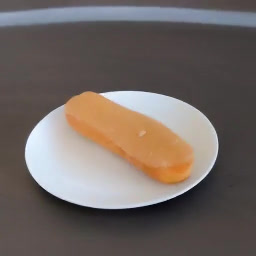} &
\includegraphics[width=0.1217\textwidth]{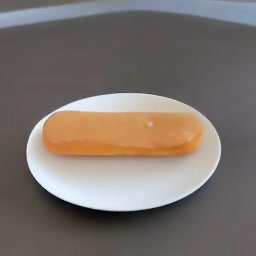} &
\includegraphics[width=0.1217\textwidth]{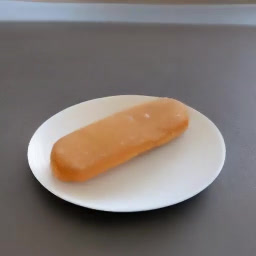} \\
\multicolumn{8}{c}{\textit{a large donut on a plate on a table}} \\

\includegraphics[width=0.1217\textwidth]{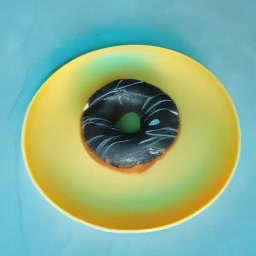} &
\includegraphics[width=0.1217\textwidth]{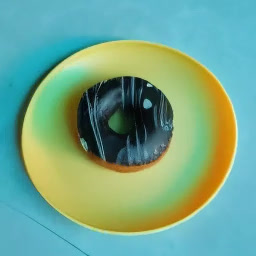} &
\includegraphics[width=0.1217\textwidth]{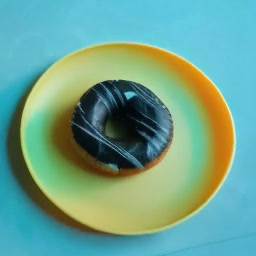} &
\includegraphics[width=0.1217\textwidth]{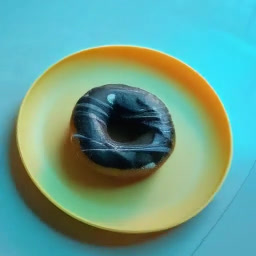} &
\includegraphics[width=0.1217\textwidth]{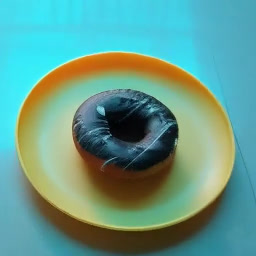} &
\includegraphics[width=0.1217\textwidth]{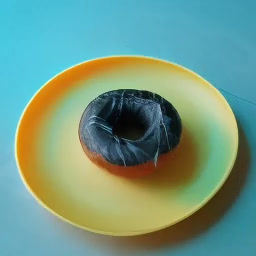} &
\includegraphics[width=0.1217\textwidth]{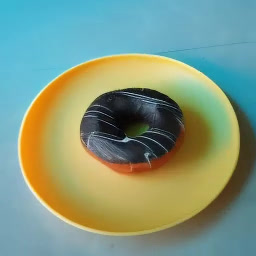} &
\includegraphics[width=0.1217\textwidth]{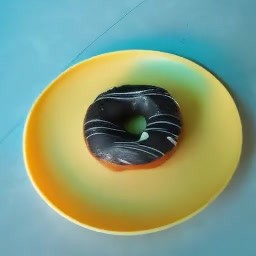} \\
\multicolumn{8}{c}{\textit{a yellow plate with a donut on it}} \\

\includegraphics[width=0.1217\textwidth]{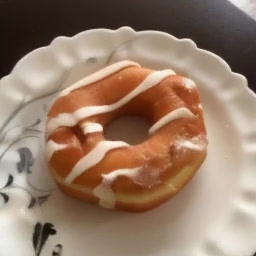} &
\includegraphics[width=0.1217\textwidth]{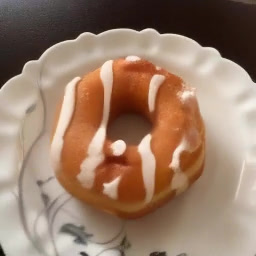} &
\includegraphics[width=0.1217\textwidth]{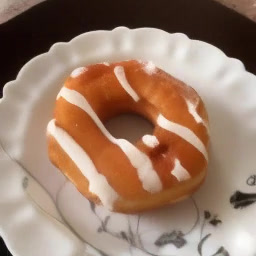} &
\includegraphics[width=0.1217\textwidth]{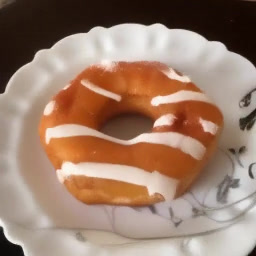} &
\includegraphics[width=0.1217\textwidth]{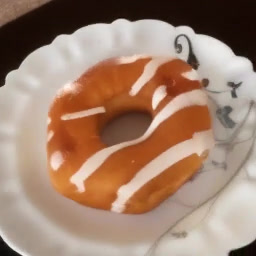} &
\includegraphics[width=0.1217\textwidth]{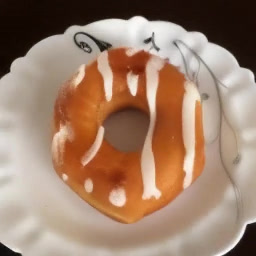} &
\includegraphics[width=0.1217\textwidth]{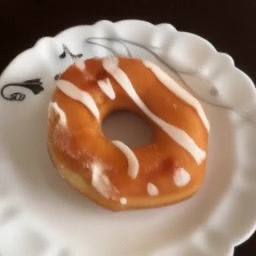} &
\includegraphics[width=0.1217\textwidth]{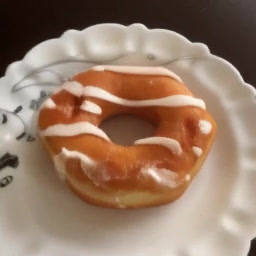} \\
\multicolumn{8}{c}{\textit{a white plate with a donut on it}} \\

\includegraphics[width=0.1217\textwidth]{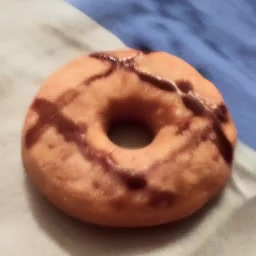} &
\includegraphics[width=0.1217\textwidth]{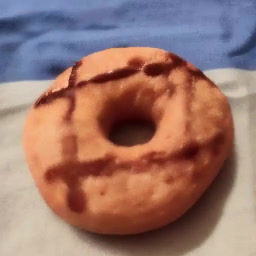} &
\includegraphics[width=0.1217\textwidth]{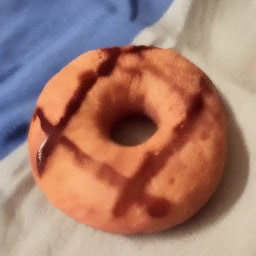} &
\includegraphics[width=0.1217\textwidth]{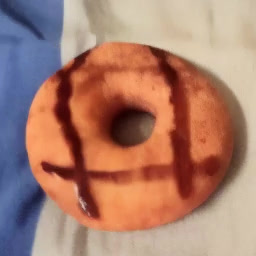} &
\includegraphics[width=0.1217\textwidth]{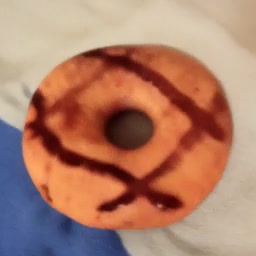} &
\includegraphics[width=0.1217\textwidth]{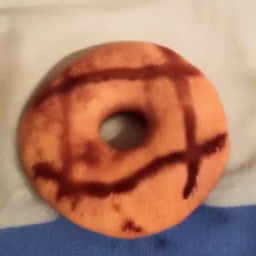} &
\includegraphics[width=0.1217\textwidth]{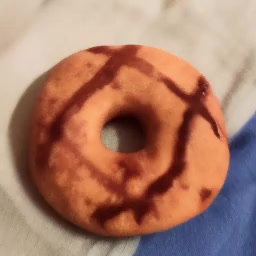} &
\includegraphics[width=0.1217\textwidth]{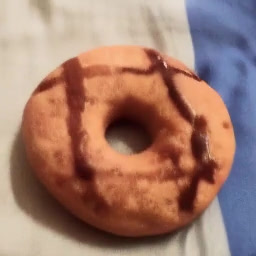} \\
\multicolumn{8}{c}{\textit{a donut sitting on a cloth}} \\

\end{tabular}
\caption{
\textbf{Additional examples of our method.}
Given a text prompt as input, we generate a smooth trajectory around an object with our autoregressive generation scheme (\cref{subsec:autoreg-gen}).
Please see the supplemental video for animations of the generated samples.
}
\label{fig:suppl-uncond-3}
\end{figure*}

\begin{figure*}
\centering
\setlength\tabcolsep{0pt}
\renewcommand\cellset{\renewcommand\arraystretch{0}%
\setlength\extrarowheight{0pt}}
\begin{tabular}{cccccccc}

\includegraphics[width=0.1217\textwidth]{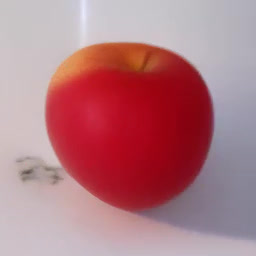} &
\includegraphics[width=0.1217\textwidth]{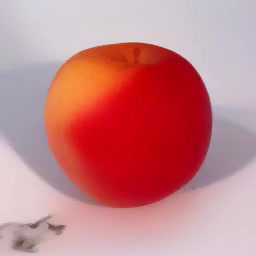} &
\includegraphics[width=0.1217\textwidth]{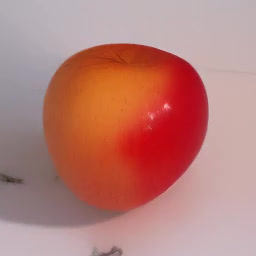} &
\includegraphics[width=0.1217\textwidth]{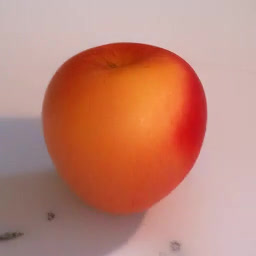} &
\includegraphics[width=0.1217\textwidth]{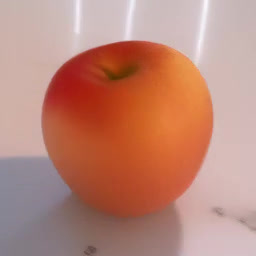} &
\includegraphics[width=0.1217\textwidth]{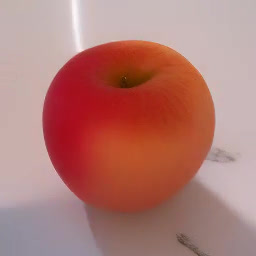} &
\includegraphics[width=0.1217\textwidth]{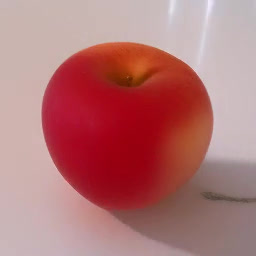} &
\includegraphics[width=0.1217\textwidth]{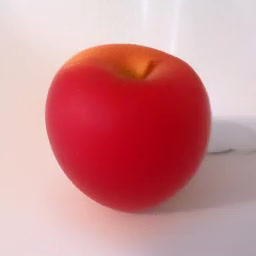} \\
\multicolumn{8}{c}{\textit{a red apple on a white counter top}} \\

\includegraphics[width=0.1217\textwidth]{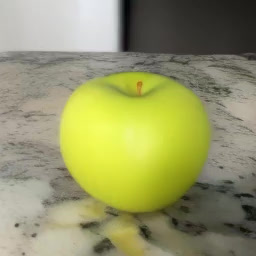} &
\includegraphics[width=0.1217\textwidth]{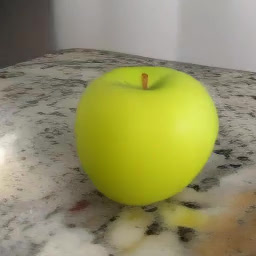} &
\includegraphics[width=0.1217\textwidth]{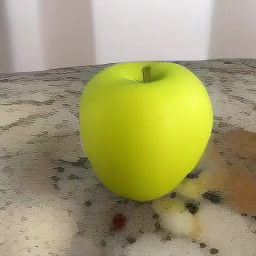} &
\includegraphics[width=0.1217\textwidth]{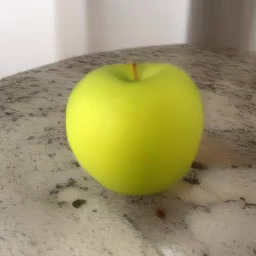} &
\includegraphics[width=0.1217\textwidth]{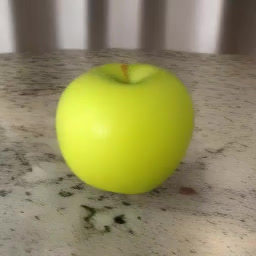} &
\includegraphics[width=0.1217\textwidth]{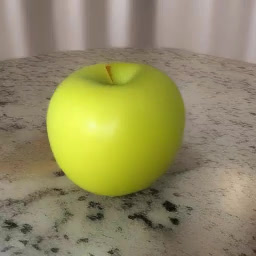} &
\includegraphics[width=0.1217\textwidth]{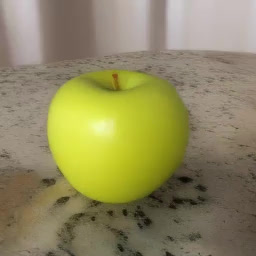} &
\includegraphics[width=0.1217\textwidth]{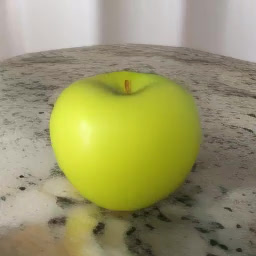} \\
\multicolumn{8}{c}{\textit{a green apple sitting on a counter}} \\

\includegraphics[width=0.1217\textwidth]{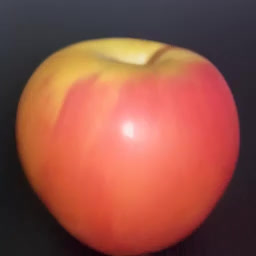} &
\includegraphics[width=0.1217\textwidth]{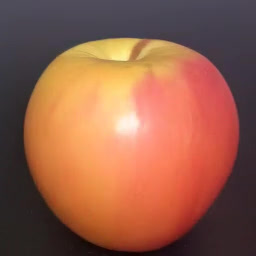} &
\includegraphics[width=0.1217\textwidth]{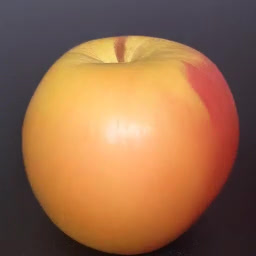} &
\includegraphics[width=0.1217\textwidth]{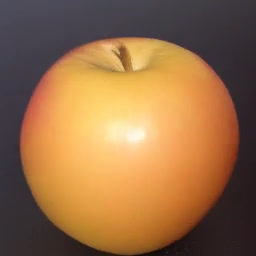} &
\includegraphics[width=0.1217\textwidth]{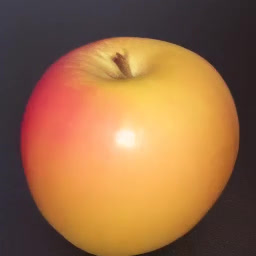} &
\includegraphics[width=0.1217\textwidth]{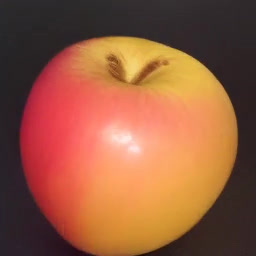} &
\includegraphics[width=0.1217\textwidth]{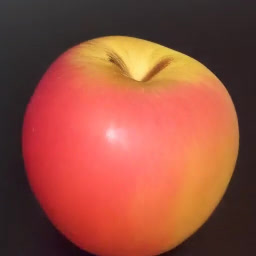} &
\includegraphics[width=0.1217\textwidth]{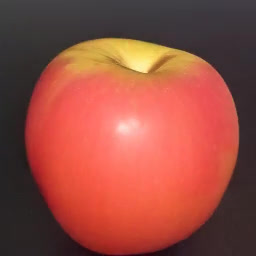} \\
\multicolumn{8}{c}{\textit{a red and yellow apple}} \\

\includegraphics[width=0.1217\textwidth]{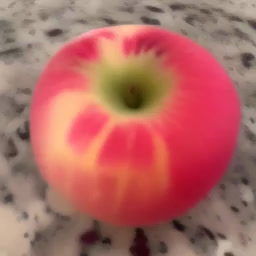} &
\includegraphics[width=0.1217\textwidth]{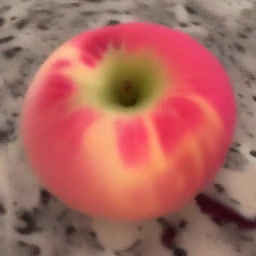} &
\includegraphics[width=0.1217\textwidth]{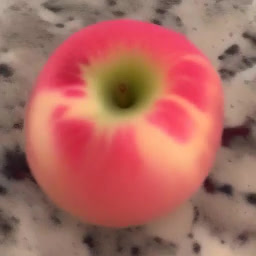} &
\includegraphics[width=0.1217\textwidth]{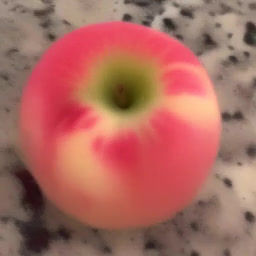} &
\includegraphics[width=0.1217\textwidth]{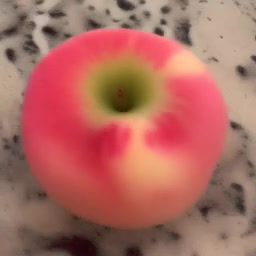} &
\includegraphics[width=0.1217\textwidth]{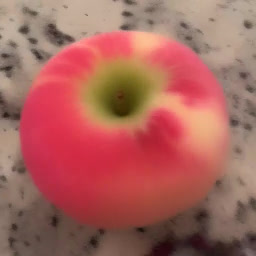} &
\includegraphics[width=0.1217\textwidth]{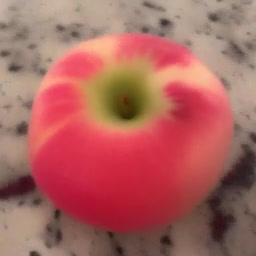} &
\includegraphics[width=0.1217\textwidth]{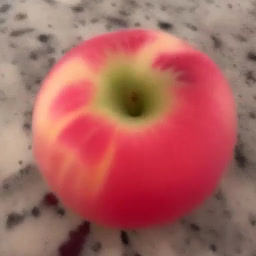} \\
\multicolumn{8}{c}{\textit{a red and yellow apple}} \\

\includegraphics[width=0.1217\textwidth]{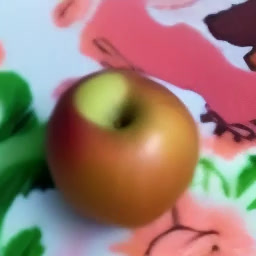} &
\includegraphics[width=0.1217\textwidth]{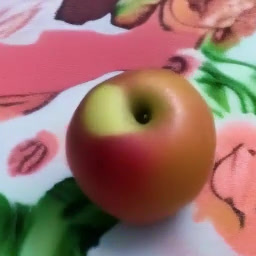} &
\includegraphics[width=0.1217\textwidth]{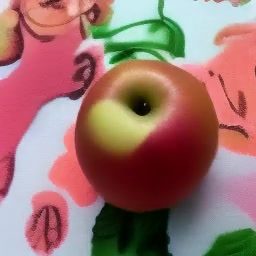} &
\includegraphics[width=0.1217\textwidth]{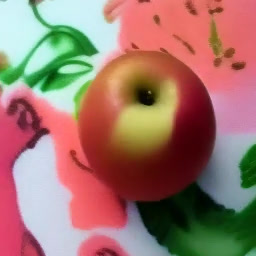} &
\includegraphics[width=0.1217\textwidth]{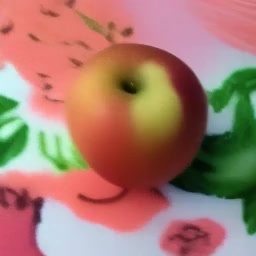} &
\includegraphics[width=0.1217\textwidth]{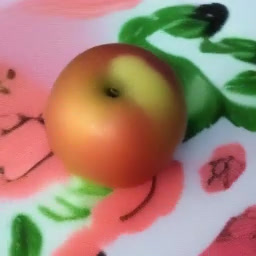} &
\includegraphics[width=0.1217\textwidth]{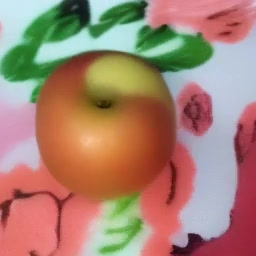} &
\includegraphics[width=0.1217\textwidth]{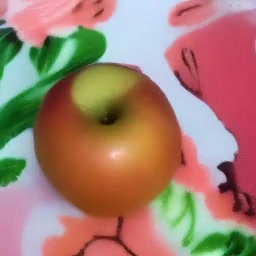} \\
\multicolumn{8}{c}{\textit{a single apple on a table cloth with a floral pattern}} \\

\includegraphics[width=0.1217\textwidth]{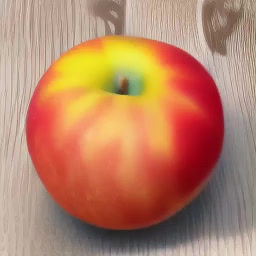} &
\includegraphics[width=0.1217\textwidth]{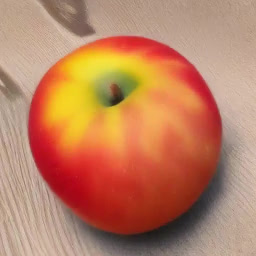} &
\includegraphics[width=0.1217\textwidth]{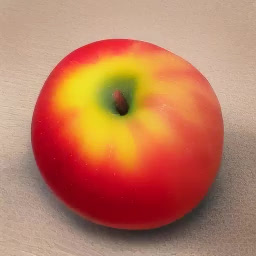} &
\includegraphics[width=0.1217\textwidth]{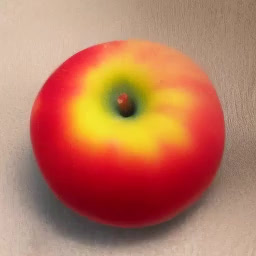} &
\includegraphics[width=0.1217\textwidth]{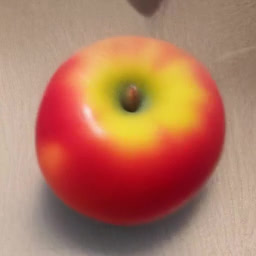} &
\includegraphics[width=0.1217\textwidth]{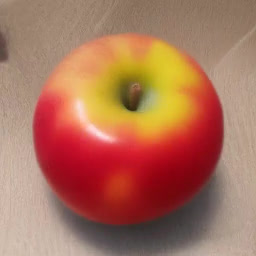} &
\includegraphics[width=0.1217\textwidth]{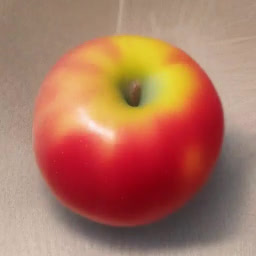} &
\includegraphics[width=0.1217\textwidth]{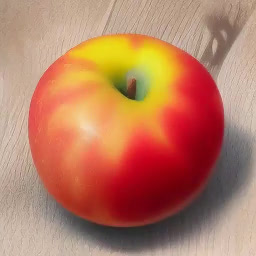} \\
\multicolumn{8}{c}{\textit{a red and yellow apple on a wooden floor}} \\

\includegraphics[width=0.1217\textwidth]{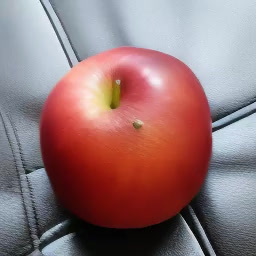} &
\includegraphics[width=0.1217\textwidth]{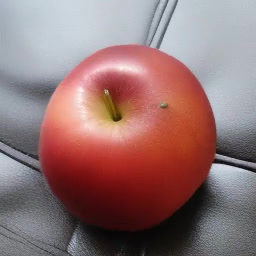} &
\includegraphics[width=0.1217\textwidth]{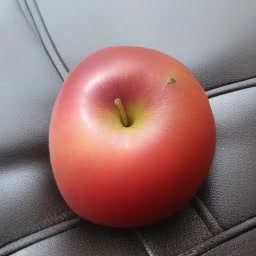} &
\includegraphics[width=0.1217\textwidth]{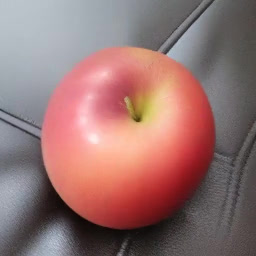} &
\includegraphics[width=0.1217\textwidth]{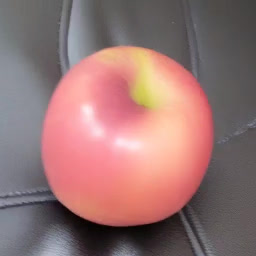} &
\includegraphics[width=0.1217\textwidth]{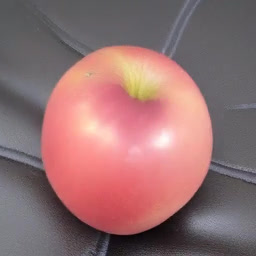} &
\includegraphics[width=0.1217\textwidth]{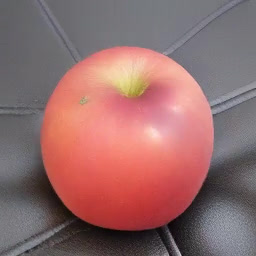} &
\includegraphics[width=0.1217\textwidth]{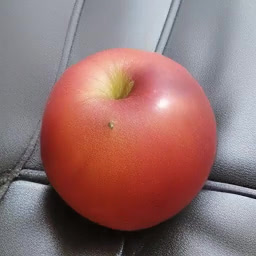} \\
\multicolumn{8}{c}{\textit{a red apple sitting on a black leather couch}} \\

\includegraphics[width=0.1217\textwidth]{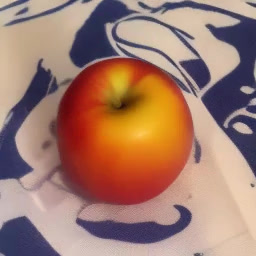} &
\includegraphics[width=0.1217\textwidth]{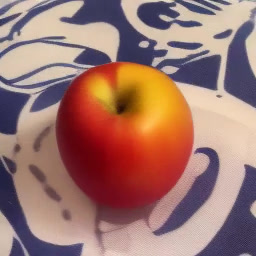} &
\includegraphics[width=0.1217\textwidth]{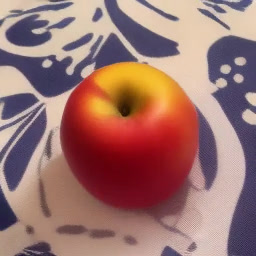} &
\includegraphics[width=0.1217\textwidth]{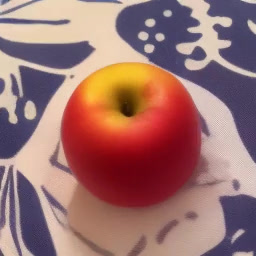} &
\includegraphics[width=0.1217\textwidth]{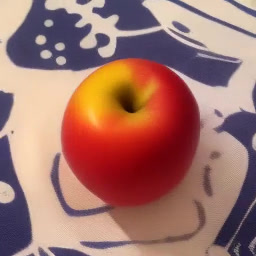} &
\includegraphics[width=0.1217\textwidth]{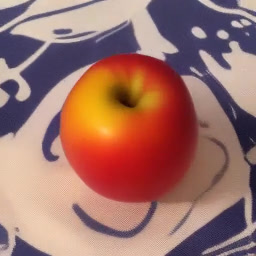} &
\includegraphics[width=0.1217\textwidth]{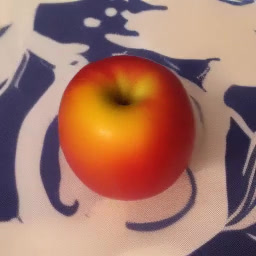} &
\includegraphics[width=0.1217\textwidth]{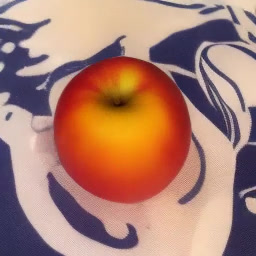} \\
\multicolumn{8}{c}{\textit{a red apple on a blue and white patterned table cloth}} \\

\end{tabular}
\caption{
\textbf{Additional examples of our method.}
Given a text prompt as input, we generate a smooth trajectory around an object with our autoregressive generation scheme (\cref{subsec:autoreg-gen}).
Please see the supplemental video for animations of the generated samples.
}
\label{fig:suppl-uncond-4}
\end{figure*}

We generate images in a similar fashion as in \cref{subsec:res-uncond}.
Concretely, we sample an (unobserved) image caption from the test set for the first batch and generate $N{=}10$ images with a guidance scale~\cite{ho2022classifier} of $\lambda_{cfg}{=}7.5$.
Then we set $\lambda_{cfg}{=}{0}$ for subsequent batches and create a total of 100 images per object.
We show additional results in ~\cref{fig:suppl-uncond-1,fig:suppl-uncond-2,fig:suppl-uncond-3,fig:suppl-uncond-4}.

\section{Optimizing a NeRF/NeuS}
\label{sec:opt-nerf-neus}

\begin{figure*}
\centering
\includegraphics[width=\textwidth]{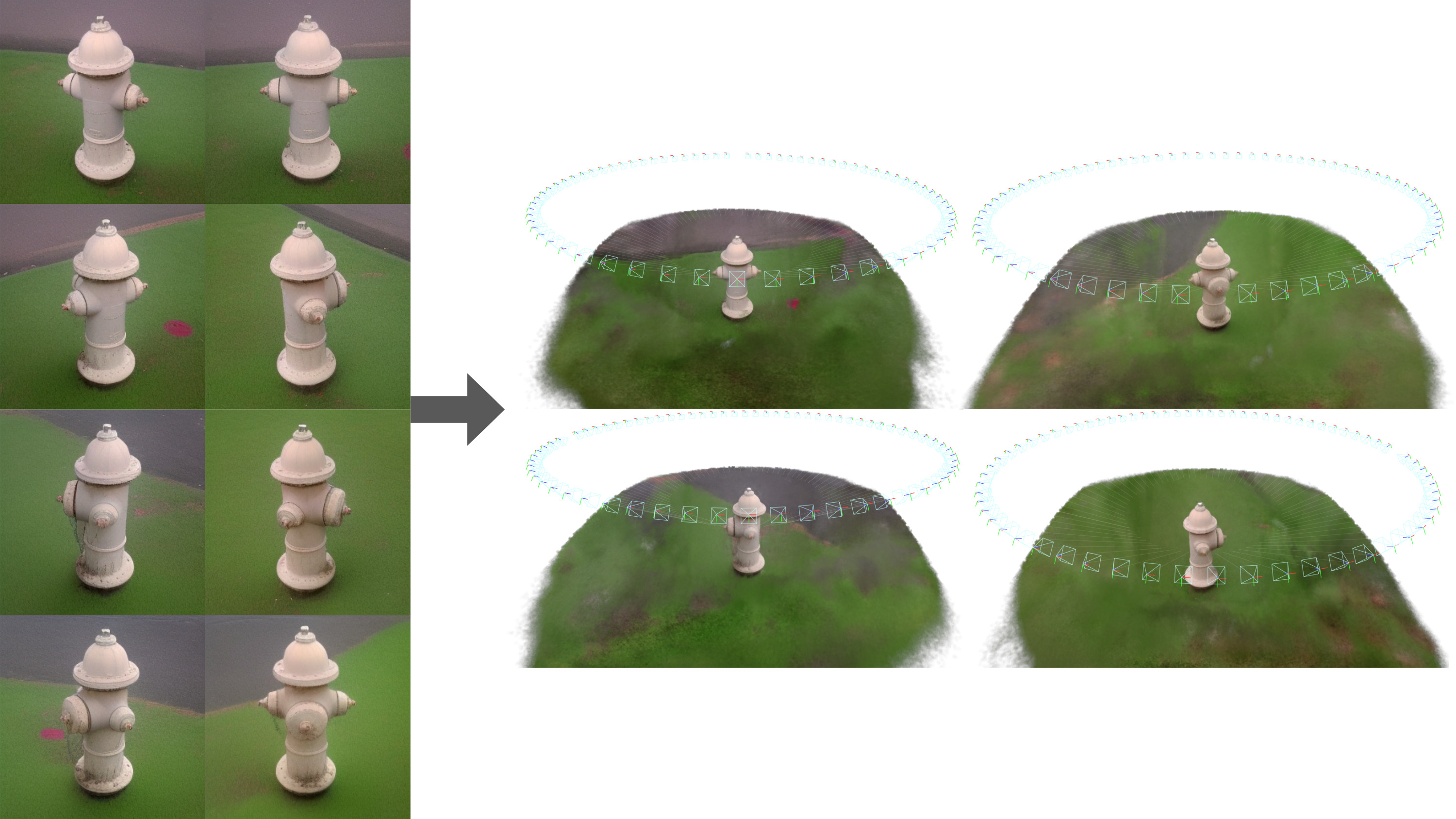}
\caption{
\textbf{NeRF~\cite{mildenhall2021nerf} optimization from our generated images.}
Left: given a text prompt as input, we generate a smooth trajectory around an object with our autoregressive generation scheme (\cref{subsec:autoreg-gen}).
In total, we generate 100 images at different camera positions.
Right: we create a NeRF using Instant-NGP~\cite{muller2022instant} from the generated images.
We show the camera positions of the generated images on top of the optimized radiance field.
}
\label{fig:suppl-nerf-ingp-hydrant}
\end{figure*}

\begin{figure*}
\centering
\includegraphics[width=\textwidth]{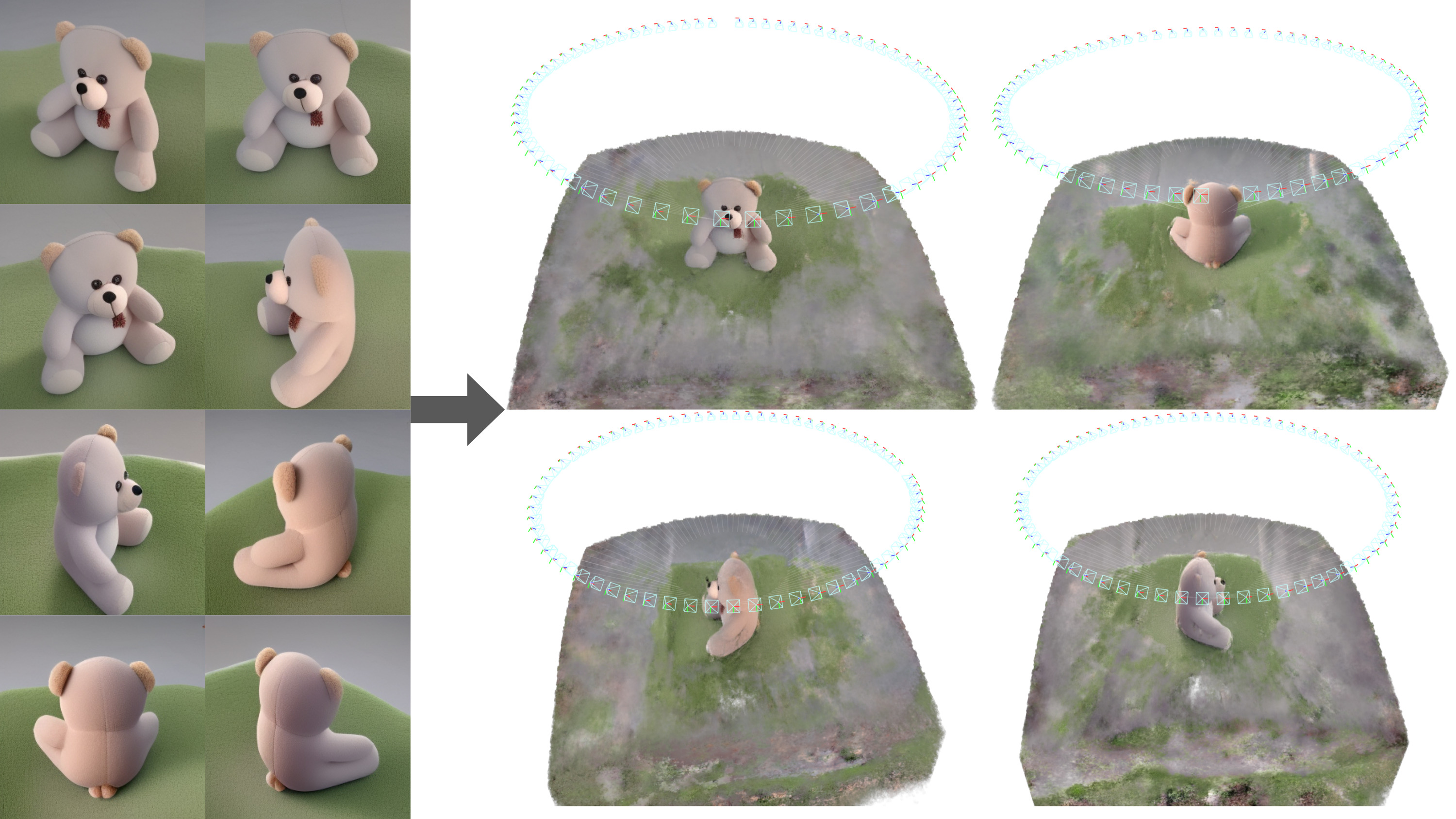}
\caption{
\textbf{NeRF~\cite{mildenhall2021nerf} optimization from our generated images.}
Left: given a text prompt as input, we generate a smooth trajectory around an object with our autoregressive generation scheme (\cref{subsec:autoreg-gen}).
In total, we generate 100 images at different camera positions.
Right: we create a NeRF using Instant-NGP~\cite{muller2022instant} from the generated images.
We show the camera positions of the generated images on top of the optimized radiance field.
}
\label{fig:suppl-nerf-ingp-teddybear}
\end{figure*}

\begin{figure*}
\centering
\includegraphics[width=\textwidth]{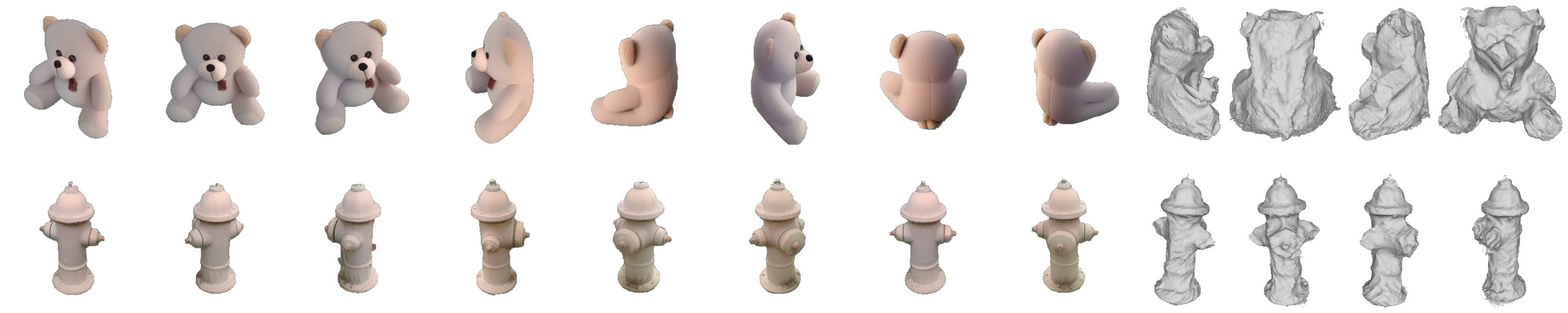}
\caption{
\textbf{Mesh extraction from our generated images.}
Given a text prompt as input, we generate a smooth trajectory around an object with our autoregressive generation scheme (\cref{subsec:autoreg-gen}).
In total, we generate 100 images at different camera positions and mask-out the background with Carvekit~\cite{carvekit}.
We then optimize a NeuS~\cite{wang2021neus} and extract the mesh from it (last 4 images per row).
}
\label{fig:suppl-neus}
\end{figure*}

Our method is capable of directly rendering images from novel camera positions in an autoregressive generation scheme (see \cref{subsec:autoreg-gen}).
This allows to render smooth trajectories around the same 3D object at arbitrary camera positions.
Depending on the use case, it might be desirable to obtain an explicit 3D representation of a generated 3D object (instead of using our method to autoregressively render new images).
We demonstrate that our generated images can be used directly to optimize a NeRF~\cite{mildenhall2021nerf} or NeuS~\cite{wang2021neus}.
Concretely, we optimize a NeRF with the Instant-NGP~\cite{muller2022instant} implementation from our generated images for 10K iterations (2 minutes).
Also, we extract a mesh by optimizing a NeuS with the \textit{neus-facto} implementation from SDFStudio~\cite{Yu2022SDFStudio, nerfstudio} for 20K iterations (15 minutes).
First, we remove the background of our generated images by applying Carvekit~\cite{carvekit} and then start the optimization with these images.
We show results in ~\cref{fig:suppl-nerf-ingp-hydrant,fig:suppl-nerf-ingp-teddybear,fig:suppl-neus}.

\end{document}